\documentclass[conference,10pt]{IEEEtran}
\ifCLASSINFOpdf
\else
\fi

\usepackage{textcomp,amssymb}
\usepackage{hyperref}
\usepackage{setspace}
\usepackage{latexsym,fancyhdr,url}
\usepackage{enumerate}
\usepackage[linesnumbered,ruled]{algorithm2e}
\usepackage{algpseudocode}
\usepackage{graphics}
\usepackage{xparse} 
\usepackage{xspace}
\usepackage{multirow}
\usepackage{csvsimple}
\usepackage{balance}
\usepackage{booktabs}
\usepackage{colortbl}
\usepackage{fancyhdr}
\usepackage{makecell}
\usepackage{xcolor, eucal}
\usepackage{array}
\usepackage{subfigure}
\usepackage{graphicx}
\usepackage{caption}
\usepackage{tabularx}
\usepackage{dsfont}
\usepackage{enumitem}
\usepackage{float}
\usepackage{amsmath}

\definecolor{gray0}{gray}{0.9}

\newtheorem{theorem}{Theorem}

\usepackage[most]{tcolorbox}

\newcommand\rev[1]{{\color{black}  #1}}

\def \toolname{\textsc{TrajDeleter}\xspace}
\def \toolnameeval{\textsc{TrajAuditor}\xspace}
\def \orlauditor{\textsc{ORL-Auditor}\xspace}

\begin{document}
%
\title{\toolname: Enabling Trajectory Forgetting in Offline Reinforcement Learning Agents}



%
\author{\IEEEauthorblockN{Chen Gong\IEEEauthorrefmark{1},
Kecen Li\IEEEauthorrefmark{2}\IEEEauthorrefmark{3},
Jin Yao\IEEEauthorrefmark{1}, and
Tianhao Wang\IEEEauthorrefmark{1}}
\IEEEauthorblockA{\IEEEauthorrefmark{1}University of Virginia}
\IEEEauthorblockA{\IEEEauthorrefmark{2}Chinese Academy of Sciences}
\thanks{\IEEEauthorrefmark{3}Work done as a remote intern at UVA.}}


\IEEEoverridecommandlockouts
\makeatletter\def\@IEEEpubidpullup{6.5\baselineskip}\makeatother
\IEEEpubid{\parbox{\columnwidth}{
    Network and Distributed System Security (NDSS) Symposium 2025 \\
    23 - 28 February 2025, San Diego, CA, USA \\
    ISBN 979-8-9894372-8-3 \\
    https://dx.doi.org/10.14722/ndss.2025.23126 \\
    www.ndss-symposium.org
}
\hspace{\columnsep}\makebox[\columnwidth]{}}

\maketitle

\begin{abstract}
Reinforcement learning (RL) trains an agent from experiences interacting with the environment. In scenarios where online interactions are impractical, offline RL, which trains the agent using pre-collected datasets, has become popular. 
While this new paradigm presents remarkable effectiveness across various real-world domains, like healthcare and energy management, there is a growing demand to enable agents to rapidly and completely eliminate the influence of specific trajectories from both the training dataset and the trained agents.
To meet this problem, this paper advocates \toolname, the first practical approach to trajectory unlearning for offline RL agents. The key idea of \toolname is to guide the agent to demonstrate deteriorating performance when it encounters states associated with unlearning trajectories. Simultaneously, it ensures the agent maintains its original performance level when facing other remaining trajectories. Additionally, we introduce \toolnameeval, a simple yet efficient method to evaluate whether \toolname successfully eliminates the specific trajectories of influence from the offline RL agent. Extensive experiments conducted on six offline RL algorithms and three tasks demonstrate that \toolname requires only about 1.5\% of the time needed for retraining from scratch. It effectively unlearns an average of 94.8\% of the targeted trajectories yet still performs well in actual environment interactions after unlearning. The replication package and agent parameters are available online\footnote{\url{https://github.com/2019ChenGong/TrajDeleter/}}.
\end{abstract}


%

\section{Introduction}
\label{sec:intro}

Reinforcement learning (RL) develops agents to learn from trajectories (sequences of states, actions, and rewards experienced by an agent as it interacts with the environment over time) and has recently made significant strides in various complex decision-making areas, including robotics control~\cite{openai2019solving,gong2022curiosity}, recommendation systems~\cite{RL4Recommender,RL4Recommender2}, and dialogue systems~\cite{RL4Dialog,RL4Dialog2}, etc. In safety-critical areas like healthcare~\cite{RL4Treatment,RL4BGC}, and even nuclear fusion~\cite{degrave2022magnetic}, direct interaction with the environment can be hazardous, since a partially trained agent might cause damage. Thus, researchers have developed offline RL, a methodology where agents are trained using static datasets pre-collected from experts, manually programmed controllers, or even random strategies~\cite{d4rl}. Offline RL paves the way for its application in situations where online interactions are either impractical or risky, and works well on a wide range of real-world fields~\cite{RL4BGC, RL4Energy,RL4AutonomousVehicles,RL4AutonomousVehicles2}.

With the success of offline RL comes the demand to delete parts of the training sets (also referred to as machine unlearning) for various reasons. For example, legislations like the European Union's General Data Protection Regulation (GDPR)~\cite{GDPR} and the California Consumer Privacy Act (CCPA)~\cite{CCPA} empowered users with the right to request their data to be deleted. The server may want to delete some data due to security or copyright reasons (the server discovered some data is poisoned, or copyrighted)~\cite{gong2022mind}. This inspires the development of an ``unlearning'' methodology tailored for offline RL, which we term \textit{offline reinforcement unlearning}.

\noindent \textbf{Existing Solutions.} 
A naive approach to unlearning is retraining without the data required to be unlearned. Furthermore, one can partition the training data and train an ensemble of models, so during unlearning, one still retrain a model but on a partition of the training set~\cite{unlearning_01}.
The inefficiency of retraining drives the development of approximate unlearning (but focuses more on the supervised learning in the image or text domain), including gradient ascent~\cite{unlearning_02_feature,yao2024machine} and contrastive learning~\cite{zhang2024contrastive}. We discuss unlearning methods in other fields in Section~\ref{sec:related}.

The unique paradigm of RL poses challenges when applying existing approximate unlearning methods. Specifically, the data in RL are a sequence of {\it trajectories}, each in the format of a tuple of state, action, and reward. 
Ye et al. first proposed the concept of reinforcement unlearning in the {\it online} setting by using environment poisoning attacks~\cite{ye2023RLunlearning}. 
However, this is not feasible in offline RL. Our work is the first to address the need for unlearning within offline RL, specifically emphasizing unlearning at the trajectory level (Ye et al.~\cite{ye2023RLunlearning} works at the aggregated level). Trajectory-level studies~\cite{sutton2018reinforcement,du2023orl} have attracted more attention, as RL algorithms are often trained within a single environment~\cite{offline_survey}.

\noindent \textbf{Our Proposal.} This paper introduces \toolname, enabling offline RL agents to unlearn trajectories. \toolname is composed of two phases, ``forgetting'' and ``convergence training''.  The forgetting phase first minimizes the value function $Q$ (a function specific to RL to be described in Section~\ref{sec:offlineRL}) for the unlearning samples. We choose to work with $Q$ because it estimates the expected cumulative reward an agent can achieve from a state. So if the agent's cumulative reward for a state is low, it means the agent is unfamiliar with a trajectory, thus forgetting. However, this process alone will make the agent unstable for the normal, remaining samples. To alleviate this, we also maximize $Q$ on remaining samples simultaneously, balancing unlearning and preventing performance degradation. 


%

Due to the notorious unstable training problem in RL~\cite{sutton2018reinforcement,offline_survey, a3c}, \rev{which presents that the challenges encountered when the learning process of an agent does not progress consistently, resulting in erratic performance, slow convergence, or failure to learn an optimal policy.} Our strategy of achieving the two opposite directions of optimization on the unlearning and remaining dataset may fail to guarantee the convergence of the unlearned agent, leading to potential instability during the training. To mitigate this concern, the second ``convergence training'' phase minimizes the discrepancies in cumulative rewards obtained by following the original and unlearned agents when encountering states in other remaining trajectories. Theoretical analysis presents that fine-tuning the unlearned agent ensures its convergence. 

It is also crucial to evaluate if the trajectories with specific impacts have been erased from the approximate unlearned agent. This evaluation forms the fundamental basis of offline reinforcement unlearning. Du et al.~\cite{du2023orl} proposed \orlauditor to audit trajectories for offline DRL models, providing the potential tool for evaluating unlearning. However, \orlauditor can be time-consuming, owing to the extensive training required for numerous \textit{shadow agents}, which explicitly exclude the unlearning trajectories from their training set. \rev{These agents are used to simulate or mimic the behavior of a target agent to distinguish between members and non-members of the training set. Specifically,} to expedite the auditing process, we introduce \toolnameeval, which fine-tunes the original agent (needed for unlearning) to create the shadow agents. We consider these shadow agents to be trained with datasets that include the targeted unlearning trajectories. In addition, we implement state perturbations along the trajectories, producing diverse auditing bases. Referring to Du et al. proposed~\cite{du2023orl}, \toolnameeval determines the success of unlearning by comparing cumulative rewards from unlearning trajectories. It assesses the similarity between results from shadow agents and the unlearned agent, with low similarity indicating successful unlearning. \toolnameeval matches the performance of \orlauditor while requiring significantly fewer computing resources. 


\noindent \textbf{Evaluations.} We extensively experiment with six offline RL algorithms on three common Mujoco evaluation tasks~\cite{mujoco} to verify the effectiveness and efficiency of \toolname. We experimented with various unlearning rates (the proportion of data required to be forgotten in a dataset).
{\toolnameeval shows high proficiency, achieving average F1-scores of 0.88, 0.87, and 0.88 across three tasks. It achieves a 97.8\% reduction in time costs while attaining an F1-score that is 0.08 points higher compared to training shadow {agents from scratch in \orlauditor.} It is a simple yet efficient tool for determining whether a specific trajectory continues influencing the unlearned agent, paving the path for our unlearning study. 

Then, we evaluate how effective is \toolname under the assessment of \toolnameeval. Our experiments show that \toolnameeval achieves the removal of 92.7\%, 99.5\%, and 90.5\% of targeted trajectories across three tasks while requiring only 1.5\% of the time needed for retraining from scratch. The average cumulative returns show a slight difference of 2.2\%, 0.9\%, and 1.6\% between \toolname-unlearned agents and retrained agents in the three tasks on average. 



We also analyze hyper-parameters of \toolname. With an increase in the number of forgetting steps, the performance of \toolname significantly improves, resulting in the unlearned agent forgetting more trajectories. We also observe an interesting fact: with an increased number of forgetting steps, \toolname shows minimal sensitivity to the values of other hyper-parameters it introduces. 
We also conduct ablation studies to study the significance of ``convergence training'' in enhancing \toolname's efficacy, and  investigate its effectiveness in defending against trajectory poisoning attacks.

\noindent \textbf{Contributions.} In summary, our contributions are three-fold:
\begin{itemize}[leftmargin=*]
    \item {{To our knowledge, we propose the first practical trajectories-level unlearning approach, \toolname, specifically tailored for offline RL agents.}}
    \item We introduce \toolnameeval, a simple yet efficient method to assess whether the unlearning method effectively erases the specific trajectories' influence on the unlearned agent.
    \item {We perform a comprehensive evaluation of \toolname. The results present the effectiveness of \toolname in offline DRL agents trained across six distinct prevalent algorithms and three tasks.} 
\end{itemize}

\rev{We recognize that approximate unlearning's compatibility with legal requirements is uncertain, but the similar situation, where technology moves faster than the legal side, is true in other areas, like differential privacy~\cite{dp} and watermark~\cite{li2022untargeted}. Approximate unlearning retains significant research value.}


\section{Background}
\label{sec:relwork}

\subsection{Offline Reinforcement Learning}
\label{sec:offlineRL}
The training of DRL agents operates a trial-and-error paradigm, with learning driven by feedback from rewards. For example, we consider an agent responsible for controlling a car. If it accelerates when confronted with a red traffic light, it receives a penalty in the form of a negative reward. Then, this agent will update its policy to avoid accelerating when a traffic light turns red. The agent learns from such experiences by interacting with the environment during the training phase, which is called the \textit{online} RL. 

\begin{figure}[!t]
    \centering
    \setlength{\abovecaptionskip}{0pt}
    \includegraphics[width=1.0\linewidth]{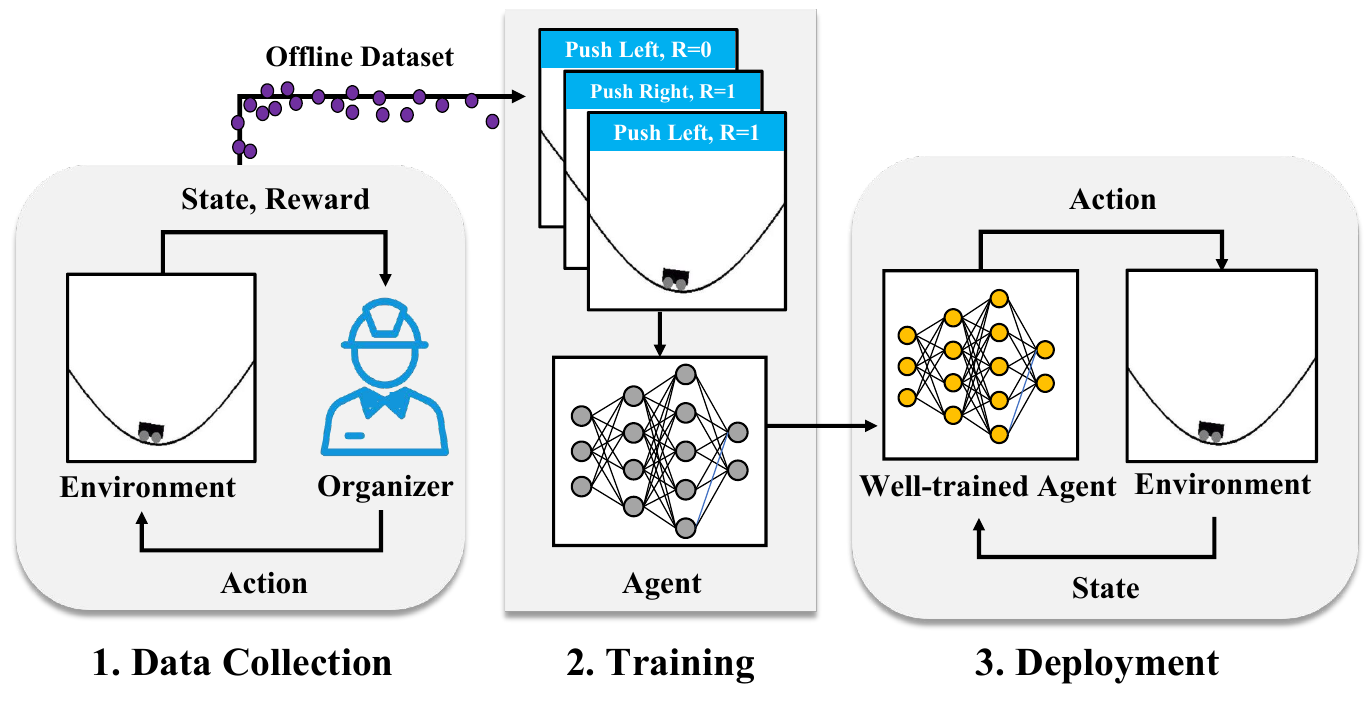}
    \caption{An example of offline RL implementations. Initially, the organizer gathers trajectories through interactions with the environments, forming the offline dataset. Then, the agent is trained using this static dataset. Once fully trained, the agent is deployed in real-world applications.}
    \vspace{-5mm}
    \label{fig:offline_RL}
\end{figure}

However, the online settings are not always feasible. For instance, a hospital aims to train an RL agent to recommend treatments to future patients. The online RL would make the agent propose treatments, observe the results, and adjust its policy accordingly. Since it is ethically and practically problematic to experiment with patients' health, the offline RL, which is designed to train agents from a \textit{pre-collected} and \textit{static} dataset eliminating the need for interactions with real environments during the training stages, is more suitable. \rev{We highlight that online RL trains an agent in real-time through continuous interaction with the environment, updating its policy based on received feedback (rewards or penalties). In contrast, offline RL trains an agent using a fixed, pre-collected dataset without accessing the environment during the training phase.} Then, as presented in Figure~\ref{fig:offline_RL}, we outline the three-step implementation process of offline RL, including data collection, training the offline RL agent, and deployment.

\noindent \textbf{Data Collection. } At timestep $t$, an agent observes a state $s_t$ and executes an action $a_t$ determined by the agent's \textit{policy}. The agent selects an action $a_t$ according to the policy $\pi(\cdot)$, which dictates which actions to execute in a given state $s_t$, i.e., $a_t \sim \pi(\cdot|s_t)$. After taking an action, the agent obtains an immediate reward from the environmental reward function, denoted as $r_t = \mathcal{R}(s_t,a_t)$. Then, the environment transitions to a new state $s_{t+1}$, determined by the transition function $s_{t+1} \sim \mathcal{T}(\cdot|s_t,a_t)$. This sequence persists until the agent reaches a termination state $s_T$. \rev{This process results in a \textit{trajectory},}
\rev{$$\tau: ( s_0,a_0,r_0, s_1,a_1,r_1,s_2, a_2,r_2,\cdots , s_{|\tau|}, a_{|\tau|}, r_{|\tau|}).$$}
Besides, the four-tuple $\langle s_t, a_t, r_t, s_{t+1} \rangle$ in each trajectory is referred to as the \textit{transition}. The organizer, also named the data provider (e.g., the hospital), gathers trajectories through multiple, distinct policies interacting with the environments. Then, these trajectories consist of the offline dataset $\mathcal{D}$ = $\{ \tau^i \}_{i=1}^N$, where $N$ represents the total number of trajectories.

\noindent \textbf{Training.} In online RL, an agent aims to learn an optimal policy $\pi^*$ that can get high expected performance from the environment. Specifically, the {\em cumulative discounted return} is the sum of discounted rewards within a given trajectory: $R(\tau) = \sum_{i=0}^T \gamma^i r_i$, where $\gamma \in (0,1)$ represents the discount factor~\cite{sutton2018reinforcement} and $T$ is the length of the trajectory. Hence, the objective of online RL can be formulated to optimize the policy and achieve the highest possible cumulative discounted return, which is defined as,
\begin{equation}
    \pi^\ast = \arg\max_\pi \mathbb{E}_{\tau \sim \pi} \left[ R(\tau^\pi)\right],
    \label{eq:rl_target}
\end{equation}
where $\tau^\pi$ indicates the trajectory generated by the policy $\pi$. The solution to Eq.~(\ref{eq:rl_target}) could be accomplished by maximizing \textit{action-state value function} $Q^\pi(s, a)$~\cite{sutton2018reinforcement}, which is defined as,
\begin{equation}     
    Q^\pi(s,a) =  \mathbb{E}_{\tau \sim \pi} \left[ R(\tau^\pi) | s_0=s, a_0=a \right].
    \label{eq:value_func}
\end{equation}
$Q^\pi(s, a)$ quantifies the expected cumulative reward an agent can achieve starting from a state $s$, taking action $a$, and following a policy $\pi$. In other words, it serves as a metric for how good it is for an agent to execute action $a$ while being in state $s$. Thus, the objective in Eq. (\ref{eq:rl_target}) can be reformulated as, $
    \pi^\ast = \arg\max_\pi \mathbb{E}_{a\sim \pi}\left[Q^\pi(s, a)\right], \forall s \in \mathcal{S},$
where $\mathcal{S}$ refers to the state space. Next, we explain how to optimize the policy in offline RL settings. 

In offline RL, the agent is only allowed to learn from the trajectories present in the offline dataset, which is defined as  $\mathcal{D}$. The agent cannot access states absent from the given offline dataset. Thus, distinguished from considering all states in state space, offline RL only takes into account the states found within the offline dataset. In particular, offline RL initially requires the agent to derive an understanding of the dynamical system underlying the environment entirely from the pre-collected offline dataset. Then, it needs to establish a policy $\pi(a|s)$ that maximizes possible performance when \textit{actually used to interact with the environment}~\cite{levine2020offline}.


\noindent \textbf{Deployment.} The well-trained offline agents are prepared for deployment. Their deployment ensures that they can make decisions based on the rich trajectories derived from the datasets, ensuring efficiency and safety in real-time applications.

\subsection{Machine Unlearning}
In response to the demands of recently introduced legislation, like the European Union's General Data Protection Regulation (GDPR)~\cite{GDPR}, a new branch of privacy-preserving machine learning arises, known as \textit{machine unlearning}. This concept requires that the specified training data points and their influence can be erased completely and quickly from both the training dataset and trained model~\cite{unlearning_01}. Specifically, we utilize the algorithm to train a model on a dataset. The well-trained model performs certain functions, such as classification, regression, and more. Upon receiving requests for the model to ``forget'' a subset of the training dataset, the unlearning algorithm is capable of altering the trained model so that it behaves as if it is trained solely on the remaining dataset (excluding the subset that is required to be forgotten). 

With the rapid progress in applying offline RL in reality~\cite{RL4Energy,RL4AutonomousVehicles,RL4Robots,RL4Treatment}, this paper focuses on the unlearning requirement in the field of offline RL. We propose \toolname, aiming to ``unlearn" selected training trajectories by updating the trained agent to completely eliminate the influence of these trajectories from the updated agent.

\subsection{Unlearning Scenarios}
\label{sec:scenario}
We outline the key requirements for advancing offline reinforcement unlearning, highlighting three typical scenarios.

\noindent \textbf{Privacy Concerns.} Unlearning has become especially relevant in privacy and data protection laws, including the European Union's GDPR~\cite{GDPR}, which enforces a ``right to erasure.'' This right also should be involved in offline RL. For instance, we use the entire offline dataset to train an agent that performs brilliantly in real-world tasks. However, there may be instances where institutions need to instruct these trained agents to ``forget'' certain trajectories containing sensitive information, like credit card details or confidential communications. Given the potential for privacy attacks on training data from agents~\cite{pan2019you,gomrokchi2022membership,yang2023valuebased}, implementing trajectory unlearning becomes crucial to mitigate these concerns.

\noindent \textbf{Trajectory Poisoning.} 
We focus on poisoning attacks within the RL area~\cite{harrison202266,Wu2023reward,ma2019policypoisoning}, where an attacker aims to mislead the trained agent by editing the training trajectory. This intervention greatly reduces the performance of the learning agent. 
We implement unlearning to restore the agent's performance efficiently, avoiding the time-consuming retraining.

\noindent \textbf{Copyright Issue.} Numerous studies indicate that our dataset may be susceptible to various forms of misuse, including infringements of intellectual property rights~\cite{boenisch2021systematic,copyright2021tessian}. The copyright issue is particularly pressing, given the increasing prevalence of sophisticated data utilization across domains. Our method enables the rapid and efficient removal of training trajectories that lack a clearly defined copyright source. This method helps mitigate legal risks associated with copyright violations and ensures compliance with evolving copyright laws and ethical standards in data usage. 



\section{Problem Setup and Preliminaries}

\subsection{Formal Problem Deﬁnition}
\label{subsec:notion}
We first introduce the concept of offline reinforcement unlearning and the formal framework for this problem. We assume the training dataset consists of $N$ trajectories and formally express it as
$\mathcal{D}$ = $\left\{ \tau^i | \tau^i = \langle s_t^i, a_t^i, r_t^i, s_{t+1}^i \rangle_{t=0}^{|\tau^i|} \right\}_{i=1}^N$.  Then, we define an offline RL algorithm as a function $\mathcal{A}(\cdot): \mathcal{D} \mapsto \Pi$, which maps a dataset
$\mathcal{D}$ to a trained agent within the hypothesis space $\Pi$. Then, we use the notion $\mathcal{D}_f = \left\{ \tau^i \right\}_{i=1}^M \ (M < N)$ to represent the subset $\mathcal{D}_f \subset \mathcal{D}$, which the agent is required to forget. Besides, the modified dataset is defined as $\mathcal{D}_m = \mathcal{D} \backslash \mathcal{D}_f$, indicating the portion of the dataset that we intend for the agent to retain. The offline reinforcement unlearning method $\mathcal{U}$: $\mathcal{A}(\mathcal{D}) \times \mathcal{D} \times \mathcal{D}_f \mapsto \Pi $, indicates a function that maps an agent $\mathcal{A}(\mathcal{D})$, along with the training dataset $\mathcal{D}$, and a subset $\mathcal{D}_f$ designated for removal, to a correspondingly unlearned agent within the hypothesis space $\Pi$. {The offline reinforcement unlearning process, $\mathcal{U}\left(\mathcal{A}(D), \mathcal{D}, \mathcal{D}_f\right)$, is defined as function that takes a trained agent $\pi \gets \mathcal{A}(\mathcal{D})$, a training dataset $\mathcal{D}$, and a dataset $\mathcal{D}_f$ that should be forgotten. This process aims to guarantee that the unlearned agent: $\pi' \gets \mathcal{U}\left(\mathcal{A}(D), \mathcal{D}, \mathcal{D}_f\right)$ behaves as an agent directly trained on the training dataset excluding $\mathcal{D}_f$.}

\subsection{Challenges}
\label{sec:challenges}

This section delves into discussing three distinct challenges associated with offline reinforcement unlearning.

\begin{itemize}[leftmargin=*]
    \item \textbf{How can we evaluate the efficacy of offline reinforcement unlearning? } 
    We focus on approximate unlearning, and should essentially assess whether a trajectory is part of the agent's training dataset. One natural approach is to leverage membership inference attacks (MIAs), and there have been MIAs against DRL~\cite{pan2019you,gomrokchi2022membership,ye2023RLunlearning}. While most of these are aimed at online RL, Du et al.~\cite{du2023orl} introduced \orlauditor to audit datasets for offline DRL models. However, their process is time-consuming due to the extensive training of numerous shadow agents. A simple yet effective method for evaluating approximate offline reinforcement unlearning is the fundamental basis of our study. 
    \item \textbf{How can we unlearn trajectories from the agent’s policy?} The objective of offline reinforcement unlearning is to eliminate the trajectories' impact on the agent, essentially to ``forget trajectories.'' Currently, there is a lack of established methods for unlearning trajectories in the field of offline RL. Therefore, the primary challenge lies in devising an effective unlearning methodology tailored for offline RL.
    \item \textbf{How can we prevent degradation of performance after unlearning?} As the training of RL especially suffering from unstable training~\cite{andrychowicz2021what,d4rl}, the unlearning process can result in performance degradation. Ensuring that the unlearned agent retains its effectiveness poses a significant challenge.
\end{itemize}
\begin{figure*}[!t]
    \centering
    \setlength{\abovecaptionskip}{0pt}
    \includegraphics[width=1.0\linewidth]{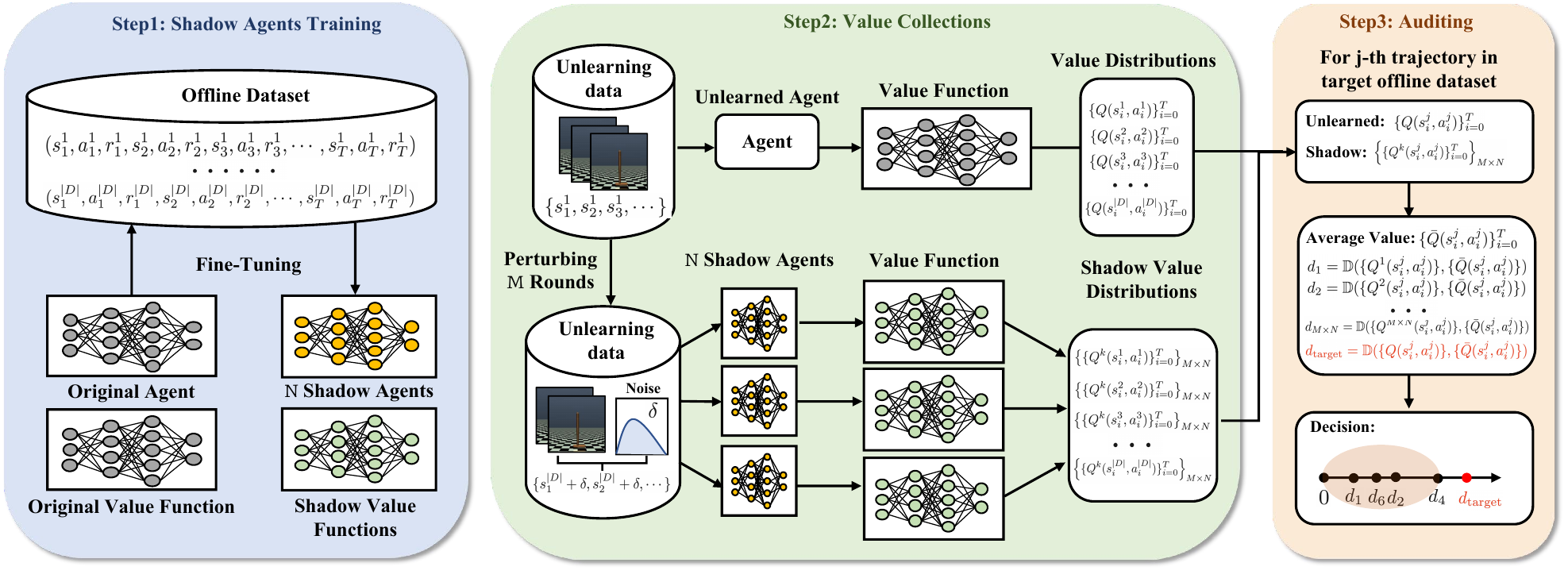}
    \caption{{The workflow of \toolnameeval. ``Shadow Agents Training'' fine-tunes the original target agent under investigation, and its value function, to gather $N$ shadow agents and shadow value functions. ``Value Collections'' calculates the value (as described in Eq. (\ref{eq:value_func})) distributions of the states of target trajectories from the unlearning agents, i.e., $\left\{Q\left(s_i^{|D|},a_i^{|D|}\right)\right\}_{i=0}^T$, where $T$ is the length of trajectories. This step introduces noise, denoted as $\delta$, to the states $M$ times, calculating the value distributions of perturbed trajectories via shadow models. Thus, for each trajectory, we obtain $M \times N$ value distributions. The ``Auditing'' assesses the distances between the average value distribution and the value distributions of both the unlearned agent $\left(d_{\text{target}}\right)$ and the shadow agents $\left(d_{M \times N}\right)$. These similarities determine whether trajectories belong to the training dataset.}}
    \vspace{-5mm}
    \label{fig:auditor}
\end{figure*}

\subsection{Our Proposals}
To address the first challenge, we propose \toolnameeval. Contrasting with the approach in \orlauditor\cite{du2023orl}, which involves training shadow agents from scratch, \toolnameeval adopts a more direct method by fine-tuning the original agent to generate the shadow agents. Besides, we introduce perturbations to the states within the trajectories to generate diverse bases for auditing. These two processes significantly reduce the time required for auditing.

To overcome the second challenge, referring to the proposed definition of ``forgetting an environment'' in Ye et al.~\cite{ye2023RLunlearning}, we interpret ``forgetting trajectories'' as the agent demonstrating reduced or deteriorating performance when encountering states involved in those trajectories. This interpretation is in line with intuitive understanding. When an agent has learned a trajectory and is familiar with its states, it can perform more effectively upon encountering similar states. This results in improved performance, as the agent is better equipped to execute optimal actions in these familiar states. 
Alternatively, when the agent encounters unfamiliar states from forgotten trajectories, it tends to make sub-optimal decisions. 


To address the third challenge, we suggest fine-tuning the unlearned agent to minimize the disparity between the value functions of the unlearned and the original agents. Specifically, this approach aims to synchronize the value function of state-action pairs in the remaining offline dataset.

\section{\toolnameeval}
\label{sub:efficacy}


\subsection{Existing Solutions}
The foundational metric of our study is to determine if a trajectory is included in the agent's training dataset, which is crucial for understanding approximate unlearning. Being the first to introduce offline reinforcement unlearning, we recognize the absence of a definitive method to quantify this. In another field of approximate unlearning, MIAs are widely used for unlearning methods' evaluation~\cite{unlearning_survey,zhang2024contrastive}. This motivates us to consider using MIAs to evaluate offline reinforcement learning as well. In supervised learning, loss is an effective metric for evaluating the effectiveness of unlearning~\cite{shokri2017mia,lira}. However, as our experiments detailed in Appendix~\ref{appsub:td_auditor}, the loss difference between trained and non-trained trajectories is minimal, rendering it unsuitable as a basis for auditing. MIAs in online RL~\cite{pan2019you,gomrokchi2022membership,ye2023RLunlearning} assume
that attackers have control over environments, enabling them to gather trajectories and manipulate them. However, in offline RL, such operations are impractical as we cannot access the environments. 
Du et al.~\cite{du2023orl} emphasize the same perspective that MIA applicable in online RL is not applicable to offline RL. 

Du et al.~\cite{du2023orl} introduced \orlauditor for auditing datasets in offline RL models. This approach leverages the concept that \textit{cumulative rewards} can serve as a unique identifier. In particular, $Q^\pi$ measures the expected cumulative reward that an agent can attain, beginning from state $s$, executing action $a$, and adhering to policy $\pi$. 
We consider the series of states $\left\{(s_t)_t^T\right\}$ within a trajectory. 
Feeding this series of states into a tested agent $\pi$ generates a sequence of state-action pairs $\left\{(s_t, \pi(s_t))_t^T\right\}$. Subsequently, by evaluating this sequence of state-action pairs with $Q^\pi$, we can derive a \textit{value vector} $\vec{Q}^{\pi} = \left\{Q^\pi(s_t, \pi(s_t))_t^T\right\}$ for a trajectory and the agent. We assemble a collection of shadow agents with the assurance that their training datasets incorporate the target trajectories. The basis for our audit relies on the similarity between the value vector of the target trajectories generated by these shadow agents and that generated by the tested agent. 


\subsection{Our Method}
Different from MIAs, in evaluation, we can involve the target model before and after unlearning. 
Therefore, we do not need to train the shadow models from scratch. It suffices to fine-tune the target original model to obtain the shadow models. It is noticed that \toolnameeval is not a type of MIA. In \toolnameeval, we should involve a model that we confirm includes the target trajectory in its training dataset, which is unpractical in MIAs.
Compared to training many shadow agents from scratch, fine-tuning significantly saves computational overheads. Moreover, to ensure a robust estimation from shadow models, we use state perturbation to ensure shadow models cover a wide range of possibilities. 

We call our method \toolnameeval and plot its main steps in Figure~\ref{fig:auditor}. 
We elaborate on the technical details of \toolnameeval as follows.

\noindent \textbf{Shadow Model Preparation.} We can define that the unlearning dataset $\mathcal{D}_f$ is included in the training dataset of the original agent $\pi$. We then directly fine-tune the original agent $\pi$ on the entire dataset $\mathcal{D}$ to gather a set of shadow agents $\{\pi_i^s\}_{i=0}^N$, where $N$ is the number of shadow agents. \textit{We believe these shadow agents have been thoroughly trained on the unlearning dataset $\mathcal{D}_f$.} Besides, as outlined in Section~\ref{sec:offlineRL}, the fine-tuning process also necessitates concurrently updating the value function, allowing us to obtain the value functions of the shadow agents $\left\{Q^{\pi_i^s}\right\}_{i=0}^N$.

\noindent \textbf{Value Collections.} For $j_{\text{th}}$ $(j=1,\cdots,|\mathcal{D}_f|)$ trajectory in the unlearning dataset $\mathcal{D}_f$, we query the unlearned agent's value vector $\vec{Q}^{\pi'}_j = Q^{\pi'}\left(s^j_t, \pi'\left(s^j_t\right)\right), (t=1,\cdots,T)$. Additionally, for the $j_{\text{th}}$ trajectory, we collect the value vectors $\vec{Q}^{\pi_i^s}_j = Q^{\pi_i^s}\left(s^j_t, \pi_i^s\left(s^j_t\right)\right), (t=1,\cdots,T; \ i=1,\cdots,N)$ generated by shadow agents. Besides, we add a low-level noise $\delta$ to perturb states to generate more value vectors for a trajectory, further reducing the requirement for extensive shadow agent training. By using perturbations across $M$ rounds $\left\{(s_t^j + \delta_h)_{h=0}^M\right\}$ and utilizing $N$ shadow agents, we gather $M \times N$ value vectors for each trajectory within the unlearning dataset, which is recorded as $\vec{Q}^{\pi_i^s}_{j,m} (i=1,\cdots, N; \ j=1,\cdots, M)$.

\noindent \textbf{Auditor.}  This step relies on the similarity between the value vector of the unlearning trajectories generated by shadow agents and that generated by the unlearned agent. Specifically, for $j_{\text{th}}$  trajectory, we define the mean of $\left\{\vec{Q}^{\pi_i^s}_{j,m}\right\}$ as $\vec{Q}^{\pi^s}_{j,\text{ave}}$. Referring to the practical adoption in \orlauditor~\cite{du2023orl}, we conduct Grubbs' hypothesis test~\cite{grubbs1950sample} to ascertain the success of removal. This determination is made when $d_{\text{target}} = \mathbb{D}\left( \vec{Q}^{\pi^s}_{j,\text{ave}}, \vec{Q}^{\pi'}_j\right)$
falls outside the distribution of $\left.\left\{\mathbb{D}\left( \vec{Q}^{\pi^s}_{j,\text{ave}}, \vec{Q}^{\pi_i^s}_{j,m}\right) \right| i \in \{1,\cdots, N\}, j \in \{1,\cdots, M\}\right\}$, where $\mathbb{D}$ denotes the Wasserstein distance~\cite{wasserstein} between two vectors. Otherwise, \toolnameeval determines that the trajectories, which are required to be forgotten, continue to influence the unlearned agent $\pi'$. 
\section{\toolname}
\label{sec:methodology}


\subsection{Strawman Unlearning Methods}
\label{subsec:strawman}

{Ye et al. proposed ``environment unlearning'' in online RL. We cannot directly apply reinforcement environment unlearning for online RL~\cite{ye2023RLunlearning} because it requires the user to poison the transition function 
$\mathcal{T}(\cdot|s_t,a_t)$ of the environment (the training for offline RL is restricted from accessing environments). 
}

{Another baseline is to perturb the rewards (it has been understood that agents learn via rewards~\cite{sutton2018reinforcement}; thus, perturbing rewards prevents agents from learning the corresponding behaviors).
We modify the rewards of unlearned trajectories by assigning random rewards sampled from the uniform distribution.  Then, we fine-tune the agent using this modified dataset for unlearning. This approach constitutes the ``random-reward'' baseline in Section~\ref{sub:baseline}. 
Moreover, we can also assign unlearned trajectories with small or the worst possible rewards, but they could adversely affect the agent's performance. 

To overcome this issue, {we propose to enhance the agent's performance on the remaining dataset in the unlearning stage, which is crucial to prevent the degradation of the agents' performance. }
The unlearning approach now seems to be successful. However, due to the unstable training in RL~\cite{d4rl,cql,bcq}, the methods mentioned may be susceptible to the issue of policy divergence {(as the empirical analysis presented in Section~\ref{subsec:ablation})}. Therefore, we need to fine-tune the agents on the remaining dataset to encourage policy convergence.

\subsection{Overview}
As discussed in Section~\ref{subsec:notion}, we aim to update the optimal unlearned policy to ensure the agent that \textit{takes actions of lower value in unlearning trajectories while still preserving its performance in the rest of the trajectories}. To achieve this goal, we structure \toolname into two distinct components that align with the first and second terms of Eq.~(\ref{eq:unlearning_target}) as follows: (1) \textit{Forgetting}, and (2) \textit{Convergence Training}.
\begin{equation}
\left\{
\begin{aligned}
    & \min \mathbb{E}_{s\sim \mathcal{D}_f}\left[Q^{\pi'}(s,\pi'(s))\right] + \max \mathbb{E}_{s\sim \mathcal{D}_m}\left[Q^{\pi'}(s,\pi'(s))\right] \\ 
    & \min \mathbb{E}_{(s,a) \sim \mathcal{D}_m}\left[ \left\|Q^{\pi'}(s,a) - Q^\pi(s,a) \right\|_{\infty}\right].
    \label{eq:unlearning_target}
\end{aligned} \right.
\end{equation}

The first term of Eq.~(\ref{eq:unlearning_target}) represents the `forgetting' phase, instructing the learned policy $\pi'$ to take suboptimal actions in the unlearning dataset $\mathcal{D}_f$, while maintaining its normal behavior in the remaining dataset $\mathcal{D}_m.$ We achieve this by updating the agent to deliberately minimize its cumulative reward in the states belonging to the unlearning trajectories, thereby steering it towards making less effective decisions in those states. Consequently, we minimize the value function for states within the unlearning trajectories, aligning the agent's actions with the unlearning objectives. Training an agent by solely minimizing the value function could lead to the issue of agent collapse -- the agent's performance tends to decline rapidly when interacting with the environment~\cite{gong2022mind}. To mitigate this problem, we also focus on maximizing the agent's value for states in the remaining dataset, thereby balancing the unlearning process and preventing the agent from deteriorating.

The second term of Eq.~(\ref{eq:unlearning_target}), aligning with the ``convergence training'' phase, aims to reduce the value differences between the original and the unlearned agent on the remaining dataset $\mathcal{D}_m$. 
This step ensures we can fine-tune the unlearned agent's convergence, as discussed in Section~\ref{secsub:convergence}. 

\subsection{Technical Details}
\label{subsec:trajdeleter_detail}

This section describes how \toolname enables offline RL agents to forget specific trajectories. We outline the processes for ``forgetting'' and ``convergence training,'' as follows.

\noindent \textbf{Forgetting.} This learning phase trains the agent to forget specific trajectories while maintaining effectiveness in other trajectories. We initially reformulate the objective function as,
$$\mathcal{L}_1 = \max \mathbb{E}_{s\sim \mathcal{D}_m} \left[Q^{\pi'}(s,\pi'(s))\right] - \mathbb{E}_{s\sim \mathcal{D}_f}\left[Q^{\pi'}(s,\pi'(s))\right].$$
Our unlearning process begins with the original policy $\pi$ (the agent required being unlearned). We use a neural network represented by $\pi'_\theta$ to denote the policy after unlearning. Our objective is to optimize the parameter $\theta$ to maximize the value function, which is given by, 
\begin{equation}
\label{eq:target_forgetting}
    \mathbb{E}_{s\sim \mathcal{D}_m, a\sim \pi'_\theta}\left[Q^{\pi'_\theta}(s,a)\right] - \lambda \mathbb{E}_{s\sim \mathcal{D}_f,a\sim \pi'_\theta}\left[Q^{\pi'_\theta}(s,a)\right],
\end{equation}
where $\lambda$ is a constant to balance the
unlearning process and prevent the agent from deteriorating. We then compute the gradient of the objective function with respect to the parameters and
iteratively apply stochastic gradient-ascend to approach a local maximum in $\mathcal{L}_1(\theta)$. Specifically, at iteration $k$, we update the policy by gradient ascent, $\theta_{k+1} \gets \theta_k + \nabla_\theta \mathcal{L}_1(\theta)|_{\theta_k}$. Based on the Policy Gradient Theorem~\cite{a3c}, we express the policy gradient of policy $\pi_\theta$ as,
\begin{equation}
\label{eq:forgetting_policy_update}
\begin{aligned}
    & \nabla_{\theta_k} \mathcal{L}_1 = \mathbb{E}_{ s \sim \mathcal{D}_m, a\sim \pi'_{\theta_k}}\left[\nabla_{\theta_k} \log \pi'_{\theta_k} (s,a)A_{\pi'_{\theta_k}}(s,a) \right] \\
    & \qquad \ \ - \lambda \mathbb{E}_{ s \sim \mathcal{D}_f, a\sim \pi'_{\theta_k}}\left[\nabla_{\theta_k} \log \pi'_{\theta_k} (s,a)A_{\pi'_{\theta_k}}(s,a) \right].
\end{aligned}
\end{equation}
Here, $A_\pi(s, a)$ indicates the advantage function, measuring the difference between the value of state-action pair $(s, a)$ and the average value of that state $s$. Using this advantage function, we can determine the improvement of taking a particular action in a given state over the average. In Eq. (\ref{eq:forgetting_policy_update}), the advantage function $A_{\pi'_{\theta_k}}(s,a)$ is defined as, $
    A_{\pi'_{\theta_k}}(s,a) = Q^{\pi'_{\theta_k}} (s,a) - \mathbb{E}_{a\sim \pi'_{\theta_k}} \left[ Q^{\pi'_{\theta_k}} (s,a) \right]$. 
Essentially, this function computes the additional reward the agent receives from choosing that action~\cite{advantage_function}. 

To address the Eq. (\ref{eq:forgetting_policy_update}), we are required to approximate function $Q^{\pi'_{\theta_k}}$ for each iteration. We start this approximation by an neural network parameterized with $\phi_k$ to model $Q^{\pi'_{\theta_k}}(s,a)$ using the TD-learning paradigm~\cite{sutton2018reinforcement}, and then update this network for next state $s'$ repeatedly by minimizing the TD-error $\mathcal{G}(\phi_k)$, 
\begin{equation}
\label{eq:forgetting_td}
 \mathbb{E}_{\mathcal{D}} \left\| Q^{\pi'_{\theta_k}}_{\phi_k}(s,a) - \left( r + \gamma \mathbb{E}_{a'\sim \pi_{\theta_k}} \left[  Q^{\pi'_{\theta_k}}_{\phi_k}(s',a') \right] \right) \right\|_2,
\end{equation}
where $(s,a,r,s') \sim \mathcal{D}$, and $\mathcal{D} = \mathcal{D}_m \cup \mathcal{D}_f$ represents the original dataset. By consistently updating the policy network and value function using the method mentioned above, the agent is effectively trained to ``forget'' the targeted trajectories.

\noindent \textbf{Convergence training.} Relying solely on this training might not ensure the convergence of $\pi'$, potentially causing instability (as conducted experiments in Section~\ref{subsec:ablation}) in the unlearning process. To address this issue, we introduce ``convergence training'' in \toolname, as depicted in the second term of Eq.~(\ref{eq:unlearning_target}).
The stage focuses on minimizing the difference between the value functions of the unlearned policy $\pi'$ and the fixed original policy $\pi$, aiming to align the value function of $\pi'_\theta$ closely with that of $\pi$ and providing the convergence guarantee for \toolname. 

This ``forgetting'' phase yields a trained policy that serves as the ``convergence training'' starting point. At iteration $h$, we initially fine-tune the value function by,
\begin{equation}
    \mathcal{L}_2(\theta) = \min \mathbb{E}_{(s,a) \sim \mathcal{D}_m}\left[ \left\|Q^{\pi'_{\theta_h}}(s,a) - Q^\pi(s,a) \right\|_{2}\right].
    \label{eq:convergence_td}
\end{equation}
Then, analogous to the implication in Eq.~(\ref{eq:forgetting_policy_update}), we update the policy $\pi'_{\theta_h}$ based on the Policy Gradient Theorem~\cite{a3c},
\begin{equation}  
    \nabla_{\theta_h} \mathcal{L}_2 = \mathbb{E}_{ s \sim \mathcal{D}_m, a\sim \pi'_{\theta_h}}\left[\nabla_{\theta_h} \log \pi'_{\theta_h} (s,a)A_{\pi'_{\theta_h}} \right].
    \label{eq:adv_convergence}
\end{equation}
After training, \toolname fine-tunes the policy for convergence. We provide the theoretical analysis in Section~\ref{secsub:convergence}.

\noindent \textbf{Summary.}
We outline \toolname in Algorithm~\ref{alg:trajdeleter}. This algorithm inputs the original agent and its value function $Q^\pi$ and the dataset targeted for unlearning $\mathcal{D}_f$. The workflow involves a two-phase approach: first ``forgetting'' specific trajectories, then reinforcing the new policy through convergence training. In the forgetting phase, it iteratively processes batches of trajectories from both $\mathcal{D}_m$ and $\mathcal{D}_f$ (Line~\ref{line:1}). The computation of advantages for each trajectory in these datasets (Line~\ref{line:5}) involves updating the unlearning policy $\pi'_\theta$ (Line~\ref{line:8}) and the value functions (Line~\ref{line:9}) to gradually forget behaviors learned from $\mathcal{D}_f$. This phase is key in systematically erasing specific behaviors from the agent's learning.

Following this, we start the convergence training phase. Here, trajectories from $\mathcal{D}_m$ are sampled (Line~\ref{line:12}). The advantage for each trajectory is computed similarly to the forgetting phase (Line~\ref{line:13}). Then, updates are made to the policy and value function parameters to reinforce the unlearned agent (Line~\ref{line:14}-\ref{line:15}). This phase ensures that the agent's behavior post-unlearning remains consistent and effective. Upon completing the iterative process described, \toolname produces a well-trained unlearned agent $\pi'_\theta$.

\subsection{Convergence Analysis}
\label{secsub:convergence}

The ``forgetting'' phase yields a trained policy that serves as the start of the next stage. Then, the policy conducts fine-tuning guided by the second term of Eq.~(\ref{eq:unlearning_target}). We provide theoretical analysis to guarantee policy convergence after fine-tuning.

We assume that the original policy $\pi$ remains fixed during training and approximate the optimal policy $\pi^\ast$, focusing solely on training the policy $\pi'$ for unlearning. Theorem~\ref{theorem:policy_convergence} states that as training progresses, the difference between the learned $Q$-function and the optimal $Q$-function diminishes.

\begin{theorem}[Interaction convergence~\cite{TosattoPDR17}]\label{theorem:policy_convergence} We assume that the offline dataset includes a diverse range of states. The state distribution generated by any policy is consistently bounded relative to the distribution in the offline dataset. Specifically, denoting the state distribution of the offline dataset as $\mu(s)$, for the state distribution $\nu(s)$ generated by any policy ${\pi_k}$, the condition $\forall s, \frac{\nu(s)}{\mu(s)}\leq C$ holds. Let $Q^\ast$ indicate the optimal value function; we have,
    $$\|Q^\ast - Q^{\pi_{k+1}}\|_\infty \leq \gamma \|Q^\ast - Q^{\pi_k}\|_\infty + \epsilon + C\|Q^{\pi_k}\|_\infty,$$
    where ${\pi_k}$ denotes a sequence of policies correlated to their respective value functions ${Q^{\pi_k} }$. Here, $\epsilon$ signifies the approximation error in value estimation: $ \| Q^{\pi_{k}} - \left(r + \gamma Q^{\pi_{k+1}}\right) \|_\infty$.
\end{theorem}

\begin{algorithm}[!t]
    \caption{Workflow of \toolname}
    \label{alg:trajdeleter}
    \textbf{Input}: \textbf{$\pi$}: The original agent, parameterized by $\theta_o$. $\pi'_\theta$: The unlearning policy, parameterized by $\theta$.
    Value Functions: $Q^\pi$ and $Q^{\pi'}$, with parameters $\phi_o$ and $\phi$, respectively.
    $\mathcal{D}$: The complete offline dataset.
    $\mathcal{D}_f$: The dataset the agent needs to forget.
    $\mathcal{D}_m$: The remaining dataset post unlearning. \\
        \textbf{Initialization}: initialize $\theta^{(0)} = \theta_o$, $\phi^{(0)} = \phi_o$. \\
        \tcp{\color{blue} Forgetting}
        \For{$k = 1, 2, 3, \cdots, K$ }{
            Sample trajectories from $\mathcal{D}_m$ and $\mathcal{D}_f$, and collect a batch of trajectories $\mathcal{D}_{Bm} = \{\tau_i\}$, $\mathcal{D}_{Bf} = \{\tau_j\}$, where $i,j=1,2,\cdots,B$. \label{line:1}\\
            For each trajectory $\tau_i$ and $\tau_j$, compute the advantage at each time step $t$ of trajectories: 
            $A_{\pi'}^{i_t}(s^i_t,a^i_t) = Q^{\pi'} \left(s^i_t,a^i_t\right) - \mathbb{E}_{a\sim \pi'} \left[ Q^{\pi'} \left(s^i_t,a\right) \right],$
            $A_{\pi'}^{j_t}(s^j_t,a^j_t) = Q^{\pi'} \left(s^j_t,a^j_t\right) - \mathbb{E}_{\pi'} \left[ Q^{\pi'} \left(s^j_t,a\right) \right].$ \label{line:5}\\
            We have $A_{\pi'}^{i_{0:|\tau_i|}}$ and $A_{\pi'}^{j_{0:|\tau_j|}}$ $(i,j = 1,\cdots,B)$ in Eq. (\ref{eq:forgetting_policy_update}). Then, we can obtain the policy gradient to maximize the Eq. (\ref{eq:target_forgetting}) by updating $\theta^{(k)}$. \label{line:8} \\
            Update $\phi^{(k)}$ by minimizing the loss of Eq. (\ref{eq:forgetting_td}). \label{line:9}\\
         }
         \tcp{\color{blue} Convergence training}
         \For{$h = 1, 2, 3, \cdots, H$ }{
            Sample trajectories from $\mathcal{D}_m$, and collect a batch of trajectories $\mathcal{D}_B = \{\tau_i\}$, where $i=1,\cdots,B$. \label{line:12}\\
            For each trajectory $\tau_i$, compute the advantage at each time step $t$ of trajectories: 
            $A_{\pi'}^{i_t}(s^i_t,a^i_t) = Q^{\pi'} \left(s^i_t,a^i_t\right) - \mathbb{E}_{a\sim \pi'} \left[ Q^{\pi'} \left(s^i_t,a\right) \right].$ \label{line:13}\\
            We have $A_{\pi'}^{i_{0:|\tau_i|}}$ $(i = 1,\cdots,B)$ in Eq. (\ref{eq:adv_convergence}) and can obtain the policy gradient to update $\theta^{(K+h)}$. \label{line:14}\\
            Update $\phi^{(K + h)}$ by minimizing the loss of Eq. (\ref{eq:convergence_td}) and Eq. (\ref{eq:forgetting_td}) on the $\mathcal{D}_{Bm}$. \label{line:15} \\
         }
         \textbf{Output}: $\pi_{\theta}'$: the well-trained unlearned agent. \\
    \end{algorithm}

This formulation implies the convergence of value function $Q$ towards the optimal value function with a sufficiently large number of learning iterations under certain conditions. Specifically, when initiating the optimization of the second term from any starting policy derived by the first term of Eq.~(\ref{eq:unlearning_target}), we maintain a consistent boundary between the optimal policy $\pi^\ast$ and the learned policy $\pi_k$~\cite{TosattoPDR17},
$
\|Q^{\pi_k} - Q^\ast\|_\infty \leq \frac{1-\gamma^{k}}{1-\gamma} \sqrt{2A C \epsilon}+\gamma^{k} \frac{R_{\max }}{1-\gamma}.
$
$R_{\text{max}}$ represent the maximum value of reward function. $A$ is the size of possible actions within the action space. The bound between performance of the optimal policy $\pi^\ast$ and the learned policy $\pi_{k}$ is,
$
    \left\| \mathcal{L}(\pi_{k}) - \mathcal{L}(\pi^\ast) \right\|_\infty \leq \frac{2}{1-\gamma} \left( \frac{1-\gamma^{k}}{1-\gamma} \sqrt{2 A C \epsilon}+\gamma^{k} \frac{R_{\max }}{1-\gamma} \right).
$
Therefore, after executing a sufficient learning process, \toolname effectively fine-tunes the policy to achieve convergence.

\section{Experimental Setup}
\label{sec:setup}

\subsection{Investigated Tasks and Datasets}


We conduct experiments on three widely robotic control tasks, {\tt Hopper}, {\tt Half-Cheetah}, and {\tt Walker2D} from MuJoCo~\cite{gym}, which are all commonly used in previous studies~\cite{du2023orl}.
In the {\tt Hopper} task, the objective for the agent is to maneuver a one-legged robot to move forward at the highest possible speed. Moreover, in the {\tt Half-Cheetah} and {\tt Walker2D} tasks, the agent is tasked with controlling a cheetah and a bipedal robot respectively, to walk forward as fast as possible. We utilized offline datasets from D4RL, a benchmark designed for offline RL algorithm evaluation. D4RL offers diverse datasets for these tasks, including \textit{medium}, \textit{random}, \textit{medium-replay}, and \textit{medium-expert}, each collected using different policies. In our experiments, we chose the \textit{medium-expert}, an agent trained to achieve the highest performance compared to others using different datasets. Further details on tasks and chosen datasets are available in Appendix~\ref{app:task_dataset}. 

\subsection{Baselines} 
\label{sub:baseline}


\noindent \textbf{Retraining from Scratch (Reference). } This baseline involves retraining the agent from scratch. Retraining the agent is applicable when the original training data is accessible and ensures complete trajectory removal. Generally, this method offers a precise guarantee for unlearning specific trajectories, but it is resource-intensive. {
This method acts as a reference for evaluating other unlearning methods.}

\begin{table*}[!t]
\small
    \centering
    \caption{Precision, recall, and F1-score of \toolnameeval. We evaluate the performance at different unlearning rates using exact unlearning methods, i.e., retraining from scratch (reference).}
    \resizebox{0.98\textwidth}{!}{
   \begin{tabular}{p{1.8cm}|c|ccc|ccc|ccc|ccc}
        \toprule
         \multirow{3}{*}{\textbf{Tasks}}   &  \multirow{3}{*}{\textbf{Algorithms}}  &\multicolumn{12}{c}{\text{\textbf{Unlearning rates}}}   \\
         \cline{3-14}
        & & \multicolumn{3}{c|}{\text{\textbf{0.01}}} & \multicolumn{3}{c|}{\text{\textbf{0.05}}} & \multicolumn{3}{c|}{\text{\textbf{0.1}}} & \multicolumn{3}{c}{\text{\textbf{0.15}}}   \\
         \cline{3-14}
         & & \textbf{Precision}  & \textbf{Recall}  & \textbf{F1} & \textbf{Precision}  & \textbf{Recall}  & \textbf{F1} & \textbf{Precision}  & \textbf{Recall}  & \textbf{F1} & \textbf{Precision}  & \textbf{Recall}  & \textbf{F1} \\
         \hline
         \multirow{6}{*}{{\tt Hopper}} 
         & BEAR & 100.0  $\%$ & 85.43 $\%$ & 0.92 & 99.68 $\%$ & 84.37 $\%$ & 0.91 & 100.0 $\%$ & 85.15 $\%$ & 0.92 & 99.75 $\%$ & 84.77 $\%$ & 0.92 \\
         & BCQ & 100.0 $\%$ & 81.79 $\%$ & 0.90 & 100.0 $\%$ & 80.14 $\%$ & 0.89 & 99.77 $\%$ & 80.75 $\%$ & 0.89 & 98.06 $\%$ & 80.01 $\%$ & 0.88 \\
         & CQL & 100.0 $\%$ & 84.70 $\%$ & 0.92 & 100.0 $\%$ & 84.92 $\%$ & 0.92 & 100.0 $\%$ & 84.46 $\%$ &  0.92 & 100.0 $\%$ & 85.57 $\%$ & 0.92 \\
         & IQL & 99.75 $\%$ & 83.25 $\%$ & 0.91 & 98.98 $\%$ & 84.06 $\%$ & 0.91 & 94.18 $\%$ & 84.54 $\%$ & 0.91 & 95.79 $\%$ & 84.77 $\%$ & 0.90 \\
         & PLAS-P & 100.0 $\%$ & 54.01 $\%$ & 0.71 & 100.0 $\%$ & 71.25 $\%$ & 0.85 & 98.69 $\%$ & 78.46 $\%$ & 0.87 & 88.24 $\%$ & 81.45 $\%$ & 0.83\\
         & TD3PlusBC & 99.03 $\%$ & 85.84 $\%$ & 0.92 & 98.30 $\%$ & 83.28 $\%$ & 0.90 & 96.00 $\%$ & 83.93 $\%$ & 0.90 & 93.33 $\%$ & 84.69 $\%$ & 0.88\\
         \cline{2-14}
         & \cellcolor{gray!20}\textbf{Average} & \cellcolor{gray!20}\textbf{99.80 $\%$} & \cellcolor{gray!20}\textbf{79.17 $\%$} & \cellcolor{gray!20}\textbf{0.88} & \cellcolor{gray!20}\textbf{99.49 $\%$} & \cellcolor{gray!20}\textbf{81.34 $\%$} & \cellcolor{gray!20}\textbf{0.90} & \cellcolor{gray!20}\textbf{97.77 $\%$} & \cellcolor{gray!20}\textbf{82.21 $\%$} & \cellcolor{gray!20}\textbf{0.90} & \cellcolor{gray!20}\textbf{95.86 $\%$} & \cellcolor{gray!20}\textbf{83.54$\%$} & \cellcolor{gray!20}\textbf{0.89} \\
          \toprule
         \multirow{6}{2em}{\shortstack{{\tt Half-} \\ {\tt Cheetah}}}
         & BEAR & 100.0 $\%$ & 83.18 $\%$ & 0.91 & 99.68 $\%$ & 84.37 $\%$ & 0.91 & 100.0 $\%$ & 84.31 $\%$ & 0.91 & 100.0 $\%$ & 82.50 $\%$ & 0.91  \\
         & BCQ & 100.0 $\%$ & 80.54 $\%$ & 0.88 & 100.0 $\%$ & 75.89 $\%$ & 0.86 & 99.75 $\%$ & 76.89 $\%$ & 0.87 & 99.25 $\%$ & 80.41 $\%$ & 0.89 \\
         & CQL & 100.0 $\%$ & 83.00 $\%$ & 0.91 & 100.0 $\%$ & 77.50 $\%$ & 0.87 & 100.0 $\%$ & 80.00 $\%$ & 0.88 & 99.72 $\%$ & 81.42 $\%$ & 0.89 \\
         & IQL & 100.0 $\%$ & 77.65 $\%$ & 0.87 & 100.0 $\%$ & 79.45 $\%$ & 0.88 & 100.0 $\%$ & 81.72 $\%$ & 0.90 & 100.0 $\%$ & 80.72 $\%$ & 0.89 \\
         & PLAS-P & 100.0 $\%$ & 62.55 $\%$ & 0.76 & 100.0 $\%$ & 72.58  $\%$ & 0.84 & 100.0 $\%$ & 65.47 $\%$ & 0.78 & 100.0 $\%$ & 62.04 $\%$ & 0.76 \\
         & TD3PlusBC & 99.45 $\%$ & 81.13 $\%$ & 0.89 & 100.0 $\%$ & 80.21 $\%$ & 0.89 & 100.0 $\%$ & 80.08 $\%$ & 0.89 & 100.0 $\%$ & 81.50 $\%$ & 0.90 \\
         \cline{2-14}
         & \cellcolor{gray!20}\textbf{Average} & \cellcolor{gray!20}\textbf{99.91 $\%$} & \cellcolor{gray!20}\textbf{77.84 $\%$} & \cellcolor{gray!20}\textbf{0.87} & \cellcolor{gray!20}\textbf{99.95 $\%$} & \cellcolor{gray!20}\textbf{78.33 $\%$} & \cellcolor{gray!20}\textbf{0.88} & \cellcolor{gray!20}\textbf{99.83 $\%$} & \cellcolor{gray!20}\textbf{79.76 $\%$} & \cellcolor{gray!20}\textbf{0.88} & \cellcolor{gray!20}\textbf{98.31 $\%$} & \cellcolor{gray!20}\textbf{78.73 $\%$} & \cellcolor{gray!20}\textbf{0.88} \\
         \toprule
         \multirow{6}{2em}{{\tt Walker2D}}
         & BEAR & 99.73 $\%$ & 82.95 $\%$ & 0.91 & 99.61 $\%$ & 81.32 $\%$ & 0.90 & 100.0 $\%$ & 83.54 $\%$ & 0.91 & 99.19 $\%$ & 83.33 $\%$ & 0.91 \\
         & BCQ & 100.0 $\%$ & 78.25 $\%$ & 0.88 & 99.84 $\%$ & 76.85 $\%$ & 0.87 & 99.32 $\%$ & 78.36 $\%$ & 0.86 & 98.88 $\%$ & 76.85 $\%$ & 0.86 \\
         & CQL & 100.0 $\%$ & 81.99 $\%$ & 0.89 &  100.0 $\%$ & 85.45 $\%$ & 0.91 & 99.80 $\%$ & 82.55 $\%$ & 0.90 & 99.35 $\%$ & 82.27 $\%$ & 0.90 \\
         & IQL & 99.75 $\%$ & 83.25 $\%$ & 0.91 & 98.15 $\%$ & 84.42 $\%$ & 0.90 & 97.32 $\%$ & 74.75 $\%$ & 0.85 & 99.04 $\%$ & 73.71 $\%$ & 0.84 \\
         & PLAS-P & 99.77 $\%$ & 80.18 $\%$ & 0.89 & 99.67 $\%$ & 78.74 $\%$ & 0.88 & 100.0 $\%$ & 79.95 $\%$ & 0.88 & 99.28 $\%$ & 76.89 $\%$ & 0.87 \\
         & TD3PlusBC & 99.76 $\%$ & 84.68 $\%$ & 0.92 & 99.15 $\%$ & 83.78 $\%$ & 0.90 & 99.65 $\%$ & 83.77 $\%$ & 0.91 & 99.26 $\%$ & 85.26 $\%$ & 0.92\\
         \cline{2-14}
         & \cellcolor{gray!20}\textbf{Average} & \cellcolor{gray!20}\textbf{99.84 $\%$} & \cellcolor{gray!20}\textbf{81.88 $\%$} & \cellcolor{gray!20}\textbf{0.90} & \cellcolor{gray!20}\textbf{99.57 $\%$} & \cellcolor{gray!20}\textbf{81.76 $\%$} & \cellcolor{gray!20}\textbf{0.89} & \cellcolor{gray!20}\textbf{99.00 $\%$} & \cellcolor{gray!20}\textbf{79.72 $\%$} & \cellcolor{gray!20}\textbf{0.88} & \cellcolor{gray!20}\textbf{98.51 $\%$} & \cellcolor{gray!20}\textbf{79.80 $\%$} & \cellcolor{gray!20}\textbf{0.87} \\
        \bottomrule
    \end{tabular}
    }
    \label{tab:trajauditor}
\end{table*}

\noindent \textbf{Fine-tuning. } This baseline extends the training of an agent using the dataset from which the targeted trajectories have been removed. We implement this fine-tuning process to adjust the agent's parameters with a limited number of iterations.

\noindent \textbf{Random-reward. } In RL, an agent's training is fundamentally guided by a reward-based paradigm~\cite{sutton2018reinforcement}. Intuitively, for a specific state, if an agent receives a high reward for an action, it is more likely to choose that action again under similar states in preference to actions associated with lower rewards. This baseline edits the reward in the trajectories selected for unlearning by assigning them random rewards. Then, the original agent is fine-tuned on this modified dataset for the unlearning process. In our experiments, random rewards are generated by sampling from a uniform distribution, where the maximum and minimum values are the highest and lowest rewards observed in the entire offline dataset.

\subsection{Implementation and Experiment Platforms}
\toolname is designed to be agent-agnostic: it should handle unlearning requests for agents trained using various offline RL algorithms. 
We select six offline RL algorithms that are prevalently used in offline RL community~\cite{d3rlpy}. Specifically, we select bootstrapping error accumulation reduction (BEAR)~\cite{kumar2019stabilizing}, batch-constrained deep Q-learning (BCQ)~\cite{bcq}, conservative Q-learning (CQL)~\cite{cql}, implicit Q-learning (IQL)~\cite{iql}, policy in the latent action apace with perturbation (PLAS-P)~\cite{plasp}, and twin delayed deep deterministic policy gradient plus behavioral cloning (TD3PlusBC)~\cite{td3plusbc}. 
We elaborate on the details of investigated algorithms in Appendix~\ref{app:offline_rl_algorithms}. We use the open-source repository for implementation~\cite{d3rlpy}. For more details on implementation and experiment platforms, please refer to Appendix~\ref{app:implement}.

\subsection{Evaluation Metrics}
\label{subsec:metrics}
We introduce metrics used to assess the \toolname from the perspectives mentioned in Appendix~\ref{app:principles}. 

\noindent \textbf{Precision, Recall and F-1 scores.} These metrics are used to assess the effectiveness of trajectory removal, i.e., the \textit{efficacy} of unlearning~\cite{chen2022graph,unlearning_survey}. {We define ``true positives'' (TP) as the trajectories that are actually included in the training dataset of the agent, and ``false negatives'' (FN) refer to the trajectories that are incorrectly marked as part of the training dataset, as identified by \toolnameeval. False negatives (FN) denote those trajectories that, despite being part of the training dataset, are erroneously classified as not included, according to the evaluation by \toolnameeval. }

Precision is the number of $\text{TP}$ over the number of TP plus the number of FP, i.e., $ \frac{\text{TP}}{\text{TP+FP}} \times 100\% $. Besides, recall is defined as the number of TP over the number of TP plus the number of FN, i.e., $ \frac{\text{TP}}{\text{TP+FN}} \times 100\%$. The F1 score, defined as $ \frac{2 \cdot \text{Precision} \cdot \text{Recall} }{ \text{Precision} + \text{Recall}} \times 100\% $, represents the harmonic mean of precision and recall. A higher F1 score denotes a more proficient method for testing trajectory removal. 

\noindent \textbf{Averaged Cumulative Return.} An unlearning method is useful only if it maintains performance levels comparable to the original agent. Hence, we consider the \textit{fidelity}
to be the second measure of performance, and apply \textit{Averaged Cumulative Return} to the quantity of the agent's performance. An agent interacts with the environment, producing a test trajectory denoted by $\tau$.
The cumulative return of this trajectory is defined as $R(\tau) = \sum_{i=0}^{|\tau|} r_i$. We collect a set of test trajectories $\mathcal{T}$. The agent's performance is then quantified by the average of the cumulative returns, i.e., $\frac{1}{|\mathcal{T}|}\sum_{\tau \in \mathcal{T}} R(\tau)$. Consistent with the evaluation in previous works~\cite{d4rl,d3rlpy}, our experiments calculate the average cumulative return over 100 test trajectories. A higher return signifies superior agent performance. 

\noindent \textbf{Time Costs.} This metric measures the \textit{efficiency} of the unlearning method in terms of how long it takes to ``unlearn'' trajectories from an agent. 
A shorter running time incurred by the agent during the unlearning of trajectories is better.


\section{EMPIRICAL EVALUATIONS}
\label{sec:eval}
This section evaluates the effectiveness of \toolname by answering the following three research questions (RQs), 
\begin{itemize}[leftmargin=*]
    \item \textbf{RQ1.} Does the \toolnameeval effectively verify the efficacy of unlearning?
    \item \textbf{RQ2.} How effective is \toolname?
    \item \textbf{RQ3.} How do hyper-parameters affect the performance of \toolname?
\end{itemize}
In the RQ1, we initially determine whether \toolnameeval for offline reinforcement unlearning can identify the presence of a trajectory in the training dataset. Then, in the RQ2, we apply \toolname along with baselines to unlearn specific trajectories across various tasks and offline RL algorithms, to present the effectiveness of \toolname from four perspectives described in Section~\ref{app:principles}. In the RQ3, we investigate the influence of hyper-parameters on the performance of \toolname.

\subsection*{\textbf{RQ1. Does the proposed \toolnameeval effectively verify the efficacy of unlearning?}}

\noindent \textbf{Experiment Design.} We define the \textit{unlearning rate} as the proportion of trajectories that need to be forgotten within the entire dataset (i.e., original dataset). 
We split the entire offline dataset, collected for the mixture of \textit{medium-expert} offline dataset of each task, to the remaining dataset and unlearning dataset across different unlearning rates, i.e., $\{0.01, 0.05, 0.10, 0.15\}$. We ensure that the unlearning and the remaining datasets are collected using the same policy. The original agent is trained on the complete offline dataset, while the unlearned agent is trained from scratch using the remaining dataset with $1 \times 10^6$ timesteps. Thus, the unlearning dataset is definitely excluded from the training dataset of unlearned agents. The number of shadow agents is 5; each fine-tuned for only $5 \times 10^3$ timesteps starting from original agents. We perturb the states of the trajectories over 4 rounds using noise sampled from a Gaussian distribution with a mean of 0 and a standard deviation of 0.05. 

\noindent \textbf{Result Analysis.} Table~\ref{tab:trajauditor} presents the precision, recall, and F1-scores achieved by \toolnameeval across agents, each trained using distinct offline RL algorithms and subjected to the dataset with various unlearning rates. This table shows that \toolnameeval consistently attains high F1-scores of 0.85 across agents we tested, showing its efficacy and robustness. \rev{The average precision across the investigated tasks and unlearning rates stands at 98.9\%, significantly higher than the recall rate of 80.3\%. We derive these results by averaging the precision and recall values presented in the last row of Table~\ref{tab:trajauditor}.} These results suggest that \toolnameeval is more likely to identify tested trajectories not included in the agents' training dataset. Offline RL algorithms often suffer from overestimating value function training~\cite{bcq,cql,plasp}, leading to inaccurate computations of the value vectors for trajectories, mistakenly categorizing them as excluded from the training dataset. We study the robustness of \toolnameeval in Appendix~\ref{supsubsec:robustness}.

 \begin{figure}[!t]
 \vspace{+5pt}
    \centering
    \includegraphics[width=0.99\linewidth]{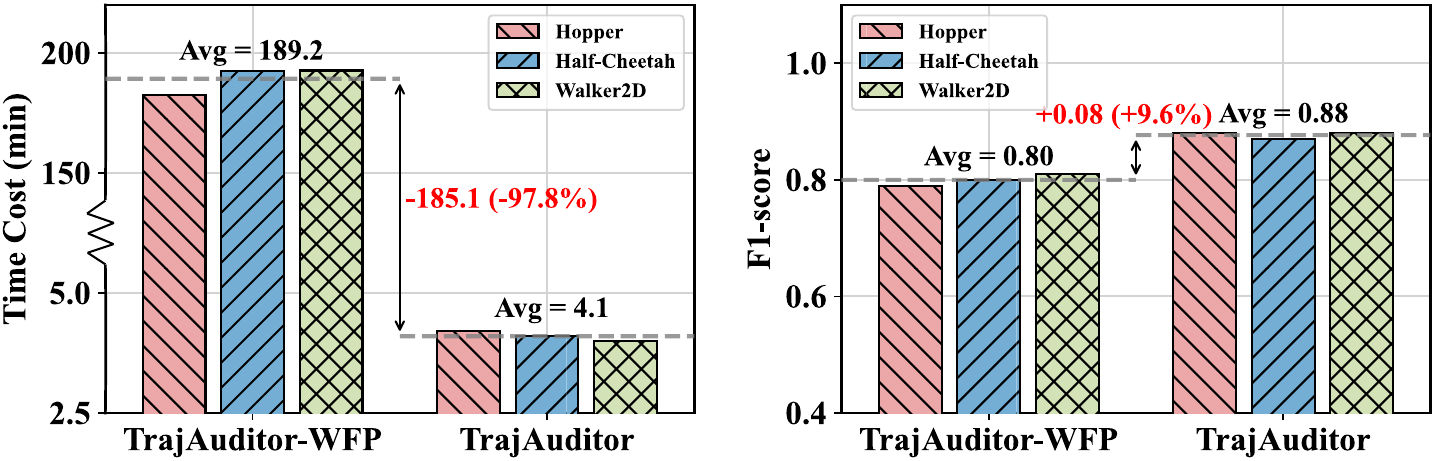}
    \caption{\rev{The time costs (left) and F1-scores (right) achieved by (1) our proposed \toolnameeval, (2) \toolnameeval without the fine-tuning component and excluding state perturbations. The `Avg' means the average of three tasks.}}
    \vspace{-6mm}
    \label{fig:auditor_abalation}
\end{figure}

 \begin{table*}[!t]
\small
    \centering
    \caption{ Percentages of positive predictions by \toolnameeval for the unlearning dataset post unlearning method application. $D_m$ indicates that the remaining dataset is not subjected to unlearning. The $D_{f,0.01}$ and $D_{f,0.05}$ denote the unlearning dataset with the size of 1\% and 5\% of the original dataset.}
    \resizebox{0.99\textwidth}{!}{
    \begin{tabular}{p{2.1cm}|c|ccc|ccc|ccc|ccc}
        \toprule
         \multirow{2}{*}{\textbf{Tasks}}   &  \multirow{2}{*}{\textbf{Algorithms}}  &\multicolumn{3}{c|}{\text{\textbf{Retraining (reference)}}} & \multicolumn{3}{c|}{\text{\textbf{Fine-tuning}}} & \multicolumn{3}{c|}{\text{\textbf{Random-reward}}}  & \multicolumn{3}{c}{\text{\textbf{TrajDeleter}}} \\
         \cline{3-14}
        & & $\mathcal{D}_m$ & $\mathcal{D}_{f,0.01}$ & $\mathcal{D}_{f,0.05}$  & $\mathcal{D}_m$ & $\mathcal{D}_{f,0.01}$ & $\mathcal{D}_{f,0.05}$ &  $\mathcal{D}_m$ & $\mathcal{D}_{f,0.01}$ & $\mathcal{D}_{f,0.05}$ & $\mathcal{D}_m$ & $\mathcal{D}_{f,0.01}$ & $\mathcal{D}_{f,0.05}$  \\
         \toprule
         \multirow{7}{*}{{\tt Hopper}} 
         & BEAR  & 89.0 $\%$ & 2.1 $\%$ & 4.2 $\%$ & 87.8 $\%$ & 90.9 $\%$ & 79.4 $\%$ & 84.2 $\%$ & 62.2 & 30.5 $\%$ & 85.1 $\%$ & 0.0 $\%$ & 0.0 $\%$ \\
          & BCQ  &  82.1 $\%$ & 0.0 $\%$ & 2.9 $\%$ & 83.4 $\%$ & 40.6 $\%$ & 47.1 $\%$ & 83.2 $\%$ & 36.4 $\%$ & 28.4 $\%$ & 80.2 $\%$ & 0.0 $\%$ & 0.0 $\%$ \\
          & CQL  &  85.2  $\%$ & 0.0 $\%$ & 0.0  $\%$ & 84.9 $\%$ & 80.6 $\%$ & 75.1 $\%$ & 88.1 $\%$ & 77.2 $\%$ & 43.5 $\%$ & 83.2 $\%$ & 0.0 $\%$ & 0.0 $\%$ \\
          & IQL & 88.1  $\%$ & 0.0 $\%$ & 5.4  $\%$ & 84.5 $\%$ & 81.1 $\%$ & 82.4 $\%$ & 84.5 $\%$ & 58.4 $\%$ & 36.7 $\%$ & 85.2 $\%$ & 0.0 $\%$ & 0.0 $\%$ \\
          & PLAS-P  & 45.8  $\%$ & 0.0 $\%$ & 0.0 $\%$ & 54.3 $\%$ & 77.6 $\%$ & 80.7 $\%$ & 57.8 $\%$ & 57.1 $\%$ & 46.7 $\%$ & 56.3 $\%$ & 40.5 $\%$ & 47.5 $\%$ \\
         & TD3PlusBC  & 82.4  $\%$ & 7.2  $\%$ & 9.8 $\%$ & 83.1 $\%$ & 75.4 $\%$ & 74.3 $\%$ & 83.8 $\%$ & 77.3 $\%$ & 60.1 $\%$ & 84.3 $\%$ & 0.0 $\%$ & 0.0 $\%$ \\
        \cline{2-14}
        & \textbf{Average} & \cellcolor{gray!20}\textbf{78.8 $\%$} & \cellcolor{gray!20}\textbf{1.6 $\%$} & \cellcolor{gray!20}\textbf{3.7 $\%$} & \cellcolor{gray!20}\textbf{79.7 $\%$} & \cellcolor{gray!20}\textbf{74.4 $\%$} & \cellcolor{gray!20}\textbf{73.2 $\%$} & \cellcolor{gray!20}\textbf{80.3 $\%$} & \cellcolor{gray!20}\textbf{61.4 $\%$} & \cellcolor{gray!20}\textbf{41.8 $\%$} & \cellcolor{gray!20}\textbf{79.1 $\%$} & \cellcolor{gray!20}\textbf{6.8 $\%$} & \cellcolor{gray!20}\textbf{7.9 $\%$} \\
         \toprule
        \multirow{7}{*}{\shortstack{{\tt Half-} \\ {\tt Cheetah}}}
        & BEAR & 80.1  $\%$ & 0.0 $\%$ & 6.5 $\%$ & 82.8 $\%$ & 60.8 $\%$ & 51.3 $\%$ & 82.9 $\%$ & 40.5 $\%$ & 10.2 $\%$ & 80.1 $\%$ & 0.0 $\%$ & 0.0 $\%$ \\
         & BCQ & 80.0  $\%$ & 0.0 $\%$ & 0.0 $\%$ & 82.7 $\%$ & 70.0 $\%$ & 28.0 $\%$ & 79.5 $\%$ & 0.0 $\%$ & 0.0 $\%$ & 83.1 $\%$ & 0.0 $\%$ & 0.0 $\%$ \\
         & CQL & 82.7  $\%$ & 0.0 $\%$ & 0.0 $\%$ & 81.7 $\%$ & 66.6 $\%$ & 52.6 $\%$ & 81.3 $\%$ & 24.6 $\%$ & 0.0 $\%$ & 77.9 $\%$ & 0.0 $\%$ & 0.0 $\%$ \\
         & IQL & 78.9  $\%$ & 0.0  $\%$ & 0.0 $\%$ & 80.7 $\%$ & 0.0 $\%$ & 0.0 $\%$ & 84.4 $\%$ & 0.0 $\%$ & 0.0 $\%$ & 82.0 $\%$ & 0.0 $\%$ & 0.0 $\%$ \\
         & PLAS-P& 53.7  $\%$ & 0.0 $\%$ & 0.0 $\%$ & 64.2 $\%$ & 74.0 $\%$ & 87.2 $\%$ & 56.7 $\%$ & 74.5 $\%$ & 77.2 $\%$ & 58.3 $\%$ & 1.8 $\%$ & 3.5 $\%$ \\
         & TD3PlusBC& 75.3  $\%$ & 0.0 $\%$ & 0.0 $\%$ & 75.6 $\%$ & 45.5 $\%$ & 54.7 $\%$ & 82.2 $\%$ & 80.0 $\%$ & 25.6 $\%$ & 78.2 $\%$ & 0.0 $\%$ & 0.0$\%$ \\
         \cline{2-14}
        & \textbf{Average} & \cellcolor{gray!20}\textbf{75.1 $\%$} & \cellcolor{gray!20}\textbf{0.0 $\%$} & \cellcolor{gray!20}\textbf{1.1 $\%$} & \cellcolor{gray!20}\textbf{78.1 $\%$} & \cellcolor{gray!20}\textbf{52.8 $\%$} & \cellcolor{gray!20}\textbf{45.6 $\%$} & \cellcolor{gray!20}\textbf{77.8 $\%$} & \cellcolor{gray!20 }\textbf{36.6 $\%$} & \cellcolor{gray!20}\textbf{18.8 $\%$} & \cellcolor{gray!20}\textbf{76.6 $\%$} & \cellcolor{gray!20}\textbf{0.3 $\%$} & \cellcolor{gray!20}\textbf{0.6 $\%$} \\
        \toprule
        \multirow{7}{*}{{\tt Walker2D}} 
        & BEAR& 82.5  $\%$ & 6.2 $\%$ & 5.2 $\%$ & 83.6 $\%$ & 69.5 $\%$ & 71.9 $\%$ & 84.5 $\%$ & 29.5 $\%$ & 8.6 $\%$ & 84.3 $\%$ & 0.0 $\%$ & 0.6 $\%$ \\
         & BCQ &  80.2 $\%$ & 0.0 $\%$ & 1.4 $\%$ & 78.5 $\%$ & 40.6 $\%$ & 31.5 $\%$ & 79.8 $\%$ & 21.6 $\%$ & 0.0 $\%$ & 79.5 $\%$ & 0.0 $\%$ & 0.2 $\%$ \\
          & CQL & 79.3  $\%$ & 0.0 $\%$ & 0.9 $\%$ & 84.2 $\%$ & 52.7 $\%$ & 64.5 $\%$ & 81.5 $\%$ & 46.5 $\%$ & 5.3 $\%$ & 82.3 $\%$ & 0.0 $\%$ & 0.0 $\%$ \\
         & IQL &  78.9 $\%$ & 0.0 $\%$ & 0.0 $\%$ & 78.5 $\%$ & 65.2 $\%$ & 49.9 $\%$ & 78.0 $\%$ & 48.6 $\%$ & 3.5 $\%$ & 77.5 $\%$ & 0.0 $\%$ & 0.3 $\%$ \\
         & PLAS-P & 80.6  $\%$ & 1.4 $\%$ & 4.3 $\%$ & 81.6 $\%$ & 66.7 $\%$ & 70.8 $\%$ & 81.9 $\%$ & 69.6 $\%$ & 74.5 $\%$ & 83.2 $\%$ & 51.6 $\%$ & 60.7 $\%$ \\
         & TD3PlusBC & 84.6  $\%$ & 1.5 $\%$ & 2.1 $\%$ & 84.3 $\%$ & 79.7 $\%$ & 58.1 $\%$ & 82.8 $\%$ & 79.5 $\%$ & 10.5 $\%$ & 84.7 $\%$ & 0.0 $\%$ & 1.3 $\%$ \\
         \cline{2-14}
          & \textbf{Average} & \cellcolor{gray!20}\textbf{81.0 $\%$} & \cellcolor{gray!20}\textbf{1.5 $\%$} & \cellcolor{gray!20}\textbf{2.3 $\%$} & \cellcolor{gray!20}\textbf{81.8 $\%$} & \cellcolor{gray!20}\textbf{62.4 $\%$} & \cellcolor{gray!20}\textbf{57.8 $\%$} & \cellcolor{gray!20}\textbf{81.4 $\%$} & \cellcolor{gray!20}\textbf{49.2 $\%$} & \cellcolor{gray!20}\textbf{17.1 $\%$} & \cellcolor{gray!20}\textbf{81.9 $\%$} & \cellcolor{gray!20}\textbf{8.6 $\%$} & \cellcolor{gray!20}\textbf{10.3 $\%$} \\
        \bottomrule
    \end{tabular}
    }
    \label{tab:efficacy}
    \vspace{-2mm}
    \end{table*}

We conduct ablation studies to highlight the importance of generating shadow agents by fine-tuning original agents and applying state perturbation in \toolnameeval. \rev{ \toolnameeval-WFP denotes the removal of state perturbations in the tested trajectory and fine-tuning of agents from the original agents. By directly training shadow models from scratch without fine-tuning or state perturbation, \toolnameeval is essentially reduced to \orlauditor. Thus, we equate \toolnameeval-WFP with \orlauditor. Figure~\ref{fig:auditor_abalation} shows that, on average across three tasks, \toolnameeval reduces time costs by 97.8\% while achieving an F1-score that is 0.08 higher compared to \toolnameeval-WFP. These results present that our method can enhance the \toolnameeval's performance. }

\noindent \textbf{Answers to RQ1}: \toolnameeval achieves average F1-scores of 0.88, 0.87, and 0.88 across three tasks, \rev{tested at four different unlearning rates.} These results present that \toolnameeval accurately identifies trajectories involved in agents' training datasets. It is a simple yet efficient tool for assessing the efficacy of offline reinforcement unlearning.



\subsection*{\textbf{RQ2. How effective is \toolname?}}

\noindent \textbf{Experiment Design.} 
We evaluate \toolname from four perspectives as we described in Section~\ref{app:principles}. Given that the unlearning dataset is typically a small portion of the original dataset, we examine scenarios where 1\% and 5\% of the offline dataset (i.e., unlearning rates of 0.01 and 0.05) are segregated as unlearning datasets. The remaining portion of the dataset is then considered as the remaining dataset. For retraining from scratch, we follow the methodology detailed in RQ1. The retrained agents serve as a reference for comparing other unlearning methods (i.e., fine-tuning, random-reward, and \toolname). The unlearning steps for the other methods are set at $1 \times 10^4$, amounting to only 1\% of the steps required for retraining ($1 \times 10^6$). The durations for ``forgetting'' ($K$) and ``convergence training'' ($H$) are set at 8000 and 2000 timesteps, respectively. Additionally, we set the balancing factor $\lambda$ at $1$ (as detailed in Section~\ref{subsec:trajdeleter_detail}).

\noindent \textbf{Result Analysis.} We evaluate the performance of \toolname from four distinct perspectives as follows.

\noindent \textit{Efficacy evaluation.} Table~\ref{tab:efficacy} presents the Percentage of Positive pRedictions (PPR) made by \toolnameeval, reflecting the extent to which the target dataset continues to influence the agents after the implementation of various unlearning methods. \rev{The last two columns in Table \ref{tab:efficacy} present that excluding the average values, only 40.5\%, 47.5\%, 51.6\%, and 60.7\% of the settings exceeded the 3.5\% in these 36 values.} These results show that \toolname can efficiently eliminate the impact of target trajectories on agents, with 32 out of 36 settings exhibiting a PPR below 3.5\% after unlearning.

Table~\ref{tab:efficacy} shows that the variance in PPR made by agents unlearned using \toolname is minimal compared to the outcomes achieved by retraining from scratch. Specifically, after retraining, \toolnameeval predicts that on average, only 2.7\%, 0.55\%, and 1.9\%\footnote{\rev{It is noticed that these figures are calculated as the mean value of PPR across two unlearning rates: (1.6\% + 3.7\%) / 2 = 2.7\%, (0.0\% + 1.1\%) / 2 = 0.55\%, and (1.5\% + 2.3\%) / 2 = 1.9\%. The other percentages mentioned in this paragraph are derived using the same method. }} of the unlearned dataset continues to influence the agent for the three tasks under investigation. In contrast, for \toolname, the average PPRs after unlearning are 7.35\%, 0.45\%, and 9.45\%. These results are significantly lower than those of the baseline methods -- after fine-tuning, the PPRs are 73.8\%, 44.2\%, and 60.1\%; while following the random-reward approach, the PPRs stand at 51.6\%, 27.7\%, and 33.2\%. \rev{By calculating the difference in average PPRs across the remaining dataset $\mathcal{D}_m$ for three tasks,} the PPRs for the remaining dataset exhibit only slight variations post-retraining, with a mere 0.9\% increase compared to the retraining method. Therefore, \toolname maintains the integrity of the trajectories' memory within the remaining dataset.

\begin{figure*}[!t]
    \centering
    \setlength{\abovecaptionskip}{2pt}
    \includegraphics[width=0.81\linewidth]{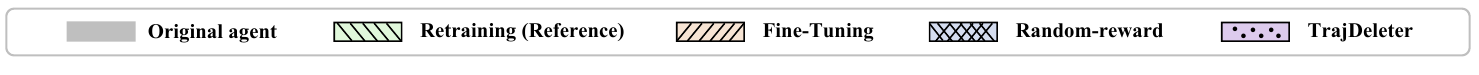}
    \includegraphics[width=1.0\linewidth]{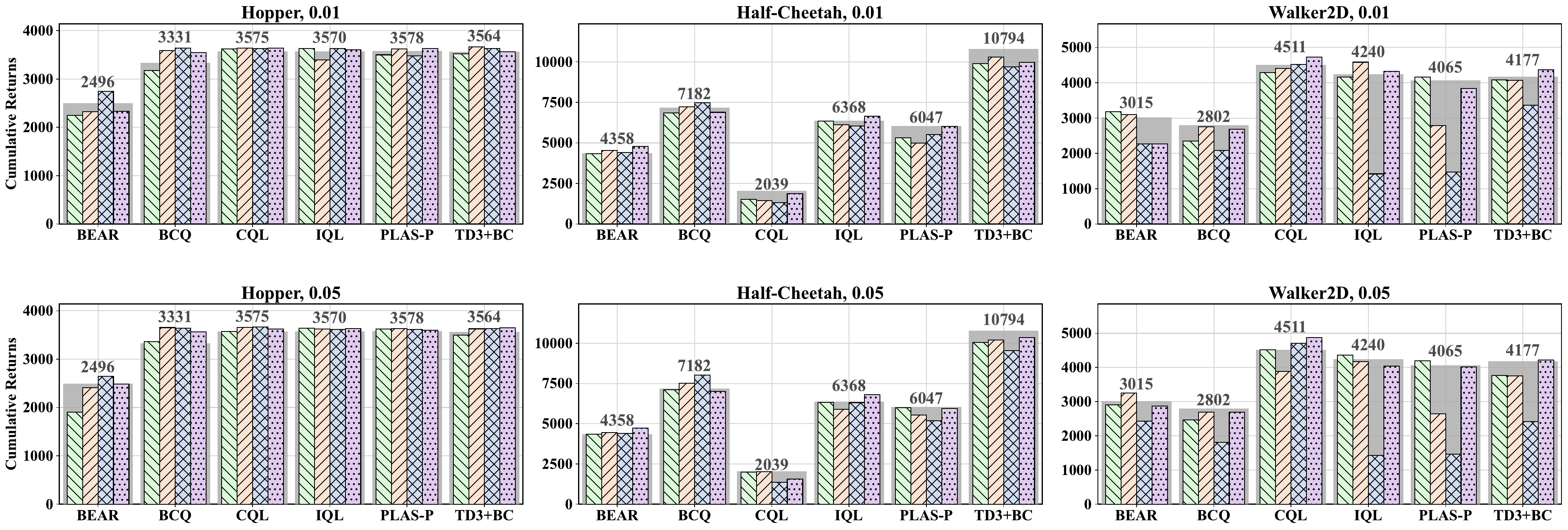}
    \vspace{-2mm}
    \caption{\rev{The cumulative returns are averaged over 100 test trajectories, collected using unlearned agents trained with 5 different random seeds. ``0.01'' and ``0.05'' represent the unlearning rates. The values beyond the gray bars indicate the cumulative returns of the original agents.} }
    \vspace{-6mm}
    \label{fig:trajdeleter_fidelity}
\end{figure*}

\vspace{3pt}
\noindent \textit{Fidelity evaluation.} Figure~\ref{fig:trajdeleter_fidelity} displays the averaged cumulative returns, calculated by averaging both the mean and variance across 100 test trajectories. As illustrated in Figure~\ref{fig:trajdeleter_fidelity}, the unlearned agents have a comparable performance level compared to those of the agents retrained from scratch. Specifically, the average cumulative returns demonstrate a marginal difference of 2.2\%, 0.9\%, and 1.6\% between the unlearned agents using \toolname and those subjected to the retraining method. These results suggest that \toolname does not negatively impact the performance of unlearned agents in real-world interactions, demonstrating the high practicality of our method. More analysis are provided in Appendix~\ref{appsub:fidelity}.

\vspace{3pt}
\noindent \textit{Efficiency evaluation.} Table~\ref{tab:time} presents the averaged time costs required for unlearning across \toolname and its baselines in the three tasks. \rev{\toolname requires only 3.4, 3.2, and 3.4 minutes for investigated tasks, significantly less than the 200.4, 223.4, and 100.4 minutes required for retraining from scratch. These results present that \toolname requires only 1.5\% of the time compared to retraining from scratch.} Compared to other methods, \toolname introduces no additional computational steps and exhibits a similar time cost, showing the high efficiency of our approach.

\noindent \textit{Agent agnostic evaluation.} 
We have selected six offline RL algorithms to train agents, determining the efficacy of \toolname in unlearning specific trajectories. Most agents (32 out of 36) can effectively forget the target trajectories without substantial degradation in performance. However, Table~\ref{tab:efficacy} shows that \toolname achieves over 40.5\% PPR when unlearning agents trained using the PLAS-P algorithm in {\tt Hopper} and {\tt Walker2D}. These results could be attributed to inaccuracies in \toolnameeval. \rev{As shown in Table~\ref{tab:trajauditor}, \toolnameeval gets explicitly low recall rates (e.g., with an unlearning rate of 0.01, the recall rate for the {\tt Hopper} task is 54.01\%) for agents trained using PLAS-P.}


\noindent  \textbf{Answers to RQ2}:
\rev{The average PPRs after unlearning are 7.35\%, 0.45\%, and 9.45\% for investigated tasks, meaning that \toolname unlearns 92.7\%, 99.5\%, 90.5\% of the target trajectories.} Yet \toolname maintains robust performance in actual environment interactions. The fact that 32 out of 36 settings display a PPR below 3.5\% shows the superior agent-agnostic capability of \toolname.

\begin{table}[!t]
\vspace{-0.2cm}
    \centering
    \small
    \caption{The time costs required for unlearning across \toolname and its baselines in the three tasks.}
    \resizebox{0.45\textwidth}{!}{
    \begin{tabular}{p{2.1cm}|ccc}
        \toprule
         \multirow{2}{*}{\textbf{Methods}}   &\multicolumn{3}{c}{\text{\textbf{Tasks}}}   \\
         \cline{2-4}
        & {\tt Hopper} & {\tt Half-Cheetah} & {\tt Walker2D}  \\
         \toprule
         \multirow{1}{*}{{Retraining}} 
         &  200.4 min & 223.4 min & 199.4 min  \\
        \multirow{1}{*}{{Fine-tuning}}
        &   3.7 min  & 3.6 min & 3.2 min  \\
        \multirow{1}{*}{{Random-reward}}
        &   3.5 min & 3.4 min & 3.2 min  \\
        \hline
        \cellcolor{gray!20}\multirow{1}{*}{{TrajDeleter}}
        &  \cellcolor{gray!20} 3.4 min & \cellcolor{gray!20} 3.2 min & \cellcolor{gray!20} 3.4 min \\
        \bottomrule
    \end{tabular}
    }
    \vspace{-3mm}
    \label{tab:time}
    \end{table}

 \begin{figure}[!t]
    \centering
    \includegraphics[width=0.81\linewidth]{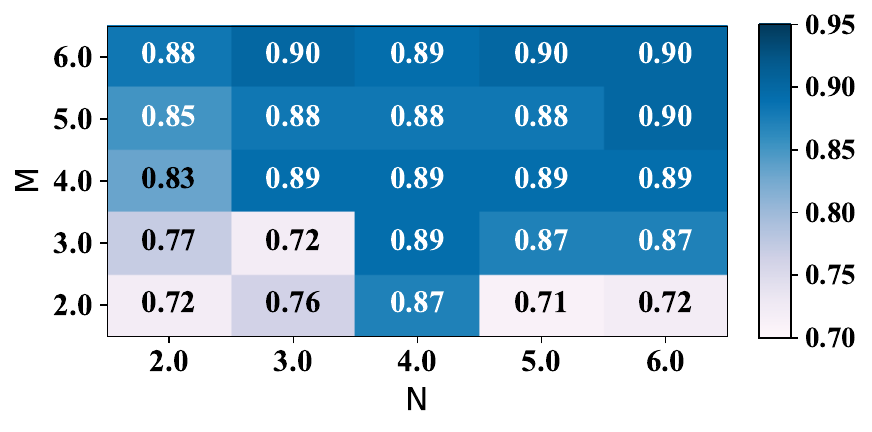}
    \vspace{-2mm}
    \caption{\rev{The average F1-score of \toolnameeval (the unlearning rate is 0.01) for the {\tt Half-Cheetah} task across $M$ and $N$.}}
    \vspace{-6mm}
    \label{fig:m_n_hyper}
\end{figure}

\subsection*{\textbf{RQ3. How do hyper-parameters affect the performance of \toolname?}} 

\noindent \textbf{Experiment Design.} This section first explores the impact of the forgetting learning steps, $K$, and the balancing factor, $\lambda$, on the unlearning performance of \toolname. The forgetting steps, $K = \{0, 2000, 4000, 6000, 8000\}$. The balancing factor $\lambda$ is to balance the
unlearning on forgetting datasets and training on remaining datasets, thereby preventing deterioration in the agent's performance. We have configured it with values $\{0.25, 0.5, 0.75, 1.0, 1.5\}$. \rev{Then, to explore the impact of the number of shadow models and perturbation rounds in \toolnameeval, denoted as $N$ and $M$, we vary these parameters across the values $\{2, 3, 4, 5, 6\}$.}

\noindent \textbf{Result Analysis.} The first, second, and third subfigures in Figure~\ref{fig:hyper-parameters} present the average trends in cumulative reward and Percentage of Positive pRedictions (PPR) as obtained by \toolnameeval for the unlearned agents across varying forgetting training steps. \rev{As the number of forgetting steps increases from 0 to 8000, the average PPRs of unlearned agents decrease from 74.4\%, 52.8\%, and 62.4\% to 7.9\%, 0.6\%, and 10.5\% for the three tasks, respectively.} In contrast, the agents' performance remains consistent. As illustrated in the fourth subfigure of Figure~\ref{fig:hyper-parameters}, when the forgetting training is adequate, the $\lambda$ exerts minimal influence on the performance of \toolname. Moreover, the last subfigure in Figure~\ref{fig:hyper-parameters} indicates that when the number of forgetting steps is small, an increase in $\lambda$ can enhance the unlearning efficiency of \toolname. Overall, $K$ exhibits greater sensitivity than $\lambda$, suggesting that the primary focus should be tuning $K$. \rev{Figure~\ref{fig:m_n_hyper} presents the average F1-score of \toolnameeval across agents trained with six offline RL algorithms for the {\tt Half-Cheetah} task, varying the parameters $M$ and $N$. In this figure, we observe that as the numbers of $M$ and $N$ increase, the F1 scores rise from 0.71 to a plateau of 0.90. } We provide a more comprehensive analysis in Appendix~\ref{appsub:hyper-parameter}.

\begin{figure*}[!t]
    \centering
    \setlength{\abovecaptionskip}{0pt}
    \includegraphics[width=1.0\linewidth]{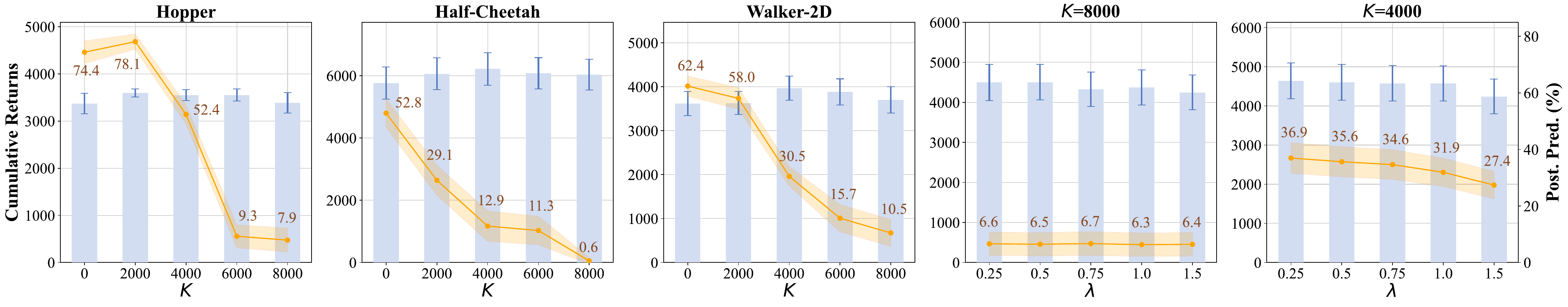}
    \caption{The influence of forgetting learning steps, $K$, and the balancing factor, $\lambda$, on the unlearning performance of \toolname. Yellow points denote the average percentages of Positive Predictions (Post. Pred.) by \toolnameeval for unlearned agents, aligning with the \textit{left} y-axis. Shaded areas denote the standard deviation. Light blue bars illustrate the average returns, corresponding to the \textit{right} y-axis, with the error bars indicating their standard deviation. The unlearning rate is 0.05.}
    \vspace{-5.5mm}
    \label{fig:hyper-parameters}
\end{figure*}

\noindent \textbf{Answers to RQ3}:  With an increase in the number of forgetting steps, the performance of \toolname improves, leading to the unlearned agent forgetting more trajectories. When the number of forgetting steps is high, \toolname exhibits minimal sensitivity to changes in the balancing factor's value.

\section{Discussions}
\label{sec:dis}


\begin{table}[!t]
\small
    \centering
    \caption{The relative changes in percentages of positive predictions of \toolnameeval on remaining dataset $\mathcal{D}_m$ and unlearning dataset $\mathcal{D}_f$, and average returns of unlearned agents just using ``forgetting'', when compared with its performance after doing ``convergence training''. The symbols `{\color{red} $\downarrow$}', `{\color{blue} $\uparrow$}', and `-' denote decreases, increases, and no changes.}
    \resizebox{0.49\textwidth}{!}{
   \begin{tabular}{p{1.2cm}|c|l|ll|ll}
        \toprule
         \multirow{3}{*}{\textbf{Tasks}}  & \multirow{3}{*}{\textbf{Algorithms}} & \multirow{3}{*}{$D_m$} & \multicolumn{4}{c}{\textbf{Unlearning rates}} \\
         \cline{4-7}
         & & & \multicolumn{2}{c|}{\text{\textbf{0.01}}} & \multicolumn{2}{c}{\text{\textbf{0.05}}} \\
         \cline{4-7}
         & &  & $D_{f}$ & \textbf{Returns} & \textbf{$D_{f}$} & \textbf{Returns} \\
         \hline
         \multirow{7}{*}{{\tt Hopper}}
         & BEAR & {\color{red} $\downarrow$}2.1\% & - 0.0 \% & {\color{red} $\downarrow$}1499 & - 0.0 \% & {\color{red} $\downarrow$}1429 \\
         & BCQ & {\color{red} $\downarrow$}7.2\% & - 0.0 \% & {\color{red} $\downarrow$}2271 & - 0.0 \% & {\color{red} $\downarrow$}2562 \\
         & CQL & {\color{red} $\downarrow$}1.3\% & - 0.0 \% & {\color{red} $\downarrow$}1903 & - 0.0 \% & {\color{red} $\downarrow$}852 \\
         & IQL &  - 0.0\% & - 0.0 \% & {\color{red} $\downarrow$}345 & - 0.0 \% & {\color{red} $\downarrow$}980 \\
         & PLAS-P & {\color{red} $\downarrow$}45.5\% & {\color{blue} $\uparrow$}0.7 \% & {\color{red} $\downarrow$}1783 & - 0.0 \% & {\color{red} $\downarrow$}1732 \\
         & TD3PlusBC & {\color{red} $\downarrow$}1.4\% & - 0.0 \% & {\color{blue} $\uparrow$}20 & - 0.0 \% & {\color{red} $\downarrow$}75 \\
        \cline{2-7}
         & \cellcolor{gray!20}  \textbf{Average} & \cellcolor{gray!20}{\color{red} $\downarrow$}9.6\% & \cellcolor{gray!20}{\color{blue} $\uparrow$} 0.1\% & \cellcolor{gray!20}{\color{red} $\downarrow$}1297 & \cellcolor{gray!20} -0.0\% & \cellcolor{gray!20}{\color{red} $\downarrow$}1272 \\
          \toprule
         \multirow{7}{*}{\shortstack{{\tt Half-} \\ {\tt Cheetah}}}
         & BEAR & - 0.0\% & - 0.0 \% & {\color{blue} $\uparrow$}772 & - 0.0 \% & {\color{blue} $\uparrow$}611 \\
         & BCQ & {\color{red} $\downarrow$}13.2\% & - 0.0 \% & {\color{blue} $\uparrow$}535 & - 0.0 \% & {\color{blue} $\uparrow$}451 \\
         & CQL & {\color{red} $\downarrow$}4.2\% & - 0.0 \% & {\color{red} $\downarrow$}100 & - 0.0 \% & {\color{red} $\downarrow$}344 \\
         & IQL & - 0.0\% & - 0.0 \% & {\color{blue} $\uparrow$}382 & - 0.0 \% & {\color{blue} $\uparrow$}431 \\
         & PLAS-P & {\color{red} $\downarrow$}1.9\% & {\color{blue} $\uparrow$}0.1 \% & {\color{red} $\downarrow$}102 & - 0.0 \% & {\color{red} $\downarrow$}105 \\
         & TD3PlusBC & {\color{red} $\downarrow$}32.1\% & - 0.0 \% & {\color{blue} $\uparrow$}218 & - 0.0 \% & {\color{red} $\downarrow$}312 \\
        \cline{2-7}
        & \cellcolor{gray!20}  \textbf{Average} & \cellcolor{gray!20}{\color{red} $\downarrow$}8.6\% & \cellcolor{gray!20}- 0.0\% & \cellcolor{gray!20}{\color{blue} $\uparrow$}284 & \cellcolor{gray!20} - 0.0\% & \cellcolor{gray!20}{\color{blue} $\uparrow$}122 \\
         \toprule
         \multirow{7}{*}{\shortstack{{\tt Walk-} \\ {\tt er2D}}} 
         & BEAR & {\color{red} $\downarrow$}5.9\% & - 0.0 \% & {\color{red} $\downarrow$}134 & - 0.0 \% & {\color{red} $\downarrow$}755 \\
         & BCQ & {\color{red} $\downarrow$}6.1\% & - 0.0 \% & {\color{blue} $\uparrow$}258 & - 0.0 \% & {\color{blue} $\uparrow$}104 \\
         & CQL & {\color{red} $\downarrow$}10.1\% & - 0.0 \% & {\color{red} $\downarrow$}2712 & - 0.0 \% & {\color{red} $\downarrow$}2064 \\
         & IQL & - 0.0 \% & - 0.0 \% & {\color{red} $\downarrow$}101 & - 0.0 \% & {\color{blue} $\uparrow$}289 \\
         & PLAS-P & {\color{red} $\downarrow$}2.1\% & {\color{blue} $\uparrow$}3.1 \%  & {\color{blue} $\uparrow$}121 & {\color{blue} $\uparrow$}2.9 \% & {\color{red} $\downarrow$}885 \\
         & TD3PlusBC & - 0.0\% & - 0.0 \% & {\color{red} $\downarrow$}488 & - 0.0 \% & {\color{red} $\downarrow$}144  \\
        \cline{2-7}
         & \cellcolor{gray!20}  \textbf{Average} & \cellcolor{gray!20}{\color{red} $\downarrow$}4.0\% & \cellcolor{gray!20}{\color{blue} $\uparrow$}0.5\% & \cellcolor{gray!20}{\color{red} $\downarrow$}509 & \cellcolor{gray!20}{\color{blue} $\uparrow$}0.5\% & \cellcolor{gray!20}{\color{red} $\downarrow$}576 \\
        \bottomrule
    \end{tabular}
    }
    \label{tab:relative_changes}
\end{table}

\subsection{Ablation Studies}
\label{subsec:ablation}
This section conducts ablation studies to emphasize the importance of ``convergence training'' in the effectiveness of \toolname. Our experiments focus solely on implementing ``forgetting'' training with \toolname, aimed at unlearning the 1\% dataset, without engaging in convergence training. In this experiment, the unlearning rates are set at 0.01 and 0.05. The results are illustrated in Table~\ref{tab:relative_changes}. We observe that \textbf{``convergence training'' leads to stronger \toolname}. We attribute the observed outcomes to the following reasons: In the absence of convergence training for the three tasks: \rev{(1) at an unlearning rate of 0.05, the average cumulative returns of agents show changes of 1272, 122, and 576, respectively, with corresponding PPR changes on the unlearning dataset $\mathcal{D}_f$ of 0.0\%, 0.0\%, and 0.5\%;} (2) \toolnameeval experiences reductions in the average percentages of positive predictions \rev{by 9.6\%, 8.6\%, and 4.0\% on the remaining dataset}. These findings suggest that while convergence training has a minimal impact on the unlearning trajectories, it assists in enhancing the performance of the unlearned agent. Furthermore, it helps in preventing adverse effects on the trajectories within the remaining dataset.

\subsection{Defending Against Trajectory Poisoning}
\label{subsec:poisoning}

We also conduct experiments to investigate the effectiveness of \toolname in defending against trajectory poisoning attacks. We poison the original dataset by adjusting the action values in the trajectories to be 1.5 times their mean value. This modification is applied to only 5\% of the whole dataset. \textit{Poisoning rates} denote the
fraction of generated poisoned trajectories in the whole
dataset.

All other experimental settings remain consistent with those used for training the original agent. Table~\ref{tab:poisoning} shows the averaged returns of the poisoned agents, compared with those of agents after unlearning poisoned trajectories. After training the agents on a poisoned dataset, we observe a decrease in their average performance, which is 37.0\%, 7.0\%, and 18.8\% across the three tasks under investigation. \toolname can mitigate the effects of poisoning in agents, thereby enhancing their performance to match that of agents who have been retrained on a non-poisoned dataset from scratch.

\subsection{\rev{The Impact of Repeated Unlearning}}

\rev{This section explores how repeated learning and unlearning of the same trajectories impact the averaged returns of agents. We repeatedly unlearn and learn the same trajectories with unlearning rates of 0.01 and 0.05, doing so $\{1, 3, 7, 10\}$ times respectively, and record the averaged returns of agents trained using six offline RL algorithms in Table~\ref{tab:reduced}. Referring to the experiments in RQ1 and RQ2, the learning and unlearning phases consist of $1 \times 10^6$ and $1 \times 10^4$ timesteps, respectively. } 

\rev{Table~\ref{tab:reduced} shows no decreasing trend in the agent's cumulative returns with an increasing number of repeated learning cycles. In particular, compared to the original agents, the largest relative change in this table for agents that have repeatedly learned and removed the same trajectories is 4.0\%. In most cases, the relative changes are less than 2\%. While the agent's performance exhibits fluctuations, this variability is a natural aspect of training, commonly observed in offline RL~\cite{levine2020offline}. We believe that repeatedly unlearning and relearning the same trajectories do not lead to damage to agents' performance. }

\begin{table}[!t]
    \centering
    \small
    \caption{\rev{The average returns of the poisoned and unlearned agents. The values in parentheses are the relative changes compared to agents retrained on a non-poisoned dataset.}}
    \resizebox{0.47\textwidth}{!}{
    \begin{tabular}{p{1.3cm}|c|ccc}
        \toprule
         \multirow{2}{*}{\textbf{Methods}} &  \multirow{2}{*}{\shortstack{{ \textbf{Poisoning}} \\ { \textbf{rates}}}} & \multicolumn{3}{c}{\text{\textbf{Tasks}}}   \\
         \cline{3-5}
        & & {\tt Hopper} & {\tt Half-Cheetah} & {\tt Walker2D}  \\
         \toprule
        \multirow{1}{*}{{Retraining}}
        & - & \multirow{1}{*}{3357}  & \multirow{1}{*}{5761} & \multirow{1}{*}{3868}  \\
        \hline
         \multirow{3}{*}{{Poisoning}}
         & 0.01 & 2169 (\textbf{-35.4\%}) & 5263 (\textbf{-8.6\%}) &  3127 (\textbf{-19.2\%}) \\
         & 0.05 & 2059 (\textbf{-38.6\%})  & 5455 (\textbf{-5.3\%}) & 3158 (\textbf{-18.4\%})  \\
         \cline{2-5}
         & \cellcolor{gray!20} \textbf{Average} & \cellcolor{gray!20} 2114 (\textbf{-37.0\%})  & \cellcolor{gray!20} 5359 (\textbf{-7.0\%}) & \cellcolor{gray!20} 3158 (\textbf{-18.8\%})  \\
        \hline
        \multirow{3}{*}{{Trajdeleter}}
        & 0.01 & 3299 (\textbf{-1.7\%}) & 6029 (\textbf{+4.6\%}) & 3949 (\textbf{+2.1\%}) \\
        & 0.05 & 3421 (\textbf{+1.9\%}) & 5679 (\textbf{-1.4\%}) & 3855 (\textbf{-0.3\%})  \\
        \cline{2-5}
         & \cellcolor{gray!20} \textbf{Average} & \cellcolor{gray!20} 3360 (\textbf{+0.1\%})  & \cellcolor{gray!20} 5854 (\textbf{+1.6\%}) & \cellcolor{gray!20} 3902 (\textbf{+0.9\%})  \\
        \bottomrule
    \end{tabular}
    }
    \vspace{-2.7mm}
    \label{tab:poisoning}
    \end{table}



\subsection{Limitation}
\rev{\noindent \textbf{Legal Uncertainty.} The approach most aligned with requirements of GDPR and other data protection regulations is to retrain the ML model using a new dataset that excludes specific data. However, this approach is overly expensive in modern ML. As a practical alternative, approximate unlearning has been proposed as a compromise that does not fully delete every impact of the data to be deleted, but is much more efficient~\cite{unlearning_02_feature,zhang2024contrastive}, and \toolname follows these works. Certified unlearning~\cite{unlearning_02_feature} offers a theoretical guarantee that the output of unlearned models closely matches that of retrained models, potentially aiding legal compliance. However, applying it to complex models is challenging, and there remains uncertainty regarding its effectiveness in reinforcement unlearning. We discuss more in Appendix~\ref{app:legal}.}

\noindent \rev{ \textbf{Potential Attacks.} One potential attack is recovery, where an adversary attempts to reconstruct individual data entries from the training datasets of models~\cite{bertran2024reconstruction}. While unlearning complicates the data recovery process, the frequent application of unlearning and subsequent parameter modifications increase the difficulty for attackers attempting to retrieve specific data. However, preventing data recovery from unlearned models is a challenging problem~\cite{chen2021machine,liu2024survey}, and this attack is possible even for retraining~\cite{bertran2024reconstruction}. In some cases, model owners may deceive users by not performing unlearning. Even when model owners genuinely implement unlearning, previous research has suggested that the output distributions of both learned and unlearned models can be utilized to construct feature vectors to train the attack model~\cite{liu2024survey}.}

\noindent \rev{ \textbf{Effectiveness Uncertainty.} 
It is noticed that as an approximate method, the unlearning effect of \toolname relies on the effectiveness of the trajectories auditing method, i.e., \toolnameeval; this is a common practice~\cite{unlearning_02_feature,unlearning_survey}. We advocate enhancing the auditor tool to improve the reliability of unlearning effectiveness. We discuss more in Appendix~\ref{app:eff_un}.}

\begin{table}[!t]
    \centering
    \small
    \caption{\rev{The averaged returns of the agents which are measured through the repeatedly addition and removal of learning specific trajectories. The values in parentheses represent the relative changes compared to the original agent's return.}}
    \resizebox{0.47\textwidth}{!}{
    \begin{tabular}{p{0.7cm}|c|ccc}
        \toprule
         \multirow{2}{*}{\textbf{Times}} &  \multirow{2}{*}{\shortstack{{ \textbf{Unlearning}} \\ { \textbf{rates}}}} & \multicolumn{3}{c}{\text{\textbf{Tasks}}}   \\
         \cline{3-5}
        & & {\tt Hopper} & {\tt Half-Cheetah} & {\tt Walker2D}  \\
         \toprule
         \multirow{2}{*}{{1}}
         & 0.01 & 3473 (\textbf{+3.4\%}) & 5788 (\textbf{+0.4\%}) & 
3946 (\textbf{+2.5\%}) \\
         & 0.05 & 3493 (\textbf{+4.0\%}) & 5692 (\textbf{-1.1\%}) & 
3929 (\textbf{+1.6\%}) \\
         \hline
        \multirow{2}{*}{{3}}
        & 0.01 & 3294 (\textbf{-1.8\%}) & 5832 (\textbf{+1.2\%}) & 
3829 (\textbf{-1.0\%}) \\
        & 0.05 & 3401 (\textbf{+1.3\%}) & 5701 (\textbf{-1.0\%}) & 3782 (\textbf{-2.2\%}) \\
        \hline
        \multirow{2}{*}{{7}}
        & 0.01 & 3449 (\textbf{+2.7\%}) & 5678 (\textbf{-1.4\%}) & 3892 (\textbf{+0.6\%}) \\
        & 0.05 & 3378 (\textbf{+0.6\%}) & 5636 (\textbf{-2.1\%}) & 
 3832 (\textbf{-0.9\%}) \\
        \hline
        \multirow{2}{*}{{10}}
        & 0.01 & 3411 (\textbf{+1.6\%}) & 5712 (\textbf{-0.8\%}) & 3910 (\textbf{+1.1\%}) \\
        & 0.05 & 3409 (\textbf{+1.5\%}) & 5865 (\textbf{+1.8\%}) & 3783 (\textbf{-2.2\%}) \\
        \bottomrule
    \end{tabular}
    }
    \vspace{-2.8mm}
    \label{tab:reduced}
    \end{table}

\section{Related works}
\label{sec:related}

We discuss related work briefly here, and we provide a more comprehensive discussion in Appendix~\ref{suppsec:related}.

\noindent \textbf{Offline RL for Real Applications}
Recently, offline RL systems work brilliantly on a wide range of real-world fields, including healthcare~\cite{RL4Treatment,RL4BGC,offline_rl_medicial}, energy management systems~\cite{RL4Energy,zhang2023mutual}, autonomous driving~\cite{RL4AutonomousVehicles,RL4AutonomousVehicles2}, and recommendation systems~\cite{RL4Recommender,RL4Recommender2}. 
In healthcare, online RL is not suitable, as it is ethically and practically problematic to experiment with patients' health. Thus, Mila et al.~\cite{RL4Treatment} used advanced offline RL to develop a policy for sepsis treatment optimization. In various areas, offline RL methods are more efficient than online RL methods~\cite{RL4Energy,RL4Dialog,RL4Dialog3}. 

\noindent \textbf{Deep Machine Unlearning}
Deep machine unlearning~\cite{unlearning_01,unlearning_02_random,unlearning_02_feature,unlearning_03_understanding} refers to eliminating the knowledge of specific data point(s) on the already trained Deep Neural Networks (DNNs). In general, deep machine unlearning is categorized into two main groups: exact unlearning~\cite{unlearning_01,unlearning_04_amnesiac_exact,unlearning_05_ARCANE_exact} and approximate unlearning methods~\cite{unlearning_06_selective_app,unlearning_07_mixed_app,unlearning_09_zero_app}. Exact unlearning involves retraining the DNN from scratch without the data meant to be forgotten, which is computationally demanding due to large datasets~\cite{unlearning_08_resource}. Bourtoule et al.~\cite{unlearning_01} proposed the SISA method by splitting the dataset into non-overlapping shards, allowing retraining on just one shard. Unlike exact unlearning, approximate unlearning estimates DNN parameters similar to retraining from scratch~\cite{unlearning_survey,unlearning_06_selective_app,unlearning_10_scrubbing_app,unlearning_03_understanding}. 
Various recent works also studied certified unlearning definitions~\cite{unlearning_11_removal_cer,unlearning_12_deletion_cer}.

\noindent \textbf{Certified Unlearning} Recent studies have explored certified unlearning definitions~\cite{unlearning_11_removal_cer, unlearning_12_deletion_cer,unlearning_02_feature}, ensuring that unlearned models are theoretically indistinguishable from those retrained from scratch. For approximate unlearning, Guo et al.\cite{unlearning_11_removal_cer} have made the distributions of unlearned and retrained models nearly identical. Warnecke et al. ~\cite{unlearning_02_feature} introduced a certified unlearning for the features and labels unlearning.

\section{Conclusions}
\label{sec:con}
This paper introduces \toolname, the first practical trajectory-level unlearning method designed specifically for offline RL agents. \toolname enables agents to erase the influence of the target trajectory and ``forget'' it. This paper emphasizes approximate unlearning, which focuses more on classic supervised learning. To verify if trajectories are truly forgotten, we introduce \toolnameeval to evaluate the success of \toolname in completely removing the influence of specific trajectories from the offline RL agent, paving the path for unlearning study. For unlearning the target trajectories, \toolname is to prompt the agent to exhibit declining performance when encountering states linked to unlearning trajectories while preserving its original performance level for other remaining trajectories. Our evaluations present that \toolname consistently forgets the trajectories efficiently while maintaining strong performance in real environment interactions. Experiment results advocate that \toolname is an effective tool for offline reinforcement unlearning.

\section*{Acknowledgment}

We thank all the anonymous reviewers for their valuable comments. This paper was partially supported by NSF CNS-2350333.






%
\bibliographystyle{IEEEtranS}
\bibliography{bib}


\appendix

\section{Appendix}

\subsection{Investigated Tasks and the Dataset}
\label{app:task_dataset}

We carry out experiments across three of MuJoCo's robotic control tasks ({\tt Hopper}, {\tt Half-Cheetah}, and {\tt Walker2D})~\cite{mujoco}. In the three robotic control tasks, there is a sensor to monitor the state of the game, collecting information about the robot. Information collected from the sensor is used to observe the agent. 

In particular, the observation in {\tt Hopper} is a vector of size 11. Observation in {\tt Half-Cheetah} and {\tt Walker2D} is a 17-dimensional vector. These vectors record the positions, velocities, angles, and angular velocities of different components of a robot. We refer the readers to the online document\footnote{\url{https://www.gymlibrary.dev/index.html}} for a detailed explanation. We elaborate on these tasks as follows.

\begin{itemize}[leftmargin=*]
\item {\tt Hopper}: In Hopper,  the robot is a two-dimensional, single-legged entity comprising four principal components: the torso at the top, the thigh in the center, the leg at the lower end, and a single foot on which the entire body rests. The objective is to maneuver the robot forward (to the right) by exerting torques on the three hinges that interconnect these four body segments.

\item {\tt Half-Cheetah}: The robot is two-dimensional, featuring nine linkages and eight joints (including two paws), in the Half-Cheetah task. The aim is to apply torque to these joints, propelling the robot to sprint forward (to the right) as fast as possible. Progress is incentivized with positive rewards for distance covered in the forward direction, while a negative reward is allocated for moving backward. The torso and head of the robot remain stationary, with torque application restricted to the remaining six joints that connect the front and rear thighs to the torso, the shins to the thighs, and the feet to the shins.

\item {\tt Walker2D}: This task is an extension of Hopper within the MuJoCo suite. Walker2D introduces a greater number of independent state and control variables to more accurately emulate real-world scenarios. The robot in Walker2D is also two-dimensional but features a bipedal design with four main components: a single torso at the top from which the two legs diverge, a pair of thighs situated below the torso, a pair of legs below the thighs, and two feet attached to the legs that support the entire structure. The objective is to coordinate the movements of both sets of feet, legs, and thighs to progress forward (to the right) by applying torques to the six hinges that connect these body parts.

\end{itemize}

The datasets used for these tasks are sourced from D4RL~\cite{d4rl}, a recently introduced and mostly studied benchmark for evaluating offline RL algorithms. D4RL provides a variety of datasets for four tasks, gathered through diverse policies, including \textit{medium}, \textit{random}, \textit{medium-replay}, and \textit{medium-expert}. We elaborate on them as follows.

\begin{itemize}[leftmargin=*]
    \item \textit{medium}: The ``medium'' dataset involves online training of a policy using Soft Actor-Critic~\cite{sac}, followed by early termination of this training process and gathering 1M transitions from this partially-trained policy.
    \item \textit{random}:  The ``random'' datasets are created by deploying a randomly initialized policy across these three tasks.
    \item \textit{medium-replay}: The ``medium-replay'' dataset consists of all transitions recorded in the replay buffer during training, up until the policy attains a ``medium'' level of performance.
    \item \textit{expert}: The "expert" dataset is created by blending transitions collected from expert and suboptimal policies. This suboptimal data is generated either through a partially trained policy or by deploying a policy that operates randomly.
\end{itemize}

\begin{table}[!t]
\centering
\caption{Brief Introduction of each task.}
\small
\resizebox{0.45\textwidth}{!}{
\begin{tabular}{l|c|c|c}
    \toprule
      Tasks   &  Observations & Action Type & Data Size \\
    \midrule 
     {\tt Hopper}     & 11 & 3  & $2 \times 10^6$  \\
    {\tt Half-Cheetah}     & 17 & 6  & $2 \times 10^6$ \\
    {\tt Walker2D}   &  17 & 6  & $2 \times 10^6$ \\
    \bottomrule
\end{tabular}
}
\label{tab:discribe_env_breif}
\end{table}

To keep our experiments at a computationally manageable scale, we select the dataset that yields the highest returns for each task. We summarize the overview of datasets investigated in our experiments in Table~\ref{tab:discribe_env}.

\subsection{Principles in Offline Reinforcement Unlearning}
\label{app:principles}
Drawing upon principles from unlearning in supervised learning~\cite{unlearning_02_feature,guo2022efficient}, we define that an optimal offline reinforcement unlearning method must adhere to four properties: (1)  effectively erase the selected trajectories, (2) maintain the agent's performance quality, (3) offer efficiency when compared to completely retraining the agent from scratch, and (4) be applicable to a variety of agents. We elaborate on these principles as follows.

\noindent \textbf{Efficacy.} The primary objective for successful offline reinforcement learning is to erase as much as possible the agent's memory of the target trajectories. As this paper pioneers the concept of an unlearning method for offline RL, there is { no established criterion} to verify the effectiveness of unlearning within this area. Consequently,
Section~\ref{sub:efficacy} introduces a metric specifically designed to assess unlearning in offline reinforcement learning.

\noindent \textbf{Fidelity.} 
An unlearning method is truly valuable only when it retains the performance of unlearned agents closely aligned with the original agents. Consequently, our objective is that the unlearned agent will not experience significant performance degradation. This property evaluates how well the model retains knowledge of the trajectories that are not intended to be forgotten.

\begin{table*}[!t]
\centering
\caption{Information of each task and the dataset.}
\small
\resizebox{1\textwidth}{!}{
\begin{tabular}{l|l|l|c|c|l|c|l}
    \toprule
     Environments & Tasks   & Chosen Datasets &  Observations & Action Shape & Action Type & Data Size & Task Type\\
    \midrule 
    \multirow{3}{*}{MuJoCo~\cite{mujoco}}& {\tt Hopper}   & ``hopper-medium-expert''  & 11 & 3 & Continuous & $2 \times 10^6$ & Robotic Control \\
    & {\tt Half-Cheetah}   & ``halfcheetah-medium-expert''  & 17 & 6 & Continuous & $2 \times 10^6$ & Robotic Control\\
    & {\tt Walker2D}  & ``walker2d-medium-expert''  &  17 & 6 & Continuous & $2 \times 10^6$ & Robotic Control\\
    \bottomrule
\end{tabular}
}
\label{tab:discribe_env}
\end{table*}

\noindent \textbf{Efficiency.} A direct unlearning approach involves retraining agents from scratch on a dataset that excludes the target trajectories. However, this approach entails substantial runtime and storage overheads. The ideal unlearning method should minimize computational resource usage while ensuring that the unlearned agents perform as the retrained agent.

\noindent \textbf{Agent Agnostic.} The optimal approach for implementing offline reinforcement unlearning should be universally applicable across agents trained with various algorithms, which means that the objectives mentioned above should be met by any agent performing that unlearning strategy.

\subsection{Implementation}
\label{app:implement} 

We use the open-source implementation~\cite{d3rlpy} of selected offline RL algorithms, provided in their official repositories with consistent hyper-parameters, such as discount factor, optimizer, and network architecture. The number of training steps for each task is $1\times 10^6$. The training of all offline RL agents is conducted on a server configured with Python 3.7.11, equipped with four NVIDIA GeForce A6000 GPUs and 512GB of memory.

\subsection{Offline RL Algorithms}
\label{app:offline_rl_algorithms}
This section introduces six investigated offline RL algorithms in our experiments: (1) bootstrapping error accumulation reduction (BEAR)~\cite{kumar2019stabilizing}, (2) batch-constrained deep Q-learning (BCQ)~\cite{bcq}, (3) conservative Q-learning (CQL)~\cite{cql}, (4) implicit Q-learning (IQL)~\cite{iql}, (5) policy in the latent action space with perturbation (PLAS-P)~\cite{plasp}, and (6) twin delayed deep deterministic policy gradient plus behavioral cloning (TD3PlusBC)~\cite{td3plusbc}. 

\subsubsection{BEAR}
BEAR enhances the accuracy of value estimation, thereby stabilizing the performance of the algorithm. BEAR employs distribution-constrained backups to diminish the accumulation of bootstrapping errors. In its approach, BEAR utilizes $K$ Q-functions, selecting the minimum Q-value for policy improvement. The policy is updated to maximize the
conservative estimate of the Q-values within the state space $S$ defined by the set of possible policies. A sampled version
of maximum mean discrepancy (MMD)~\cite{MMD} is used to ensure that the actions proposed by the learned policy closely align with those in the dataset. The optimization problem, formulated during the policy improvement phase, is as follows,
$$
\pi_\phi:=\max _{\pi \in \Delta_{|S|}} \mathbb{E}_{s \sim \mathcal{D}} \mathbb{E}_{a \sim \pi(\cdot \mid s)}\left[\min _{j=1, . ., K} \hat{Q}_j(s, a)\right] 
$$
$$
\text { s.t. } \mathbb{E}_{s \sim \mathcal{D}}[\operatorname{MMD}(\mathcal{D}(s), \pi(\cdot \mid s))] \leq \varepsilon,$$
where $\varepsilon$ represents a carefully selected threshold parameter.

\subsubsection{BCQ}
BCQ encourages the agent to generate varied and exploratory actions to enhance performance. BCQ is
the first practical data-driven offline RL algorithm. The core concept behind BCQ involves incorporating a generative model to achieve the concept of batch-constrained learning, which means minimizing the difference between the proposed actions and the recorded actions in the dataset. To ensure a variety of actions, BCQ constructs a perturbation model to slightly modify each selected action. It then selects the action with the highest estimated value using a $Q$-network, which is trained to predict the expected cumulative reward for a specific state-action combination. Therefore, the objective function of BCQ can be formulated as,
$$ \pi(s)=\mathop{\arg\min}\limits_{a_i+ \xi_{\phi}(s, a_i,\Phi)} Q_{\theta}(s, a_i+ \xi_{\phi}(s, a_i,\Phi)), \left\{a_i \sim G_{\omega}(s)   \right\}^n_{i=1}, $$
where $G_{\omega}$ is a Conditional Variational Auto-Encoder (CVAE)~\cite{SohnLY15CVAE}, which generates $n$ potential actions. The perturbation model, denoted as $\xi_{\phi}(s, a_i,\Phi)$, applies a bounded noise to an action $a$, where the noise is restricted within the interval $[-\Phi, \Phi]$. The value function $Q_{\theta}$ scores these generated actions and selects the one with the highest estimated value. The perturbation model $\xi_{\phi}$ is optimized to maximize $Q_{\theta}(s, a)$ leveraging the deterministic policy gradient algorithm~\cite{Silver14DPG} by sampling $a \sim G_{\omega}(s)$,
$$\Phi \leftarrow \mathop{\arg\min}\limits_{\phi} \sum_{(s,a) \in \mathcal{B}} Q_{\theta}(a_i+ \xi_{\phi}(s, a_i,\Phi)),$$
where $\mathcal{B}$ signifies a batch of state-action pairs. BCQ modifies clipped double Q-learning~\cite{td3} to impose a penalty on the uncertainty associated with future state predictions. The revised learning objective is outlined as follows,
$$
r+\gamma \max _{a_i}\left[\lambda \min _{j=1,2} Q_{\theta_j^{\prime}}\left(s^{\prime}, a_i\right)+(1-\lambda) \max _{j=1,2} Q_{\theta_j^{\prime}}\left(s^{\prime}, a_i\right)\right],
$$
where $\left\{ Q_{\theta_1}, Q_{\theta_2}  \right\}$ are two Q-networks and $a_i $ represents the actions outputed from the generative model $G_{\omega}$ with perturbation.

\subsubsection{CQL}
CQL highlights a conservative approach to Q-value estimation, which secures the stability of the training process. By providing a conservative estimate of $Q$-values and introducing regularization techniques, CQL strikes a balance between exploration and safety in the absence of real-time interaction with the environment. It has demonstrated promising results in various applications, including robotics, autonomous systems, and decision-making tasks, where collecting new data is not feasible or safe.

At its core, CQL aims to minimize Q-values across a carefully selected distribution of state-action pairs. This objective is further tightened by adding a maximization component based on the data distribution.
Through this process, CQL generates Q-values that provide a lower bound for the value of a policy $\pi$. For policy optimization, CQL employs $\mu (a|s)$, representing the expected Q-value within a specific state-action distribution, to approximate
the policy that maximizes the current Q-function iteration. In addition, CQL incorporates regularization of Q-values during its training phase. An example of such a regularizer, $R(\mu)$, could be the KL-divergence relative to a prior distribution,
$\mu(a|s)$, namely $R(\mu) = -D_{KL}(\mu, \rho)$.

\subsubsection{IQL}
IQL uses implicit quantile networks to estimate the distributional values of state-action pairs, rather than merely calculating their expected value as prior methods, and thus provides a more robust method for value estimation. Compared to the batch-constrained idea of BCQ, IQL strictly avoids querying values of the actions, which are not in the pre-collected dataset. The loss function of IQL is defined as the following,
$$
\mathcal{L}(\theta)=\mathbb{E}_{\left(s, a, s^{\prime}, a^{\prime}\right) \sim \mathcal{D}}\left[(\mathcal{L}_2^{\tau}(r+\gamma Q_{\hat{\theta}}\left(s^{\prime}, a^{\prime}\right)-Q_\theta(s, a)))^2\right],
$$
where $\mathcal{L}_2^\tau(u)$ is defined as $|\tau- \mathds{  1 } (u<0)| u^2$, with $s^{\prime}$ and $ a^{\prime}$ denoting the subsequent state and action of $s$ and $ a$, respectively. $Q_{\theta}(s, a)$ and $Q_{\hat{\theta}}(s, a)$ correspond to the parameterized Q-function and its target network counterpart. IQL leverages a distinct value function to estimate the expectile, which induces the following loss,
$$
\mathcal{L}_V(\psi)=\mathbb{E}_{(s, a) \sim \mathcal{D}}\left[\mathcal{L}_2^\tau(Q_{\hat{\theta}}(s, a)-V_\psi(s))\right].
$$
Following this, the Q-function is updated through the application of the value function, utilizing the Mean Squared Error (MSE) loss in the subsequent form,
$$
\mathcal{L}_Q(\theta)=\mathbb{E}_{\left(s, a, s^{\prime}\right) \sim \mathcal{D}}\left[\left(r(s, a)+\gamma V_\psi\left(s^{\prime}\right)-Q_\theta(s, a)\right)^2\right].
$$

IQL employs a separate step to extract the policy while circumventing the use of actions beyond the sample space. The structured loss function is delineated as,
$$
\mathcal{L}_\pi(\phi)=\mathbb{E}_{(s, a) \sim \mathcal{D}}\left[\exp (\beta(Q_{\hat{\theta}}(s, a)-V_\psi(s))) \log \pi_\phi(a \mid s)\right],
$$
where $\beta \in  [0, \infty]$ is an inverse temperature.

\subsubsection{PLAS-P}
To ensure stable training, PLAS-P represents the policy by a CVAE in the action latent space to implicitly constrain the policy to output actions within the support range of the dataset. Initially,  
PLAS-P trains a deterministic policy $z = \pi(s)$ to map a state $s$ to a “latent action” $z$. Subsequently,  a pretrained decoder $p_{\beta}(a|s, z)$ projects the latent action into the actual action space. The goal of the CVAE is to maximize $\log p(a|s)$ by optimizing its lower bound,
$$
\begin{aligned}
\max _{\alpha, \beta} \log p(a \mid s) &\geq \max _{\alpha, \beta} \mathbb{E}_{z \sim q_\alpha}\left[\log p_\beta(a \mid s, z)\right] \\
&-\mathcal{D}_{KL}\left[q_\alpha(z \mid a, s) \| P(z \mid s)\right],
\end{aligned}
$$
where $q$ is the encoder, and $\alpha$ and $\beta$ denote the parameters of the encoder and the decoder, respectively.
The guiding principle is that if the latent policy $z = \pi(s)$ is constrained to produce a latent action $z$ which is highly probable under the prior $p(z|s)$, then the overall policy, as represented by $p_{\beta}(a|s, z = \pi(s))$, is likely to be more probable
under the behavior policy $p(a|s)$.

\begin{table*}[!t]
\small
    \centering
    \caption{The average TD error across trajectories collected using the same behaviour policy. The term ``Original'' refers to TD errors of agents trained using the entire offline dataset. ``Utilized'' and ``Excluded'' indicate the TD errors on trajectories used for training the agent and those not used for training, respectively. ``Diff'' indicates the differences between these two sets.}
    \resizebox{1\textwidth}{!}{
    \begin{tabular}{p{1.4cm}|c|ccc|ccc|ccc|ccc}
        \toprule
         \multirow{3}{*}{\textbf{Tasks}}   &  \multirow{3}{*}{\textbf{Algorithms}} & \multicolumn{3}{c|}{\multirow{2}{*}{\textbf{Original}}} &\multicolumn{9}{c}{\text{\textbf{Unlearning rates}}}   \\
         \cline{6-14}
        & & & & & \multicolumn{3}{c|}{\text{\textbf{0.1}}} & \multicolumn{3}{c|}{\text{\textbf{0.15}}} & \multicolumn{3}{c}{\text{\textbf{0.2}}}  \\
        \cline{3-14}
        & & \textbf{Utilized} & \textbf{Excluded} & \textbf{Diff }& \textbf{Utilized} & \textbf{Excluded} & \textbf{Diff } & \textbf{Utilized} & \textbf{Excluded} & \textbf{Diff } & \textbf{Utilized} & \textbf{Excluded} & \textbf{Diff }  \\
         \toprule
         \multirow{7}{*}{{\tt Hopper}} 
         & BEAR & 5.8 & 5.8 & \cellcolor{gray!20} 0.0 & 6.3 & 6.3 & \cellcolor{gray!20} 0.0 &  9.0 & 9.1  & \cellcolor{gray!20} 0.1 & 5.2 & 5.2  & \cellcolor{gray!20} 0.0 \\
         & BCQ & 7.5 & 7.5 & \cellcolor{gray!20} 0.0 & 6.9 & 6.9 & \cellcolor{gray!20} 0.0 & 7.1 & 7.1 & \cellcolor{gray!20} 0 & 6.3  & 6.3 & \cellcolor{gray!20} 0.0 \\
         & CQL & 10.6 & 10.6 & \cellcolor{gray!20} 0.0 & 10.9 &  10.9 & \cellcolor{gray!20} 0.0 &  6.5 & 6.5 & \cellcolor{gray!20}  0.0 & 9.4 & 9.4 & \cellcolor{gray!20} 0.0 \\
         & IQL & 0.5 & 0.5 & \cellcolor{gray!20} 0.0 & 0.5 & 0.5 & \cellcolor{gray!20} 0.0 & 0.4 & 0.3 & \cellcolor{gray!20} 0.1 & 0.4 & 0.4 & \cellcolor{gray!20} 0.0 \\
         & PLAS-P & 11.4 & 11.4 & \cellcolor{gray!20} 0.0  & 11.0 & 11.0 & \cellcolor{gray!20} 0.0 & 10.6 & 10.7 & \cellcolor{gray!20} 0.1 & 10.3 & 10.3 & \cellcolor{gray!20} 0.0 \\
         & TD3PlusBC & 4.2 & 4.2 & \cellcolor{gray!20} 0.0  & 4.0 & 4.0 & \cellcolor{gray!20} 0.0 & 3.8 & 3.8 & \cellcolor{gray!20} 0.0 & 3.7 & 3.7 & \cellcolor{gray!20} 0.0 \\
        \cline{2-14}
         & \textbf{Average} & \textbf{6.8} & \textbf{6.8} & \cellcolor{gray!20} \textbf{0.0} & \textbf{6.5} & \textbf{6.5} & \cellcolor{gray!20} \textbf{0.0} & \textbf{6.2} & \textbf{6.2} & \cellcolor{gray!20} \textbf{0.05} & \textbf{5.9} & \textbf{5.9} & \cellcolor{gray!20} \textbf{0.0} \\
         \toprule
        \multirow{7}{1em}{{\tt Half-} {\tt Cheetah}}
        & BEAR & 83.1 & 83.1 & \cellcolor{gray!20} 0.0 & 90.0  & 90.6 & \cellcolor{gray!20} 0.6 & 89.9 & 89.9 & \cellcolor{gray!20} 0.0 & 76.1 & 76.1 & \cellcolor{gray!20} 0.0 \\
         & BCQ & 124.3 & 123.5 & \cellcolor{gray!20} 0.8 &  189.8 & 189.8 & \cellcolor{gray!20} 0.0 & 101.4 & 100.9 & \cellcolor{gray!20} 0.5 & 92.7 & 92.7 & \cellcolor{gray!20} 0.0 \\
         & CQL & 48.5 & 48.5 & \cellcolor{gray!20} 0.0 & 62.2 & 62.2 & \cellcolor{gray!20} 0.0 & 68.2 & 67.8 & \cellcolor{gray!20} 0.4 & 49.3 & 49.3 & \cellcolor{gray!20} 0.0 \\
         & IQL & 0.8 & 0.8 & \cellcolor{gray!20} 0.0 & 0.6 & 0.6 & \cellcolor{gray!20} 0.0 & 0.7 & 0.7 & \cellcolor{gray!20} 0.0 & 0.7 & 0.8 & \cellcolor{gray!20} 0.1\\
         & PLAS-P & 70.4 & 70.4 & \cellcolor{gray!20} 0.0 & 70.9 & 70.9  & \cellcolor{gray!20} 0.0 & 64.8 & 64.8 & \cellcolor{gray!20} 0.0 & 58.2 & 58.2 & \cellcolor{gray!20} 0.0\\
         & TD3PlusBC & 111.2 & 114.3 &\cellcolor{gray!20} 3.1 & 115.3 & 115.3 & \cellcolor{gray!20} 0.0 & 117.9 & 117.9 &\cellcolor{gray!20} 0.0 & 126.3 & 126.3 & \cellcolor{gray!20} 0.0 \\
         \cline{2-14}
         & \textbf{Average} & \textbf{73.1} & \textbf{73.3} & \cellcolor{gray!20} \textbf{0.2} & \textbf{88.1} & \textbf{88.0} & \cellcolor{gray!20} \textbf{0.0} & \textbf{73.8} & \textbf{73.6} & \cellcolor{gray!20} \textbf{0.2} & \textbf{67.2} & \textbf{67.2} & \cellcolor{gray!20} \textbf{0.0} \\
         \toprule
        \multirow{7}{*}{{\tt Walker2D}} 
        & BEAR &  113.1 & 113.4  & \cellcolor{gray!20} 0.4 & 110.4 & 110.9 & \cellcolor{gray!20} 0.5 & 66.7 & 68.8 & \cellcolor{gray!20} 2.1 & 70.3 & 70.3 & \cellcolor{gray!20} 0.0 \\
        & BCQ & 116.5 & 116.5 & \cellcolor{gray!20} 0.0 & 99.8 & 98.4 & \cellcolor{gray!20} 1.4 & 78.3 & 78.3 & \cellcolor{gray!20} 0.0 & 78.3 & 78.3 & \cellcolor{gray!20} 0.0\\
         & CQL  & 121.2 & 121.2 & \cellcolor{gray!20} 0.0 & 91.6 & 91.6 & \cellcolor{gray!20} 0.0 & 64.6 & 64.6 & \cellcolor{gray!20} 0.0 & 67.9 & 67.9 & \cellcolor{gray!20} 0.0 \\
         & IQL  & 12.5 & 12.8 & \cellcolor{gray!20} 0.3 & 10.7 & 10.7 & \cellcolor{gray!20} 0.0 & 6.5 & 6.5 & \cellcolor{gray!20} 0.0 & 6.4 & 6.4 & \cellcolor{gray!20} 0.0 \\
         & PLAS-P  & 81.8 & 81.8 & \cellcolor{gray!20} 0.0 & 67.7 & 69.8 & \cellcolor{gray!20} 2.1 & 49.7 & 49.7 & \cellcolor{gray!20} 0.0 & 48.3 & 49.3 & \cellcolor{gray!20} 1.0 \\
         & TD3PlusBC  & 72.7 & 72.3 & \cellcolor{gray!20} 0.4 & 66.1 & 66.1 & \cellcolor{gray!20} 0.0 & 45.1 & 45.1 & \cellcolor{gray!20} 0.0 & 43.6 & 45.0 & \cellcolor{gray!20} 1.4 \\
         \cline{2-14}
         & \textbf{Average} & \textbf{86.3} & \textbf{86.4} & \cellcolor{gray!20} \textbf{0.1} & \textbf{74.4} & \textbf{73.7} & \cellcolor{gray!20} \textbf{0.7} & \textbf{51.8} & \textbf{52.1} & \cellcolor{gray!20} \textbf{0.3} & \textbf{52.5} & \textbf{52.9} & \cellcolor{gray!20} \textbf{0.4} \\

        \bottomrule
    \end{tabular}
    }
    \label{tab:td_error}
    \end{table*}

\subsubsection{TD3PlusBC}
TD3PlusBC is a streamlined yet remarkably efficient offline reinforcement learning algorithm that builds upon the Twin Delayed Deep Deterministic Policy Gradient (TD3)~\cite{td3} by introducing a regularization term based on Behavioral Cloning (BC)~\cite{BC}, which encourages the policy to prioritize actions that are present in the dataset $\mathcal{D}$,
$$
\pi=\underset{\pi}{\operatorname{argmax}} \mathbb{E}_{(s, a) \sim \mathcal{D}}\left[\lambda Q(s, \pi(s))-(\pi(s)-a)^2\right],
$$
where $\lambda=\frac{\alpha}{\frac{1}{N} \sum_{\left(s_i, a_i\right)}\left|Q\left(s_i, a_i\right)\right|}$ for the dataset comprises  $N$ instances of state-action pairs denoted by $s_i$ and $ a_i$. To achieve a meaningful enhancement in performance for offline reinforcement learning, TD3PlusBC applies a normalization process to each state feature within the provided dataset, which is described by the equation,
$s_i=\frac{s_i-\mu_i}{\sigma_i+\epsilon}$
,
where $s_i$ represents  the $i$th feature of
the state $s$, and $\mu_i$ and $\sigma_i$ are the mean and standard deviation of the $i$th feature throughout the dataset, respectively.

\subsection{Supplementary experiments}
\label{app:implement}

\subsubsection{Error-Based Auditing}
\label{appsub:td_auditor} 
This section investigates the variations in expected TD error (as described in Eq. (\ref{eq:forgetting_td})) between trajectories that are included in and excluded from the agent's training dataset. Thus, we can assess whether TD error is an effective auditing metric. TD error quantifies the difference between the expected reward that a model anticipates for a specific action in a given state and the actual reward it observes, in addition to the predicted reward for the next state. This prediction relies on the agent's current policy and value function. This error shares similar concepts with loss in supervised learning. The ideal assumption is that the TD errors of trajectories included in the tested agents' training dataset would be significantly smaller than those obtained from trajectories excluded from the tested agents' training dataset. 

To verify this, we train the agents using the entire offline dataset. Additionally, we retrain other agents from scratch using different unlearning rates, namely $\{0.1, 0.15, 0.2\}$. It is noticed that \textbf{the trajectories in the unlearning dataset and the remaining dataset are collected using the same behavior policy.} This ensures that the unlearning dataset is included in the original agents but excluded from the retrained agents. Subsequently, we can precisely investigate the disparity in TD errors between trajectories included in the agents' training dataset and those excluded from it.

Table~\ref{tab:td_error} presents the average TD error across trajectories for different agents. Upon observation, it becomes evident that there is minimal difference in TD errors between trajectories included in the agents' training dataset and those excluded from it. \rev{
For example, the Hopper task with CQL shows a consistent 0.0 difference. In the Half-Cheetah task, BCQ has a 0.8 difference, while BEAR at a 0.1 unlearning rate shows 0.6. The Walker2D task generally has small differences, with a maximum of 2.1 for PLAS-P at a 0.1 unlearning rate. }

However, when the trajectories in the unlearning dataset and remaining dataset are collected using different behavior policies (unlearning dataset from the `expert' policy and remaining dataset from the `medium' policy), a significant difference is observed, as shown in Table~\ref{tab:td_error_2}. For instance, the Hopper task shows average differences of 37.9, 150.8, and 75.5 for unlearning rates of 0.1, 0.15, and 0.2, respectively.

Therefore,  the difference in TD errors may simply caused by the trajectories' collected policies. These results suggest that TD error may not be a suitable metric for auditing.

\begin{table*}[!t]
\small
    \centering
    \caption{The average TD errors across trajectories are collected using different behavior policies for various agents. The term ``Original'' refers to TD errors of agents trained using the entire offline dataset. ``Utilized'' and ``Excluded'' indicate the TD errors on trajectories used for training the agent and those not used for training, which are collected using different policies.}
    \resizebox{1\textwidth}{!}{
    \begin{tabular}{p{1.4cm}|c|ccc|ccc|ccc|ccc}
        \toprule
         \multirow{3}{*}{\textbf{Tasks}}   &  \multirow{3}{*}{\textbf{Algorithms}} & \multicolumn{3}{c|}{\multirow{2}{*}{\textbf{Original}}} &\multicolumn{9}{c}{\text{\textbf{Unlearning rates}}}   \\
         \cline{6-14}
        & & & & & \multicolumn{3}{c|}{\text{\textbf{0.1}}} & \multicolumn{3}{c|}{\text{\textbf{0.15}}} & \multicolumn{3}{c}{\text{\textbf{0.2}}}  \\
        \cline{3-14}
        & & \textbf{Utilized} & \textbf{Excluded} & \textbf{Diff }& \textbf{Utilized} & \textbf{Excluded} & \textbf{Diff } & \textbf{Utilized} & \textbf{Excluded} & \textbf{Diff } & \textbf{Utilized} & \textbf{Excluded} & \textbf{Diff }  \\
         \toprule
         \multirow{7}{*}{{\tt Hopper}} 
         & BEAR & 6.8 & 37.1 & \cellcolor{gray!20} 30.3 & 7.2 & 56.5 & \cellcolor{gray!20} 49.3 &  9.0 & 413.2  & \cellcolor{gray!20} 404.2 & 5.2 & 149.0  & \cellcolor{gray!20} 143.8 \\
         & BCQ & 7.5 & 30.4 & \cellcolor{gray!20} 22.9 & 6.9 & 82.3 & \cellcolor{gray!20} 75.4 & 7.1 & 173.6 & \cellcolor{gray!20} 166.5 & 6.3  & 145.0 & \cellcolor{gray!20} 138.7 \\
         & CQL & 10.6 & 43.4 & \cellcolor{gray!20} 32.8 & 10.9 &  76.4 & \cellcolor{gray!20} 65.5 &  6.5 & 257.1 & \cellcolor{gray!20}  250.6 & 9.4 & 110.0 & \cellcolor{gray!20} 100.6 \\
         & IQL & 0.5 & 1.8 & \cellcolor{gray!20} 1.3 & 0.5 & 2.6 & \cellcolor{gray!20} 2.1 & 0.4 & 9.5 & \cellcolor{gray!20} 9.1 & 0.4 & 7.3 & \cellcolor{gray!20} 6.9\\
         & PLAS-P & 11.4 & 13.0 & \cellcolor{gray!20} 1.6  & 11.0 & 13.1 & \cellcolor{gray!20} 2.1 & 10.6 & 13.2 & \cellcolor{gray!20} 2.6 & 10.3 & 13.1 & \cellcolor{gray!20} 2.8 \\
         & TD3PlusBC & 4.2 & 15.9 & \cellcolor{gray!20} 11.7  & 4.0 & 37.3 & \cellcolor{gray!20} 33.3 & 3.8 & 75.6 & \cellcolor{gray!20} 72.8 & 3.7 & 64.0 & \cellcolor{gray!20} 60.3 \\
        \cline{2-14}
         & \textbf{Average} & \textbf{6.8} & \textbf{23.6} & \cellcolor{gray!20} \textbf{16.8} & \textbf{6.8} & \textbf{44.7} & \cellcolor{gray!20} \textbf{37.9} & \textbf{6.2} & \textbf{157.0} & \cellcolor{gray!20} \textbf{150.8} & \textbf{5.9} & \textbf{81.4} & \cellcolor{gray!20} \textbf{75.5} \\
         \toprule
        \multirow{7}{1em}{{\tt Half-} {\tt Cheetah}}
        & BEAR & 83.1 & 197.3 & \cellcolor{gray!20} 114.2 & 90.0  & 245.5 & \cellcolor{gray!20} 155.5 & 89.9 & 254.5 & \cellcolor{gray!20} 164.6 & 76.1 & 240.4 & \cellcolor{gray!20} 164.3 \\
         & BCQ & 124.3 & 212.7 & \cellcolor{gray!20} 88.4 &  189.8 & 417.9 & \cellcolor{gray!20} 228.1 & 101.4 & 255.7 & \cellcolor{gray!20} 154.3 & 92.7 & 268.1 & \cellcolor{gray!20} 175.4 \\
         & CQL & 48.5 & 132.5 & \cellcolor{gray!20} 84.0 & 62.2 & 131.3 & \cellcolor{gray!20} 69.1 & 68.2 & 150.7 & \cellcolor{gray!20} 82.5 & 49.3 & 150.0 & \cellcolor{gray!20} 100.7 \\
         & IQL & 0.6 & 0.8 & \cellcolor{gray!20} 0.2 & 0.6 & 0.6 & \cellcolor{gray!20} 0 & 0.4 & 0.7 & \cellcolor{gray!20} 0.3 & 0.5 & 0.8 & \cellcolor{gray!20} 0.3\\
         & PLAS-P & 70.4 & 157.1 & \cellcolor{gray!20} 86.7 & 70.9 & 157.5 & \cellcolor{gray!20} 86.6 & 64.8 & 156.6 & \cellcolor{gray!20} 91.8 & 58.2 & 156.7 & \cellcolor{gray!20} 98.5\\
         & TD3PlusBC & 111.2 & 118.3 &\cellcolor{gray!20} 7.1 & 115.3 & 126.4 & \cellcolor{gray!20} 11.1 & 117.9 & 144.1 &\cellcolor{gray!20} 26.2 & 126.3 & 160.6 & \cellcolor{gray!20} 34.3 \\
         \cline{2-14}
         & \textbf{Average} & \textbf{73.0} & \textbf{136.4} & \cellcolor{gray!20} \textbf{63.4} & \textbf{88.1} & \textbf{179.9} & \cellcolor{gray!20} \textbf{91.8} & \textbf{73.8} & \textbf{160.4} & \cellcolor{gray!20} \textbf{86.6} &\textbf{ 67.2} & \textbf{162.8} & \cellcolor{gray!20} \textbf{95.6} \\
         \toprule
        \multirow{7}{*}{{\tt Walker2D}} 
        & BEAR &  113.1 & 123.4  & \cellcolor{gray!20} 10.3 & 110.4 & 210.9 & \cellcolor{gray!20} 100.5 & 66.7 & 368.8 & \cellcolor{gray!20} 302.1 & 70.3 & 417.0 & \cellcolor{gray!20} 346.7\\
        & BCQ & 116.5 & 157.7 & \cellcolor{gray!20} 41.2 & 99.8 & 198.4 & \cellcolor{gray!20} 98.6 & 78.3 & 348.4 & \cellcolor{gray!20} 270.1 & 78.3 & 297.1 & \cellcolor{gray!20} 218.8\\
         & CQL  & 121.2 & 139.9 & \cellcolor{gray!20} 18.7 & 91.6 & 173.3 & \cellcolor{gray!20} 81.7 & 64.6 & 521.6 & \cellcolor{gray!20} 457 & 67.9 & 485.3 & \cellcolor{gray!20} 417.4 \\
         & IQL  & 10.5 & 12.8 & \cellcolor{gray!20} 2.3 & 10.7 & 22.3 & \cellcolor{gray!20} 11.6 & 6.5 & 32.8 & \cellcolor{gray!20} 26.3 & 6.4 & 27.7 & \cellcolor{gray!20} 21.3 \\
         & PLAS-P  & 81.8 & 114.5 & \cellcolor{gray!20} 32.7 & 67.7 & 149.8 & \cellcolor{gray!20} 82.1 & 49.7 & 192.8 & \cellcolor{gray!20} 143.1 & 48.3 & 182.3 & \cellcolor{gray!20} 134 \\
         & TD3PlusBC  & 72.7 & 77.3 & \cellcolor{gray!20} 4.6 & 66.1 & 82.8 & \cellcolor{gray!20} 16.7 & 45.1 & 161.3 & \cellcolor{gray!20} 116.2 & 43.6 & 175.8 & \cellcolor{gray!20} 132.2 \\
         \cline{2-14}
         & \textbf{Average} & \textbf{86.0} & \textbf{104.3} & \cellcolor{gray!20} \textbf{18.3} & \textbf{74.4} & \textbf{139.6} & \cellcolor{gray!20} \textbf{65.2} & \textbf{51.8} & \textbf{271.0} & \cellcolor{gray!20} \textbf{219.2} & \textbf{52.5} & \textbf{264.2} & \cellcolor{gray!20} \textbf{211.7} \\
        \bottomrule
    \end{tabular}
    }
    \label{tab:td_error_2}
    \end{table*}

\subsubsection{Fidelity evaluation of \toolname}
\label{appsub:fidelity} 

\begin{table*}[!t]
\vspace{0.1cm}
\small
    \centering
    \caption{The cumulative returns (mean $\pm$ standard variance) are averaged with mean and variance over 100 test trajectories, collected using agents trained with 5 different random seeds. ``0.01'' and ``0.05'' represent the unlearning rates. }
    \resizebox{1\textwidth}{!}{
    \begin{tabular}{c|c|cc|cc|cc|cc}
        \toprule
         \multirow{3}{*}{\textbf{Tasks}}   &  \multirow{3}{*}{\textbf{Algorithms}}  &\multicolumn{8}{c}{\text{\textbf{Unlearning Methods}}}   \\
         \cline{3-10}
        & & \multicolumn{2}{c|}{\text{\textbf{Retraining (Reference)}}} & \multicolumn{2}{c|}{\text{\textbf{Fine-tuning}}}  & \multicolumn{2}{c|}{\text{\textbf{Random-reward}}}  & \multicolumn{2}{c}{\text{\textbf{TrajDeleter}}} \\
        \cline{3-10}
        & & \textbf{0.01} & \textbf{0.05} & \textbf{0.01} & \textbf{0.05}  & \textbf{0.01} & \textbf{0.05}  & \textbf{0.01} & \textbf{0.05} \\
         \toprule
         \multirow{6}{*}{{\tt Hopper}} 
         & BEAR  & 2249 $\pm$ 818 & 1906 $\pm$ 812 &  2324 $\pm$ 79 & 2411 $\pm$ 122 & 2741 $\pm$ 17 & 2648 $\pm$ 19 & 2332 $\pm$ 345  & 2488 $\pm$ 398 \\
         & BCQ  & 3177 $\pm$ 64 & 3365 $\pm$ 40 & 3587 $\pm$ 83 & 3653 $\pm$ 7 & 3641 $\pm$ 7 & 3638 $\pm$ 25 & 3542 $\pm$ 153 & 3562 $\pm$ 83\\
         & CQL &  3625 $\pm$ 9 & 3571 $\pm$ 90 & 3635 $\pm$ 4 & 3655 $\pm$ 9 & 3630 $\pm$ 3 & 3664 $\pm$ 5 & 3637 $\pm$ 3 & 3626 $\pm$ 19 \\
         & IQL & 3633 $\pm$ 7 & 3638 $\pm$ 9 & 3394 $\pm$ 107 & 3621 $\pm$ 9 & 3631 $\pm$ 25 & 3618 $\pm$ 8 & 3601 $\pm$ 67 & 3631 $\pm$ 22 \\
         & PLAS-P &  3499 $\pm$ 14 & 3623 $\pm$ 24 & 3620 $\pm$ 22 & 3630 $\pm$ 5.2 & 3480 $\pm$ 157 & 3615 $\pm$ 17  & 3630 $\pm$ 5 & 3598 $\pm$ 63 \\
         & TD3PlusBC &  3523 $\pm$ 9 & 3498 $\pm$ 7 & 3664 $\pm$ 5.7 & 3628 $\pm$ 3 & 3632 $\pm$ 5 & 3630 $\pm$ 4  & 3562 $\pm$ 158  & 3651 $\pm$ 6  \\
         \cline{2-10} & \cellcolor{gray!20} \textbf{Average} &\cellcolor{gray!20} \textbf{3284 $\pm$ 154} &  \cellcolor{gray!20} \textbf{3267 $\pm$ 164}  & \cellcolor{gray!20} \textbf{3371 $\pm$ 50} &  \cellcolor{gray!20} \textbf{3433 $\pm$ 26}  & \cellcolor{gray!20} \textbf{3459 $\pm$ 36} &  \cellcolor{gray!20} \textbf{3469 $\pm$ 13}  & \cellcolor{gray!20} \textbf{3384 $\pm$ 122} &  \cellcolor{gray!20} \textbf{3426 $\pm$ 99} 
         \\
         \toprule
        \multirow{6}{*}{{\tt Half-}\newline {\tt Cheetah}} 
        & BEAR  & 4345 $\pm$ 785& 4343 $\pm$ 101 & 4526 $\pm$ 492 & 4438 $\pm$ 361 & 4417 $\pm$ 336 & 4392 $\pm$ 294  & 4776 $\pm$ 930  & 4735 $\pm$ 657 \\
         & BCQ &  6841 $\pm$ 1136 & 7098 $\pm$ 1076 & 7220 $\pm$ 1083 & 7523 $\pm$ 1288 & 7472 $\pm$ 1173 & 8016 $\pm$ 510 & 6885 $\pm$ 1175  & 7011 $\pm$ 841 \\
         & CQL &  1507 $\pm$ 237 & 1988 $\pm$ 235 & 1452 $\pm$ 170 & 2023 $\pm$ 104 & 1306 $\pm$ 141 & 1353 $\pm$ 206 & 1865 $\pm$ 194 & 1556 $\pm$ 171\\
         & IQL &  6345 $\pm$ 991 & 6321 $\pm$ 1048 & 6108 $\pm$ 932 & 5882 $\pm$ 859 & 6034 $\pm$ 950  & 6312 $\pm$ 868  & 6657 $\pm$ 916 & 6812 $\pm$ 952 \\
         & PLAS-P & 5321 $\pm$ 1758 & 5984 $\pm$ 944 & 4996 $\pm$ 727 & 5547 $\pm$ 1088 & 5524 $\pm$ 741 & 5174 $\pm$ 934  & 6023 $\pm$ 942  & 5947 $\pm$ 1046\\
         & TD3PlusBC &  9878 $\pm$ 1124 & 10039 $\pm$ 1681 & 10284 $\pm$ 694 & 10200 $\pm$ 857 & 9685 $\pm$ 930 & 9547 $\pm$ 930 & 9973 $\pm$ 655 & 10363 $\pm$ 842\\
         \cline{2-10} & \cellcolor{gray!20} \textbf{Average} &\cellcolor{gray!20} \textbf{5706 $\pm$ 1005} &  \cellcolor{gray!20} \textbf{5962 $\pm$ 848}  & \cellcolor{gray!20} \textbf{5764 $\pm$ 683} &  \cellcolor{gray!20} \textbf{5936 $\pm$ 760}  & \cellcolor{gray!20} \textbf{5740 $\pm$ 712} &  \cellcolor{gray!20} \textbf{5799 $\pm$ 624}  & \cellcolor{gray!20} \textbf{6030 $\pm$ 802} &  \cellcolor{gray!20} \textbf{6071 $\pm$ 752} 
         \\
         \toprule
        \multirow{6}{*}{{\tt Walker2D}} 
        & BEAR &  3181 $\pm$ 496 &
        2907 $\pm$ 517& 3102 $\pm$ 426 & 3254 $\pm$ 481 & 2263 $\pm$ 648 & 2434 $\pm$ 638 & 2265 $\pm$ 559 & 2873 $\pm$ 474 \\
        & BCQ &  2346 $\pm$ 590 & 2462 $\pm$ 524& 2758 $\pm$ 435 & 2689 $\pm$ 366 & 2083 $\pm$ 581 & 1814 $\pm$ 492 & 2688 $\pm$ 472  & 2694 $\pm$ 437\\
         & CQL  & 4286 $\pm$ 287 & 4516 $\pm$ 306 & 4408 $\pm$ 463 & 3890 $\pm$ 586 & 4521 $\pm$ 337 & 4712 $\pm$ 552 & 4724 $\pm$ 159 & 4880 $\pm$ 138 \\
         & IQL &  4159 $\pm$ 257&  4365 $\pm$ 505& 4583 $\pm$ 443 & 4170 $\pm$ 416 & 1419 $\pm$ 228 & 1426 $\pm$ 845 & 4326 $\pm$ 390  & 4041 $\pm$ 441\\
         & PLAS-P &  4163 $\pm$ 329 & 4192 $\pm$ 325 & 2784 $\pm$ 249 & 2641 $\pm$ 449 & 1471 $\pm$ 462 & 1460 $\pm$ 418  & 3841 $\pm$ 481 & 4017 $\pm$ 474\\
         & TD3PlusBC   & 4077 $\pm$ 756 & 3767 $\pm$ 511 & 4071 $\pm$ 482 & 3753 $\pm$ 716 & 3369 $\pm$ 284 & 2412 $\pm$ 643  & 4365 $\pm$ 365 & 4223 $\pm$ 464\\
         \cline{2-10} & \cellcolor{gray!20} \textbf{Average} &\cellcolor{gray!20} \textbf{3702 $\pm$ 453} &  \cellcolor{gray!20} \textbf{3702 $\pm$ 448}  & \cellcolor{gray!20} \textbf{3618 $\pm$ 416} &  \cellcolor{gray!20} \textbf{3400 $\pm$ 502}  & \cellcolor{gray!20} \textbf{2521 $\pm$ 423} &  \cellcolor{gray!20} \textbf{2376 $\pm$ 598}  & \cellcolor{gray!20} \textbf{3702 $\pm$ 404} &  \cellcolor{gray!20} \textbf{3788 $\pm$ 405} 
         \\
        \bottomrule
    \end{tabular}
    }
    \label{apptab:fidelity}
    \end{table*}

    \begin{table*}[!t]
\small
    \centering
    \caption{The cumulative returns (mean $\pm$ standard variance) are averaged with mean and variance over 100 test trajectories, collected using agents trained with 5 different random seeds. Percentage of Positive Predictions (Post. Pred.) of \toolname for unlearned agents on the unlearning dataset.}
    \resizebox{1\textwidth}{!}{
    \begin{tabular}{c|c|cc|cc|cc|cc}
        \toprule
         \multirow{3}{*}{\textbf{Tasks}}   &  \multirow{3}{*}{\textbf{Algorithms}}  &\multicolumn{8}{c}{\text{\textbf{Forgetting Training Steps}}}   \\
         \cline{3-10}
        & & \multicolumn{2}{c|}{\text{\textbf{2000}}} & \multicolumn{2}{c|}{\text{\textbf{4000}}}  & \multicolumn{2}{c|}{\text{\textbf{6000}}}  & \multicolumn{2}{c}{\text{\textbf{8000}}} \\
        \cline{3-10}
        &  & \textbf{Returns} & \textbf{Post. Pred.}  & \textbf{Returns} & \textbf{Post. Pred.}   & \textbf{Returns} & \textbf{Post. Pred.}  & \textbf{Returns} & \textbf{Post. Pred.} \\
         \toprule
         \multirow{6}{*}{{\tt Hopper}} 
         & BEAR & 3428 $\pm$ 202  & 89.4 \% & 3253 $\pm$ 383 & 59.1 \% & 3201 $\pm$ 206 & 0.0 \%  & 2332 $\pm$ 345  & 0.0 \% \\
         & BCQ & 3630 $\pm$ 28  & 77.2 \% & 3653 $\pm$ 9  & 28.1 \% & 3644 $\pm$ 4 & 0.0 \% & 3542 $\pm$ 153 & 0.0 \%\\
         & CQL & 3628 $\pm$ 34  & 79.5 \% & 3628 $\pm$ 6  & 54.6 \% & 3648 $\pm$ 5 & 5.5 \%  & 3637 $\pm$ 3 & 0.0 \% \\
         & IQL & 3637 $\pm$ 4  & 71.2 \% & 3595 $\pm$ 95 & 51.5 \% & 3598 $\pm$ 98  &  4.7 \% & 3601 $\pm$ 67 & 0.0 \% \\
         & PLAS-P & 3634 $\pm$ 6  & 68.1 \% & 3530 $\pm$ 215 & 63.6 \% & 3596 $\pm$ 55 &  45.5 \% & 3630 $\pm$ 5 & 47.5 \% \\
         & TD3PlusBC &  3622 $\pm$ 6 & 83.3 \% & 3624 $\pm$ 7 & 57.2 \% & 3631 $\pm$ 8 & 0.0 \%  & 3562 $\pm$ 158  & 0.0 \%  \\
         \cline{2-10} & \cellcolor{gray!20} \textbf{Average} &\cellcolor{gray!20} \textbf{3597 $\pm$ 47} &  \cellcolor{gray!20} \textbf{78.1\%}  & \cellcolor{gray!20} \textbf{3547 $\pm$ 119} &  \cellcolor{gray!20} \textbf{52.4 \%}  & \cellcolor{gray!20} \textbf{3553 $\pm$ 63} &  \cellcolor{gray!20} \textbf{9.3 \%}  & \cellcolor{gray!20} \textbf{3384 $\pm$ 122} &  \cellcolor{gray!20} \textbf{7.9 \%}
         \\
         \toprule
        \multirow{6}{*}{{\tt Half-}\newline {\tt Cheetah}} 
        & BEAR & 4861 $\pm$ 412  & 54.7 \% & 4552 $\pm$ 350 & 0.0 \% & 4723 $\pm$ 845 &  0.0 \% & 4776 $\pm$ 930  & 0.0 \% \\
         & BCQ &  7571 $\pm$ 997 & 17.0 \% & 7315 $\pm$ 1051 & 0.0 \% & 7024 $\pm$ 1058  &  0.0 \%  & 6885 $\pm$ 1175  & 0.0 \% \\
         & CQL &  1541 $\pm$ 204 & 27.5 \% & 1527 $\pm$ 155 & 0.0 \% & 1705 $\pm$ 173 &  0.0 \% & 1865 $\pm$ 194 & 0.0 \% \\
         & IQL &  6051 $\pm$ 1029 & 0.0 \% & 6597 $\pm$ 1128  & 0.0 \% & 7016 $\pm$ 975 & 0.0 \%  & 6657 $\pm$ 916 & 0.0 \% \\
         & PLAS-P&  6287 $\pm$ 1306 & 75.3 \% & 6893 $\pm$ 950 & 77.3 \% &  6025 $\pm$ 994 &  67.9 \% & 6023 $\pm$ 942  & 3.5 \% \\
         & TD3PlusBC & 10048 $\pm$ 394  & 0.0 \% & 10418 $\pm$ 816 & 0.0 \% & 9969 $\pm$ 759  & 0.0 \% & 9973 $\pm$ 655 & 0.0 \% \\
         \cline{2-10} & \cellcolor{gray!20} \textbf{Average} &\cellcolor{gray!20} \textbf{6060 $\pm$ 724} &  \cellcolor{gray!20} \textbf{29.1 \%} &
         \cellcolor{gray!20} \textbf{6217 $\pm$ 742} &  \cellcolor{gray!20} \textbf{12.9 \%}  & \cellcolor{gray!20} \textbf{6077 $\pm$ 801} &  \cellcolor{gray!20} \textbf{11.3 \%}  & \cellcolor{gray!20} \textbf{6030 $\pm$ 802} &  \cellcolor{gray!20} \textbf{0.6 \%}  
         \\
         \toprule
        \multirow{6}{*}{{\tt Walker2D}} 
        & BEAR & 2944 $\pm$ 559  & 58.9\% & 2832 $\pm$ 496 & 27.0 \%  & 2636 $\pm$ 508 & 7.4 \% & 2265 $\pm$ 559 & 0.6 \% \\
        & BCQ & 2912 $\pm$ 432  & 67.3 \% & 3287 $\pm$ 399 & 27.2 \% & 2803 $\pm$ 439  & 4.3 \%  & 2688 $\pm$ 472  & 0.2 \% \\
         & CQL &  4643 $\pm$ 393 & 33.3 \% & 4959 $\pm$ 137 & 23.2 \% & 4954 $\pm$ 196 & 9.4 \%  & 4724 $\pm$ 159 & 0.0 \% \\
         & IQL & 3341 $\pm$ 558 & 49.2 \% & 4434 $\pm$ 342 & 26.1 \%  & 4431 $\pm$ 357 & 5.7 \%  & 4326 $\pm$ 390  & 0.0 \% \\
         & PLAS-P & 3448 $\pm$ 296  & 59.4 \% & 3748 $\pm$ 484 & 56.2 \% & 3946 $\pm$ 463  & 67.2 \%  & 3841 $\pm$ 481 & 60.7 \% \\
         & TD3PlusBC   & 4484 $\pm$ 394  & 80.1 \% & 4561 $\pm$ 516 & 23.2 \% & 4536 $\pm$ 420 & 0.0 \% & 4365 $\pm$ 365 & 1.3 \% \\
         \cline{2-10} & \cellcolor{gray!20} \textbf{Average} &\cellcolor{gray!20} \textbf{3629 $\pm$ 439} &  \cellcolor{gray!20} \textbf{58.0 \%}  & \cellcolor{gray!20} \textbf{3970 $\pm$ 396} &  \cellcolor{gray!20} \textbf{30.5 \%}  & \cellcolor{gray!20} \textbf{3884 $\pm$ 397} &  \cellcolor{gray!20} \textbf{15.7 \%}  & \cellcolor{gray!20} \textbf{3702 $\pm$ 404} &  \cellcolor{gray!20} \textbf{10.5 \%}
         \\
        \bottomrule
    \end{tabular}
    }
    \label{tab:training_steps}
\end{table*}

We consider a setting where 1\% and 5\% of the offline dataset (unlearning rates of 0.01 and 0.05) are allocated as unlearning datasets, while the remaining data serves as the reference dataset. Retrained agents are used as a benchmark for comparing other unlearning methods (fine-tuning, random-reward, and \toolname). The unlearning steps for these methods are set at $1 \times 10^4$, which is only 1\% of the steps required for retraining ($1 \times 10^6$). The ``forgetting'' duration is 8000 timesteps, ``convergence training'' duration is 2000 timesteps.

Table~\ref{apptab:fidelity} presents the cumulative returns, represented as mean $\pm$ standard deviation, calculated by averaging the mean and variance across 100 test trajectories. These trajectories are gathered using agents trained with five distinct random seeds.

As delineated in Table~\ref{apptab:fidelity}, the agents unlearned through \toolname exhibit a performance level that closely matches that of agents retrained from the ground up. Specifically, the average cumulative returns reveal only a slight variance when comparing the unlearned agents using \toolname to those undergoing the retraining process. This finding implies that \toolname does not detrimentally affect the agents' operational effectiveness in actual environments, thus underscoring the practical viability of our method. When using the random-reward method for unlearning, there is an average performance decrease of 11.4\% (\rev{we derived this figure by averaging the performance changes relative to those of retrained agents}) in the agents, indicating that trajectories involving random rewards can detrimentally affect the performance of unlearned agents. Employing the fine-tuning method for unlearning agents results in only minimal performance degradation. However, as illustrated in Table~\ref{tab:efficacy}, fine-tuning fails to effectively eliminate the impact of target trajectories on the original agents. Consequently, the fine-tuning approach is not an optimal method for unlearning.

\begin{table*}[!t]
\vspace{0.1cm}
\small
    \centering
    \caption{The cumulative returns (mean $\pm$ standard variance) and percentage of Positive Predictions (Post. Pred.) of TrajDeleter for unlearned agents on the unlearning dataset. The forgetting training steps is 8000. The unlearning rate is 0.05.}
    \resizebox{1\textwidth}{!}{
    \begin{tabular}{c|c|cc|cc|cc|cc}
        \toprule
         \multirow{3}{*}{\textbf{Tasks}}   &  \multirow{3}{*}{\textbf{Algorithms}}  &\multicolumn{8}{c}{\text{\textbf{Value of Balancing Factor}}}   \\
         \cline{3-10}
        & & \multicolumn{2}{c|}{\text{\textbf{0.25}}} & \multicolumn{2}{c|}{\text{\textbf{0.5}}}  & \multicolumn{2}{c|}{\text{\textbf{0.75}}}  & \multicolumn{2}{c}{\text{\textbf{1.5}}} \\
        \cline{3-10}
        &  & \textbf{Returns} &\textbf{ Post. Pred.}  & \textbf{Returns} &\textbf{ Post. Pred.}  & \textbf{Returns} &\textbf{ Post. Pred.}  & \textbf{Returns} &\textbf{ Post. Pred.} \\
         \toprule
         \multirow{6}{*}{{\tt Hopper}} 
         & BEAR & 2676 $\pm$ 307  & 0.0 \% & 2821 $\pm$ 331 & 0.0 \% & 2815 $\pm$ 300 & 0.0 \%  & 2512 $\pm$ 231  & 0.0 \% \\
         & BCQ & 3520 $\pm$ 6  & 0.0 \% & 3602 $\pm$ 13  & 0.0 \% & 3522 $\pm$ 5 & 0.0 \% & 3621 $\pm$ 232 & 0.0 \%\\
         & CQL & 3637 $\pm$ 71  & 0.0 \% & 3510 $\pm$ 4  & 0.0 \% & 3425 $\pm$ 19 & 0.0 \%  & 3680 $\pm$ 6 & 0.0 \% \\
         & IQL & 3725 $\pm$ 16  & 0.0 \% & 3568 $\pm$ 104 & 0.0 \% & 3511 $\pm$ 110  &  0.0 \% & 3310 $\pm$ 37 & 0.0 \% \\
         & PLAS-P & 3600 $\pm$ 8  & 56.8 \% & 3734 $\pm$ 532 & 55.4 \% & 3600 $\pm$ 37 &  50.6 \% & 3397 $\pm$ 60 & 45.5 \% \\
         & TD3PlusBC &  3750 $\pm$ 5 & 0.0 \%  & 3638 $\pm$ 10 & 0.0 \% & 3731 $\pm$ 103 & 0.0 \%  & 3774 $\pm$ 61  & 0.0 \%  \\
         \cline{2-10} & \cellcolor{gray!20} \textbf{Average} &\cellcolor{gray!20} \textbf{3485 $\pm$ 69} &  \cellcolor{gray!20} \textbf{9.5 \%}  & \cellcolor{gray!20} \textbf{3479 $\pm$ 166} &  \cellcolor{gray!20} \textbf{9.2 \%}  & \cellcolor{gray!20} \textbf{3434 $\pm$ 96} &  \cellcolor{gray!20} \textbf{8.4 \%}  & \cellcolor{gray!20} \textbf{3382 $\pm$ 105} &  \cellcolor{gray!20} \textbf{7.6 \%}
         \\
         \toprule
        \multirow{6}{*}{{\tt Half-}\newline {\tt Cheetah}} 
        & BEAR & 4629 $\pm$ 428  & 0.0 \% & 4711 $\pm$ 551 & 0.0 \% & 4905 $\pm$ 843 &  0.0 \% & 4843 $\pm$ 726  & 0.0 \% \\
         & BCQ &  6604 $\pm$ 946 & 0.0 \% & 7546 $\pm$ 1082 & 0.0 \% & 6723 $\pm$ 1119  &  0.0 \%  & 6823 $\pm$ 1310  & 0.0 \% \\
         & CQL &  1428 $\pm$ 199 & 0.0 \% & 1689 $\pm$ 133 & 0.0 \% & 1515 $\pm$ 103 &  0.0 \% & 1715 $\pm$ 107 & 0.0 \% \\
         & IQL &  6515 $\pm$ 996 & 0.0 \% & 6597 $\pm$ 831  & 0.0 \% & 6129 $\pm$ 460 & 0.0 \%  & 6629 $\pm$ 1046 & 0.0 \% \\
         & PLAS-P&  6104 $\pm$ 1037 & 10.7 \% & 6786 $\pm$ 1050 & 7.6 \% &  6122 $\pm$ 1086 &  9.1 \% & 6012 $\pm$ 986  & 5.5 \% \\
         & TD3PlusBC & 11042 $\pm$ 312  & 0.0 \% & 9910 $\pm$ 1060 & 0.0 \% & 9992 $\pm$  623  & 0.0 \% & 9559 $\pm$ 1066 & 0.0 \% \\
         \cline{2-10} & \cellcolor{gray!20} \textbf{Average} &\cellcolor{gray!20} \textbf{6054 $\pm$ 653} &  \cellcolor{gray!20} \textbf{1.8 \%} & \cellcolor{gray!20} \textbf{6207 $\pm$ 785} &  \cellcolor{gray!20} \textbf{1.3 \%} & \cellcolor{gray!20} \textbf{5898 $\pm$ 706} &  \cellcolor{gray!20} \textbf{1.5 \%} & \cellcolor{gray!20} \textbf{5930 $\pm$ 874} &  \cellcolor{gray!20} \textbf{0.9 \%}
         \\
         \toprule
        \multirow{6}{*}{{\tt Walker2D}} 
        & BEAR & 3108 $\pm$ 386  & 1.4 \% & 2602 $\pm$ 451 & 0.0 \%  & 2739 $\pm$ 430 & 0.0 \% & 2210 $\pm$ 581 & 0.9 \% \\
        & BCQ & 2731 $\pm$ 456  & 0.0 \% & 2945 $\pm$ 456 & 0.0 \% & 2937 $\pm$ 200  & 0.1 \% & 2643 $\pm$ 658  & 0.0 \% \\
         & CQL &  4719 $\pm$ 226 & 0.0 \% & 4123 $\pm$ 233 & 0.0 \% & 4421 $\pm$ 210 & 0.0 \%  & 4132$\pm$ 365 & 0.0 \% \\
         & IQL & 4823 $\pm$ 423 & 0.0 \% & 4712 $\pm$ 471 & 0.0  \%  & 4426 $\pm$ 310 & 0.0 \%  & 4195 $\pm$ 408  & 0.0 \% \\
         & PLAS-P & 3455 $\pm$ 300  & 49.1 \% & 3927 $\pm$ 522 & 53.1 \% & 3173 $\pm$ 514  & 60.6 \%  & 3328 $\pm$ 510 & 61.9 \% \\
         & TD3PlusBC   & 4891 $\pm$ 126  & 1.1 \% & 4649 $\pm$ 479 & 0.4 \% & 4211 $\pm$ 385 & 0.0 \% & 4110 $\pm$ 286 & 1.9 \% \\
         \cline{2-10} & \cellcolor{gray!20} \textbf{Average} &\cellcolor{gray!20} \textbf{3955 $\pm$ 320} &  \cellcolor{gray!20} \textbf{8.6 \%} & \cellcolor{gray!20} \textbf{3826 $\pm$ 435} &  \cellcolor{gray!20} \textbf{8.9 \%} & \cellcolor{gray!20} \textbf{3651 $\pm$ 342} &  \cellcolor{gray!20} \textbf{10.1 \%} & \cellcolor{gray!20} \textbf{3436 $\pm$ 468} &  \cellcolor{gray!20} \textbf{10.8 \%}
         \\
        \bottomrule
    \end{tabular}
    }

    \label{tab:lambda1}
    \end{table*}

\begin{table*}[!t]
\small
    \centering
    \caption{The cumulative returns (mean $\pm$ standard variance) and percentage of Positive Predictions (Post. Pred.) of TrajDeleter for unlearned agents on the unlearning dataset. The forgetting training steps is 4000.}
    \resizebox{1\textwidth}{!}{
    \begin{tabular}{c|c|cc|cc|cc|cc}
        \toprule
         \multirow{3}{*}{\textbf{Tasks}}   &  \multirow{3}{*}{\textbf{Algorithms}}  &\multicolumn{8}{c}{\text{\textbf{Value of Balancing Factor}}}   \\
         \cline{3-10}
        & & \multicolumn{2}{c|}{\text{\textbf{0.25}}} & \multicolumn{2}{c|}{\text{\textbf{0.5}}}  & \multicolumn{2}{c|}{\text{\textbf{0.75}}}  & \multicolumn{2}{c}{\text{\textbf{1.5}}} \\
        \cline{3-10}
        &  & \textbf{Returns} & \textbf{Post. Pred.}  & \textbf{Returns} & \textbf{Post. Pred.}   & \textbf{Returns} & \textbf{Post. Pred.}  & \textbf{Returns} & \textbf{Post. Pred.} \\
         \toprule
         \multirow{6}{*}{{\tt Hopper}} 
         & BEAR & 3641 $\pm$ 36 & 54.5 \% & 3577 $\pm$ 130 & 57.3 \% & 3659 $\pm$ 24 & 60.1 \% & 3085 $\pm$ 120 & 41.3 \% \\
         & BCQ & 3558 $\pm$ 154  & 31.5 \% & 3653 $\pm$ 6 & 33.2 \% & 3641 $\pm$ 4 & 27.3 \% & 3642 $\pm$ 8 & 21.1 \% \\
         & CQL &  3633 $\pm$ 8 & 62.3 \% & 3639 $\pm$ 5 & 63.4 \% & 3638 $\pm$ 6 & 63.6 \% & 3610 $\pm$ 6 & 40.4 \% \\
         & IQL & 3630 $\pm$ 8  & 59.8 \% & 3569 $\pm$ 305 & 59.1 \% & 3622 $\pm$ 36 & 60.3 \% & 3638 $\pm$ 21 & 50.0 \% \\
         & PLAS-P & 3614 $\pm$ 36 & 63.6 \% & 3623 $\pm$ 26 & 54.5 \% & 3618 $\pm$ 35 & 63.8 \% & 3625 $\pm$ 6 & 51.9 \% \\
         & TD3PlusBC & 3624 $\pm$ 9  & 81.9 \% & 3622 $\pm$ 8 & 77.2 \% & 3610 $\pm$ 3 & 72.3 \% & 3598 $\pm$ 6 & 61.2 \% \\
         \cline{2-10} & \cellcolor{gray!20} \textbf{Average} &\cellcolor{gray!20} \textbf{3617 $\pm$ 42} &  \cellcolor{gray!20} \textbf{58.9 \%} & \cellcolor{gray!20} \textbf{3614 $\pm$ 80} &  \cellcolor{gray!20} \textbf{57.5 \%} & \cellcolor{gray!20} \textbf{3631 $\pm$ 18} &  \cellcolor{gray!20} \textbf{57.9 \%} & \cellcolor{gray!20} \textbf{3533 $\pm$ 28} &  \cellcolor{gray!20} \textbf{44.3 \%}
         \\
         \toprule
        \multirow{6}{*}{{\tt Half-}\newline {\tt Cheetah}} 
        & BEAR & 4626 $\pm$ 619  & 0.0 \% & 4731 $\pm$ 547 & 0.0 \% & 4698 $\pm$ 512 & 0.0 \% & 4649 $\pm$ 477 & 0.0 \% \\
         & BCQ & 8095 $\pm$ 958 & 0.0 \% & 7874 $\pm$ 1173 & 0.0 \% & 8044 $\pm$ 1032 & 0.0 \% & 7705 $\pm$ 977 & 0.0 \% \\
         & CQL & 1495 $\pm$ 146  & 0.0 \% & 1533 $\pm$ 194 & 0.0 \% & 1600 $\pm$ 195 & 0.0 \% & 1105 $\pm$ 159 & 0.0 \% \\
         & IQL & 6752 $\pm$ 835 & 0.0 \% & 6879 $\pm$ 794 & 0.0 \% & 6845 $\pm$ 985 & 0.0 \% & 6667 $\pm$ 957 & 0.0 \% \\
         & PLAS-P& 7286 $\pm$ 1144  & 73.4 \% & 6577 $\pm$ 1148 & 77.8 \% & 6293 $\pm$ 1072 & 75.8 \% & 6167 $\pm$ 943 & 71.3 \% \\
         & TD3PlusBC & 10021 $\pm$ 791 & 0.0 \% & 10354 $\pm$ 788 & 0.0 \% & 10089 $\pm$ 718 & 0.0 \% & 9432 $\pm$ 705 & 0.0 \% \\
         \cline{2-10} & \cellcolor{gray!20} \textbf{Average} & \cellcolor{gray!20} \textbf{6379 $\pm$ 749} &  \cellcolor{gray!20} \textbf{12.2 \%} & \cellcolor{gray!20} \textbf{6325 $\pm$ 774} &  \cellcolor{gray!20} \textbf{13.0 \%} & \cellcolor{gray!20} \textbf{6262 $\pm$ 752} &  \cellcolor{gray!20} \textbf{12.6 \%} & \cellcolor{gray!20} \textbf{5954 $\pm$ 703} &  \cellcolor{gray!20} \textbf{11.9 \%}  
         \\
         \toprule
        \multirow{7}{*}{{\tt Walker2D}} 
        & BEAR & 2999 $\pm$ 570  & 26.0 \% & 2932 $\pm$ 517 & 33.3 \% & 2772 $\pm$ 689 & 30.4 \% & 2645 $\pm$ 533 & 17.3 \% \\
        & BCQ & 2601 $\pm$ 395  & 21.9 \% & 2522 $\pm$ 454 & 24.6 \% & 2446 $\pm$ 534 & 22.9 \% & 2197 $\pm$ 575 & 21.7 \% \\
         & CQL & 4947 $\pm$ 186 & 59.3 \% & 4940 $\pm$ 270 & 31.3 \% & 4998 $\pm$ 81 & 27.3 \% & 3426 $\pm$ 778 & 20.5 \% \\
         & IQL & 4701 $\pm$ 489  & 29.3 \% & 4525 $\pm$ 348 & 27.5 \% & 4788 $\pm$ 257 & 28.3 \% & 4179 $\pm$ 376 & 20.7 \% \\
         & PLAS-P & 3434 $\pm$ 576 & 77.1 \% & 3424 $\pm$ 565 & 79.3 \% & 3248 $\pm$ 459 & 66.6 \% & 2659 $\pm$ 435 & 59.7 \% \\
         & TD3PlusBC   & 4857 $\pm$ 510  & 24.5 \% & 4873 $\pm$ 426 & 23.1 \% & 4808 $\pm$ 195 & 24.7 \% & 4323 $\pm$ 470 & 15.9 \% \\
         \cline{2-10} & \cellcolor{gray!20} \textbf{Average} &
         \cellcolor{gray!20} \textbf{3923 $\pm$ 454} &  \cellcolor{gray!20} \textbf{39.7 \%} & \cellcolor{gray!20} \textbf{3869 $\pm$ 430} &  \cellcolor{gray!20} \textbf{36.5 \%} & \cellcolor{gray!20} \textbf{3843 $\pm$ 369} &  \cellcolor{gray!20} \textbf{33.4 \%} & \cellcolor{gray!20} \textbf{3238 $\pm$ 528} &  \cellcolor{gray!20} \textbf{26.0 \%}  
         \\
        \bottomrule
    \end{tabular}
    }
    \vspace{1mm}
    \label{tab:lambda2}
    \end{table*}
    
\subsubsection{Hyper-Parameter Analysis}
\label{appsub:hyper-parameter} 

This section delves into how the forgetting learning steps, denoted as $K$, and the balancing factor, $\lambda$, influence the unlearning efficacy of \toolname. We define the forgetting learning steps, $K$, with the values $\{0, 2000, 4000, 6000, 8000\}$. To maintain a total of 10,000 unlearning steps, the corresponding training steps for convergence are accordingly set to $\{10000, 8000, 6000, 4000, 2000\}$. \rev{Additionally,  we fix the forgetting step at $K=8000$ and vary the convergence training steps $H$ across the values $\{0, 2000, 4000, 6000, 8000, 10000\}$.} Meanwhile, the balancing factor $\lambda$ is pivotal in moderating the unlearning process for forgotten datasets and the training for remaining datasets, crucially preserving the agent's performance standards. This factor is adjusted through various values: $\{0.25, 0.5, 0.75, 1.0, 1.5\}$. In these experiments, the unlearning rate is 0.05.

As defined in Section~\ref{subsec:metrics}, higher averaged cumulative returns indicate better overall utility of the unlearned agent, while a lower percentage of positive predictions means more complete forgetting of targeted trajectories.
Table~\ref{tab:training_steps} illustrates the impact of forgetting learning steps $K$ on the unlearning efficacy in \toolname. The data show that for all scenarios except the PLAS-P algorithm in the Walker2D task, an increase in $K$ leads to a significant reduction in the percentage of positive predictions. Concurrently, the average cumulative returns within each task remain consistent. These results indicate a strong sensitivity of unlearning efficacy to the magnitude of forgetting learning steps $K$, suggesting that higher $K$ values enhance unlearning effectiveness. \rev{Figure~\ref{fig:hyper-parameters-H} also presents that with only 2000 steps of convergence training, the unlearned agents can recover their performance and achieve a level of average cumulative returns comparable to that of agents before unlearning. The cumulative returns and average percentages of Positive Prediction (Post. Pred.) remain stable for agents trained with an increasing number of convergence steps.} Furthermore, the utility of the unlearned agents is largely unaffected by changes in $K$, implying that a higher $K$ may be beneficial for the overall unlearning processes.

\begin{figure*}[!t]
    \centering
    \setlength{\abovecaptionskip}{0pt}
    \includegraphics[width=1\linewidth]{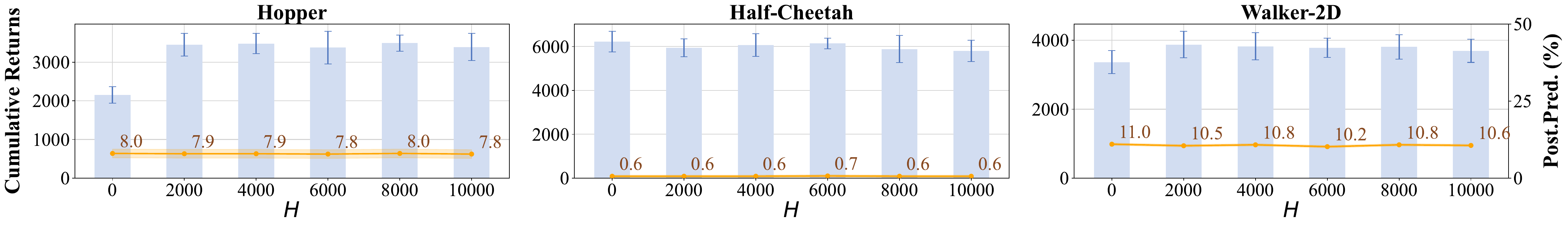}
    \caption{\rev{The influence of convergence learning steps, $H$, on the unlearning performance of \toolname. Yellow points denote the average percentages of Positive Predictions (Post. Pred.) by \toolnameeval for unlearned agents, aligning with the \textit{left} y-axis. Shaded areas denote the standard deviation. Light blue bars illustrate the average returns, corresponding to the \textit{right} y-axis, with the error bars indicating their standard deviation. The unlearning rate is 0.01.}}
    \label{fig:hyper-parameters-H}
    \vspace{-6.5mm}
\end{figure*}

\begin{table}[!t]
\small
    \centering
    \caption{The relative changes in precision, recall, and f1-score of \toolnameeval after perturbing the tested trajectories, when compared with its performance on unaltered trajectories.}
    \resizebox{0.48\textwidth}{!}{
   \begin{tabular}{c|c|ccc}
        \toprule
         \textbf{Tasks}   & \textbf{Algorithms} & \textbf{Precision} & \textbf{Recall} & \textbf{F1-score} \\
         \hline
         \multirow{7}{*}{{\tt Hopper}}
         & BEAR & -4.0 \% & -0.5 \%  & -0.02 \\
         & BCQ & -2.7 \% & 0.0 \% & -0.01 \\
         & CQL & -5.9 \% & 0.0 \% & -0.03 \\
         & IQL & -4.9 \% & 0.0 \% & -0.01 \\
         & PLAS-P & -3.0 \% & -1.2 \% & -0.01 \\
         & TD3PlusBC & -9.1 \% & 0.0 \% & -0.05 \\
        \cline{2-5}
         & \cellcolor{gray!20}  \textbf{Average} & \cellcolor{gray!20} \textbf{-4.9 \%} & \cellcolor{gray!20}\textbf{-0.3 \%} & \cellcolor{gray!20} \textbf{-0.02} \\
          \toprule
         \multirow{7}{*}{{\tt Half-Cheetah}}
         & BEAR & +12.0 \% & -13.8 \% & -0.02 \\
         & BCQ & -1.3 \% & -1.5 \% & -0.01 \\
         & CQL & -3.0 \% & +2.0 \% & -0.01 \\
         & IQL & +2.9 \% & +1.6 \% & +0.02 \\
         & PLAS-P & -3.9 \% & 0.0 \% & -0.02 \\
         & TD3PlusBC & -6.0 \% & -0.5 \% & -0.04 \\
        \cline{2-5}
         & \cellcolor{gray!20} \textbf{Average} & \cellcolor{gray!20} \textbf{+0.8 \%} & \cellcolor{gray!20} \textbf{ -2.0 \% } & \cellcolor{gray!20}\textbf{-0.02} \\
         \toprule
         \multirow{7}{*}{{\tt Walker2D}} 
         & BEAR & -1.8 \% & -1.8 \% & -0.02 \\
         & BCQ & 0.0 \% & 0.0 \% & 0.0 \\
         & CQL & -0.3 \% & -1.9 \% & -0.01 \\
         & IQL & +0.4 \% & -1.8 \% & -0.01 \\
         & PLAS-P & +0.6 \% & -4.7 \% & -0.03 \\
         & TD3PlusBC &  -1.5 \% & 0.0 \% & -0.01 \\
        \cline{2-5}
         & \cellcolor{gray!20} \textbf{Average} & \cellcolor{gray!20} \textbf{-0.4 \%} & \cellcolor{gray!20} \textbf{-1.7 \%} & \cellcolor{gray!20} \textbf{-0.02} \\
        \bottomrule
    \end{tabular}
    }
    \vspace{-1.0mm}
    \label{appsupp_tab:auditor_rob}
\end{table}

Table~\ref{tab:lambda1} delineates the influence of the balancing factor $\lambda$ on \toolname's unlearning efficacy with a constant forgetting training step count of 8000. In this scenario, most algorithms, excluding PLAS-P, exhibit nearly zero positive predictions, indicating robust unlearning efficacy that remains stable as $\lambda$ increases. Conversely, when the forgetting training steps are set at 4000, as shown in Table~\ref{tab:lambda2}, 
\rev{the average percentage of positive predictions in the three tasks decreases from 58.9\%, 12.2\%, and 39.7\% to 44.3\%, 11.9\%, and 26.0\%, respectively, as $\lambda$ increases.}
This trend suggests that a larger $\lambda$ value leads to improved unlearning efficacy when the number of forgetting steps is small. Correspondingly, the average cumulative returns demonstrate a reduction in three tasks when $\lambda$ ranges from 0.25 to 1.5.

In summary, the forgetting learning steps $K$ exert a more significant impact on unlearning efficacy compared to the balancing factor $\lambda$. A higher $K$ enhances unlearning efficacy while maintaining the utility of the agent. Conversely, an increased $\lambda$ value improves unlearning efficacy only when $K$ is moderate, but it may negatively affect unlearned agents.

\subsection{Additional Discussions}

This section explores the robustness of \toolnameeval and threats to validity, and the limitations of our paper.

\subsubsection{Robustness of \toolnameeval} 
\label{supsubsec:robustness}

Prior studies also explore the robustness of the auditing~\cite{du2023orl}: whether trajectory with some random noises can still be accurately identified as part of an agent's training dataset. For example, the MuJoCo environments also add small random noises to sensor information (i.e., observing states). Additionally, the offline RL agent deployed in real-world decision-making tasks that frequently utilizes Gaussian noise to improve the agents' generalization capabilities~\cite{alhinai2020introduction}. Thus, introducing Gaussian noise into the states of trajectories is a subtle approach to evade detection by an auditor. An effective trajectory auditing method expects to detect trajectories reliably, even if such noise exists. 

We follow the MuJoCo documentation\footnote{\url{https://www.gymlibrary.dev/environments/mujoco/}} to introduce Gaussian noise into each state of the trajectories. This noise follows a distribution with a mean of 0 and a standard deviation of 0.05.
Our experimental results presented in Table~\ref{appsupp_tab:auditor_rob} reveal that \toolnameeval is only slightly affected by the Gaussian noise. When processing trajectories that have been perturbed with this noise, the average F1-score of \toolnameeval decreases by both 0.02 across the three evaluated tasks compared to the performance on original trajectories. This consistency in performance highlights the robustness of \toolnameeval to maintain accuracy and reliability under varying conditions. 

\subsubsection{Threats to Validity}
\label{subsub:threat}

The performance of offline RL algorithms is highly contingent on hyperparameters. To mitigate potential threats to internal validity, we adhere to the hyper-parameter settings provided in the D3RLPY~\cite{d3rlpy} replication package. We benchmark the performance of agents developed in this study against those documented in D3RLPY~\cite{d3rlpy}, affirming the accurate replication of investigated offline RL algorithms.
RL agents usually face the notorious problem of unstable performance~\cite{sutton2018reinforcement,sac,cql}. To reduce the impact of environmental randomness and enhance the construct validity, we train the agents using five different random seeds, and each agent is allowed to interact with the environment to generate 100 trajectories. We then use the average cumulative returns from these trajectories to measure the agent's performance. The findings presented in this paper might have limited applicability to other 
offline datasets or algorithms. Our experiments are performed using the dataset from the benchmark recently introduced in~\cite{d4rl}, as well as six advanced offline RL algorithms, to relieve the threats to external validity.

\subsubsection{Effectiveness Uncertainty.}
\label{app:eff_un}
\rev{We agree that, as proposed in \toolname, poor performance on specific trajectories cannot ensure that the agents have forgotten the trajectories. It is noticed that as an approximate method, the unlearning effect relies more on empirical evaluations; this is a common practice. In addition, the evaluation of \toolname depends on the effectiveness of the trajectories auditing method, i.e., \toolnameeval. This issue is also a common challenge in previous unlearning research across various fields~\cite{unlearning_02_feature,unlearning_survey}. We propose potential solutions by enhancing the auditor tools. Improving the auditor tool enhances unlearning methods and increases their trustworthiness. We remain open-minded about this issue, emphasizing that our conclusions are based on the presented effectiveness of \toolnameeval.}

\subsubsection{Towards Legal Requirements}
\label{app:legal}

\rev{This paper focuses on approximate unlearning. We summarize challenges and limitations that need further exploration to satisfy legal requirements.

\begin{itemize}[leftmargin=*]
\item \textbf{Adequacy of Data Removal}: Approximate unlearning may not effectively remove all influence of an individual's data from the model as GDPR and other data protection regulations require, potentially leaving residual data that could still affect the model's output. We need to verify and validate the effectiveness of data unlearning methods.

\item \textbf{Legal Acceptance}: Legal standards require both transparency and demonstrable efficacy; thus, developers of approximate unlearning techniques need to ensure that their methods are transparent and understandable to regulators.

\item \textbf{Evolving Legal Standards}: Changes in data protection laws could impact the acceptance of approximate unlearning methods. Legal reforms or clarifications could either facilitate or hinder the adoption of these methods.

\end{itemize}

Despite these uncertainties, approximate unlearning reduces computational overhead and enables more dynamic data management practices in ML, making it a valuable research area. Ensuring that approximate unlearning meets the same standards as full retraining is challenging. We have made every effort to address this issue in our paper. We also recommend developing certified unlearning methods to advance practical unlearning applications. Besides, interdisciplinary research that includes legal experts and technologists is crucial for navigating the compliance landscape and crafting approximate unlearning methods that are both technically effective and legally robust.}

\subsection{Related works}
\label{suppsec:related}

\subsubsection{Offline RL for Real-World Applications}
Recently, offline RL systems work brilliantly on a wide range of real-world fields, including healthcare~\cite{RL4Treatment,RL4BGC,offline_rl_medicial}, energy management systems~\cite{RL4Energy}, autonomous driving~\cite{RL4AutonomousVehicles,RL4AutonomousVehicles2}, recommendation systems~\cite{RL4Recommender,RL4Recommender2} and dialog systems~\cite{RL4Dialog,RL4Dialog2}. 
In healthcare, online RL is not suitable, as it is ethically and practically problematic to experiment with patients' health. Thus, Mila et al.~\cite{RL4Treatment} used advanced offline RL methods to develop a policy for recommending diabetes and sepsis treatment optimization. Meanwhile, Emerson et al.~\cite{RL4BGC} applied offline RL to determine the optimal insulin dose for maintaining blood glucose levels within a healthy range. Additionally, in various areas, using existing data to learn a policy proved significantly more efficient than online RL methods~\cite{RL4Energy,RL4AutonomousVehicles,RL4Dialog,RL4Dialog3}. In energy management, Zhan et al.~\cite{RL4Energy} introduced a model-based offline RL algorithm aimed at refining the energy combustion control strategy for thermal power generating units (TPGUs). By integrating extensive amounts of historical data from TPGUs with low-fidelity simulation data, they could derive a policy that operates TPGUs within safety constraints. In autonomous driving, researchers gather diverse driving behaviors from multiple drivers and then train the planning algorithm using offline RL methods~\cite{RL4AutonomousVehicles,RL4AutonomousVehicles2,RL4AutonomousVehicles3}. Offline RL systems have also made great progress in developing dialog systems. Verma et al.~\cite{RL4Dialog} integrated offline RL methods with language models to propose training realistic dialogue agents, promoting the efficiency of completing human-interactive tasks. 

As pointed out by Prudencio et al.~\cite{offline_survey}, many research opportunities remain for evaluating, testing, and safeguarding offline RL systems, especially when using them in the real world. This paper addresses a particular privacy challenge to advance the practical implementation of offline RL.

\subsubsection{Machine Unlearning}
Deep machine unlearning~\cite{unlearning_01,unlearning_02_random,unlearning_02_feature,unlearning_03_understanding} refers to eliminating the knowledge of specific data point(s) on the already trained Deep Neural Networks (DNNs). This concept gains particular significance in privacy and data protection legislation, such as the European Union's General Data Protection Regulation (GDPR)~\cite{GDPR}, which mandates a ``right to erasure." Deep machine unlearning is categorized into two main groups: exact unlearning~\cite{unlearning_01,unlearning_04_amnesiac_exact,unlearning_05_ARCANE_exact} and approximate unlearning methods~\cite{unlearning_03_understanding,unlearning_06_selective_app,unlearning_07_mixed_app,unlearning_09_zero_app}. In exact unlearning, the most straightforward approach is retraining the DNN from scratch thought, excluding the data that is requested to be forgotten from the training set. This is computationally intensive, especially considering DNNs are typically trained on large datasets~\cite{unlearning_08_resource}. Bourtoule et al.~\cite{unlearning_01} introduced the SISA method for training DNNs by dividing the dataset into non-overlapping shards, thereby diminishing the necessity for complete retraining since the DNN can be retrained on just one of these shards. Leveraging the one-class classifier, Yan et al.~\cite{unlearning_05_ARCANE_exact} developed a method that accelerates the SISA while also ensuring the accuracy of the retrained model within a large number of unlearning requests. 

Unlike exact unlearning guaranteeing that the outputs of an unlearned DNN and a fully retrained DNN are indistinguishable, approximate unlearning aims to estimate the parameters of DNNs in a manner analogous to retraining the network from scratch~\cite{unlearning_survey}. Warnecke et al.~\cite{unlearning_02_feature} established changes in the training dataset to closed-form updates of the DNN parameters, enabling direct adjustments to the DNN parameters in response to unlearning requests. However, this method is only applicable to tabular data. Golatkar et al.~\cite{unlearning_06_selective_app,unlearning_10_scrubbing_app} presented an approximation of the training process based on the neural tangent kernel and used it to predict the updated DNNs parameters
after unlearning. Unrolling SGD~\cite{unlearning_03_understanding} operated approximate unlearning by directly stochastic ascent using data to be unlearned.

This paper is the first to develop a method to satisfy the eager unlearning requirements within offline RL.

\subsubsection{Certified Unlearning}
A range of recent works studied certified unlearning definitions~\cite{unlearning_11_removal_cer,unlearning_12_deletion_cer}. Certified unlearning
offers the theoretical guarantee that the unlearned model is indistinguishable from a DNN retrained from scratch on the remaining dataset. Besides, for approximate unlearning, Guo et al.~\cite{unlearning_11_removal_cer}, and Sekhari et al.~\cite{sekhari2021remember} have proposed methods to ensure that the distributions of an unlearned model and a retrained model are nearly indistinguishable, using differential privacy techniques. Warnecke et al. ~\cite{unlearning_02_feature} introduced a certified unlearning scheme for systematically removing learned features and labels from models.

\end{document}